\RequirePackage{tikz}
\usetikzlibrary{
	positioning,      calc, arrows,
	arrows.meta,      fit,  shapes,
	shapes.multipart, decorations
}

\documentclass{article}

\usepackage[preprint]{neurips_2021}

\usepackage[mode=build]{standalone}
\usepackage[utf8]{inputenc} 
\usepackage{url}            
\usepackage{booktabs}       
\usepackage{amsfonts}       
\usepackage{amsmath}
\usepackage{nicefrac}       
\usepackage{xcolor}         
\usepackage{csquotes}
\usepackage{tikz}
\usepackage{multicol}
\usepackage{pgfplots}
\usepackage{hyperref}
\pgfplotsset{compat=newest}
\usepgfplotslibrary{groupplots,dateplot}
\usetikzlibrary{patterns,shapes.arrows}

\usepackage{acro}

\DeclareAcronym{gan}{
  short = GAN,
  long = generative adversarial neural network,
}
\DeclareAcronym{cnn}{
    short = CNN,
    long = convolutional neural network,
}
\DeclareAcronym{fwt}{
  short = FWT,
  long = fast wavelet transform,
}
\DeclareAcronym{ifwt}{
  short = iFWT,
  long = inverse fast wavelet transform,
}
\DeclareAcronym{fft}{
    short = FFT,
    long = fast Fourier transform,
}
\DeclareAcronym{dct}{
  short = DCT,
  long = discrete cosine transform,
}
\DeclareAcronym{knn}{
    short = kNN,
    long = $k$-nearest neighbors,
}
\DeclareAcronym{prnu}{
	short = PRNU,
	long = photoresponse non-uniformity,
    cite = marra2019gans
}
\DeclareAcronym{ffhq}{
  short = FFHQ,
  long = Flickr Faces High Quality,
}
\DeclareAcronym{lsun}{
  short = LSUN,
  long = Large-scale Scene UNderstanding
}
\DeclareAcronym{celeba}{
  short = CelebA,
  long = Large-scale Celeb Faces Attributes
}
\DeclareAcronym{ln}{
  short = $\ln$,
  long = natural logarithm
}
\DeclareAcronym{fft2}{
  short = fft2,
  long = Two dimensional fast Fourier transform
}
\DeclareAcronym{ffpp}{
  short = ff++,
  long = Face Forensics++
}

\usepackage{xcolor}

\title{Wavelet-Packets for Deepfake Image Analysis and Detection}

\author{%
  Moritz Wolter \\
  High Performance Computing \& Analytics Lab, \\
  Universität Bonn \\
  \texttt{moritz.wolter@uni-bonn.de} \\
  \And
  Felix Blanke  \\
  Fraunhofer SCAI \\
  University of Bonn \\
  \texttt{felix.blanke@scai.fraunhofer.de}
  \And
  Raoul Heese  \\
  Fraunhofer Center for Machine Learning \\
  and Fraunhofer ITWM \\
  \texttt{raoul.heese@itwm.fraunhofer.de}
  \And
  Jochen Garcke   \\
  Institute for Numerical Simulation, \\
  University of Bonn \\
  Fraunhofer Center for Machine Learning \\ and Fraunhofer SCAI\\
  \texttt{garcke@ins.uni-bonn.de}
}

\begin{document}

\maketitle

\begin{abstract}
As neural networks become able to generate realistic artificial images, they have the potential to improve movies, music, video games and make the internet an even more creative and inspiring place. Yet, the latest technology potentially enables new digital ways to lie. In response, the need for a diverse and reliable method toolbox arises to identify artificial images and other content. Previous work primarily relies on pixel-space \aclp{cnn} or the Fourier transform. Synthesized fake image analysis and detection methods based on a multi-scale wavelet representation, localized in both space and frequency, have been absent thus far. The wavelet transform conserves spatial information to a degree, allowing us to present a new analysis. Comparing the wavelet coefficients of real and fake images significant differences are identified, where this representation also allows further interpretations. Additionally, this paper proposes to learn a model for the detection of synthetic images based on the wavelet-packet representation of natural and \ac{gan}-generated images. Our forensic classifiers exhibit competitive or improved performance at small network sizes, as we demonstrate on the \ac{ffhq}, \ac{celeba} and \ac{lsun} source identification problems. Furthermore, we study the binary \ac{ffpp} fake-detection problem.
\end{abstract}

\section{Introduction}\label{sec:introduction}
While \acp{gan} can extract useful representations from data, translate textual descriptions into images, transfer scene and style information between images or detect objects~\citep{review-gan}, they can also enable abusive actors to quickly and easily generate potentially damaging, highly realistic fake images, colloquially called deepfakes~\citep{guardian2019DeepFakeDemocracy,frank2020frequency}.
As the internet becomes a more prominent virtual public space for political discourse and social media outlet~\citep{Applebaum2020twilight}, deepfakes present a looming threat to its integrity that must be met with techniques for differentiating the real and trustworthy from the fake.

Previous algorithmic techniques for separating real from computer-hallucinated images of people have relied on identifying fingerprints in \textit{either} the spatial~\citep{yu2019attributing,marra2019gans} or frequency~\citep{frank2020frequency,dzanic2019fourier} domain.
However, to the best of our knowledge, no techniques have jointly considered the two domains in a multi-scale fashion, for which we propose to employ wavelet-packet coefficients representing a spatio-frequency representation.

In this paper, we make the following contributions:
\begin{itemize}
  \item We present a wavelet-\textit{packet}-based analysis of \ac{gan}-generated images. Compared to existing work in the frequency domain, we examine the spatio-frequency properties of \ac{gan}-generated content for the first time. We find differences between real and synthetic images in both the wavelet-packet mean and standard deviation, with increasing frequency and at the edges. The differences in mean and standard deviation also holds between different sources of synthetic images.
  \item To the best of our knowledge, we present the first application and implementation of \textit{boundary} wavelets for image analysis in the deep learning context. Proper implementation of boundary-wavelet filters allows us to share identical network architectures across different wavelet lengths.
  \item As a result of the aforementioned wavelet-\textit{packet}-based analysis, we build classifiers to identify image sources. We work with fixed seed values for reproducibility and report mean and standard deviations over five runs whenever possible. Our systematic analysis shows improved or competitive performance.
  \item Integrating existing Fourier-based methods, we introduce fusion networks combining both approaches. Our best networks outperform previous purely Fourier or pixel-based approaches on the CelebA and Lsun bedrooms benchmarks. Both benchmarks have been previously studied in \cite{yu2019attributing} and \cite{frank2020frequency}.
\end{itemize}
We believe our virtual public spaces and social media outlets will benefit from a growing, diverse toolbox of techniques enabling automatic detection of \ac{gan}-generated content, where the introduced wavelet-based approach provides a competitive and interpretable addition.
The source code for our wavelet toolbox, the first publicly available toolbox for fast boundary-wavelet transforms in the python world, and our experiments is available at \url{https://github.com/v0lta/PyTorch-Wavelet-Toolbox} and \url{https://github.com/gan-police/frequency-forensics}.

\section{Motivation}\label{sec:motivation}
\enquote{Wavelets are localized waves. Instead of oscillating forever, they drop to zero.}~\cite{strang1996wavelets}.
This important observation explains why wavelets allow us to
conserve some spatio-temporal relations.
The Fourier transform uses sine and cosine waves, which never stop fluctuating.
Consequently, Fourier delivers an excellent frequency resolution while losing all spatio-temporal information.
We want to learn more about the spatial organization of differences in the frequency domain, which is
our first motivation to work with wavelets.

The second part of our motivation lies in the representation power of wavelets.
Images often contain sharp borders where pixel values change rapidly. Monochromatic areas are
also common, where the signal barely changes at all.
Sharp borders or steps are hard to represent in the Fourier space. As the sine and cosine waves do not fit well, we observe the Gibbs-phenomenon \citep{strang1996wavelets} as high-frequency coefficients pollute the spectrum.
For a dull, flat region, Fourier requires an infinite sum \citep{van2004wavelets}.
The Fourier transform is uneconomical in these two cases.
Furthermore, the fast wavelet transform can be used with many different wavelets. Having a choice that allows us to
explore and choose a basis which suits our needs is our second motivation to work with wavelets.



\begin{figure}
\centering
\includegraphics[width=\textwidth]{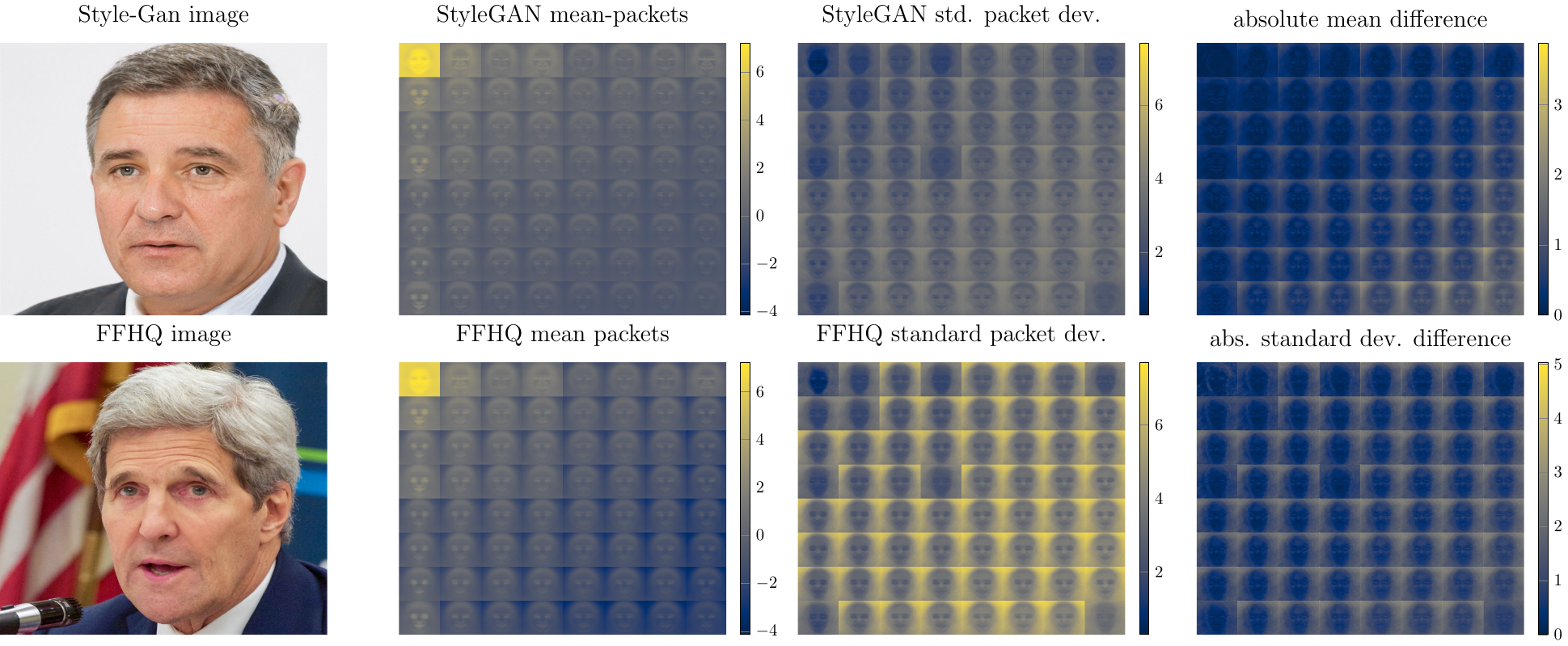}
\caption{%
This figure compares single time-domain images and absolute $\ln$-scaled mean wavelet packets and their standard deviations. The first column shows two example images, the second mean packets, the third standard deviations, and the fourth mean and standard deviation differences. Mean values and standard deviations have been computed on 5k $1024 \times 1024$ \ac{ffhq}- and StyleGAN-generated images each.
The order of the packets has the frequency increasing from the top left to the bottom right. We observe significant differences. The mean is significantly different for higher frequencies and at image edges. The standard deviations differ in the background across the entire spectrum. In the supplementary, Figure~\ref{fig:packetlabels} shows the exact filter sequences for each packet. Best viewed in color.%
}
\label{fig:proof_of_concept}
\end{figure}

Before discussing the mechanics of the \ac{fwt} and its packet variant, we present a proof of concept experiment in Figure~\ref{fig:proof_of_concept}.
We investigate the coefficients of the level 3 Haar wavelet packet transform.
Computations use 5k $1024 \times 1024$ pixel images from \acf{ffhq} and 5k $1024 \times 1024$ pixel images generated by StyleGAN.
The leftmost columns of Figure~\ref{fig:proof_of_concept} show a sample from both \ac{ffhq} and a StyleGAN generated one. Mean wavelet packets appear in the second, and their standard deviation in the third column. Finally, the rightmost column plots the absolute mean packets and standard deviation differences.
For improved visibility, we rescale the absolute values of the packet representation, averaged first over the three color bands, using the \ac{ln}.

We find that the mean wavelet packets of \ac{gan}-generated images are often significantly brighter. The differences become more pronounced as the frequency increases from the top left to the bottom right. The differences in high-frequency packets independently confirm \cite{dzanic2019fourier}, who saw the most significant differences for high-frequency Fourier-coefficients.

The locality of the wavelet filters allows face-like shapes to appear, but the image edges stand out, allowing us to pinpoint the frequency disparities. We now have an intuition regarding the local origin of the differences.
We note that Haar wavelet packets do not use padding or orthogonalization and conclude that the differences stem from \ac{gan} generation.
For the standard deviations, the picture is reversed. Instead of the GAN the \ac{ffhq}-packets appear brighter. The StyleGAN packets do not deviate as much as the data in the original data set.
The observation suggests that the \ac{gan} did not capture the complete variance of the original data. Our evidence indicates that \ac{gan}-generated backgrounds are of particular monotony across all frequency bands.

In the next sections, we survey related work and discuss how the \acf{fwt} and wavelet packets in Figure~\ref{fig:proof_of_concept} are computed.

\section{Related work}\label{sec:related}

\subsection{Generative adversarial networks}
The advent of \acp{gan}~\citep{goodfellow2014generative} heralded several successful image generation projects.
We highlight the contribution of the progressive growth technique, in which a small network is trained on smaller images then gradually increased in complexity/number of weights and image size, on the ability of optimization to converge at high resolutions of $1024 \times 1024$ pixels \citep{karras2019pro}.
As a supplement, style transfer methods \citep{gatys2016imagestyle} have been integrated into new style-based generators.
The resulting style-\acp{gan} have increased the statistical variation of the hallucinated faces \citep{karras2019stylebased,karras2020analyzing}.
Finally, regularization (e.g.,\ using spectral methods) has increased the stability of the optimization process~\citep{Miyato2018Spectral}.
The recent advances have allowed large-scale training \citep{Brock2019Large}, which in turn improves the quality further.

Conditional methods allow to partially manipulate images, \cite{antipov2017face} for example proposes to use conditional \acp{gan} for face aging. Similarly, \acp{gan} allow the transfer of facial expressions \citep{deepfake2022github,thies2019deferred}, an ability of particular importance from a detection point of view.
For a recent exhaustive review on \ac{gan} theory, architecture, algorithms, and applications, we refer to~\cite{review-gan}.

\subsection{Diffusion Probabilistic Models}
As an alternative class of approaches to \acp{gan}, recently, two closely related probabilistic generative models, diffusion (probabilistic) models and score-based generative models, respectively, were shown to produce high-quality images, see~\cite{Ho2020diffusionmodel,dhariwal2021diffusion,song2021scorebased} and their references.
In short, images are generated by reversing in the sampling phase a gradual noising process used during training.
Further, \cite{dhariwal2021diffusion} proposes to use gradients from a classifier to guide a diffusion model during sampling, with the aim to trade-off diversity for fidelity.

\subsection{Deepfake detection}
Deepfake detectors broadly fall into two categories. The first group works in the frequency domain. Projects include the study of natural and deep network generated images in the frequency space created by the \ac{dct}, as well as detectors based on Fourier features~\citep{zhang2019detecting,durall2019unmasking,dzanic2019fourier,frank2020frequency,durall2020watch,Giudice2021}, these provide frequency space information, but the global nature of these bases means all spatial relations are missing. In particular, \cite{frank2020frequency} visualizes the \ac{dct} transformed images and identifies artifacts created by different upsampling methods. We learn that the transformed images are efficient classification features, which allow significant parameter reductions in GAN identification classifiers. Instead of relying on the \ac{dct}, \cite{dzanic2019fourier} studied the distribution of Fourier coefficients for real and \ac{gan}-generated images. After noting significant deviations of the mean frequencies for \ac{gan} generated- and real-images, classifiers are trained. Similarly, \cite{Giudice2021} studies statistics of the \ac{dct} coefficients and uses estimations for each GAN-engine for classification. \cite{dzanic2019fourier} found high-frequency discrepancies for GAN-generated imagery, built detectors, and spoofed the newly trained Fourier classifiers by manually adapting the coefficients in Fourier-space. Finally, \cite{he2021Beyond} combined 2d-FFT and DCT features and observed improved performance.

The second group of classifiers works in the spatial or pixel domain, among others, \citep{yu2019attributing,Wang2020,wang2020cnn,zhao2021multi,zhao2021learning} train (convolutional) neural networks directly on the raw images to identify various \ac{gan} architectures. Building upon pixel-CNN \cite{wang2021representative} adds neural attention.
According to \citep{yu2019attributing}, the classifier features in the final layers constitute the fingerprint for each \ac{gan} and are interesting to visualize.
Instead of relying on deep-learning, \citep{marra2019gans} proposed \ac{gan}-fingerprints working exclusively with denoising-filter and mean operations, whereas~\citep{Guarnera2020} computes convolutional traces.

\cite{Tang2021} uses a simple first-level discrete wavelet transform, this means no multi-scale representation as a preprocessing step to compute spectral correlations between the color bands of an image, which are then further processed to obtain features for a classifier.
\cite{younus2020effective} use a Haar-wavelet transform to study edge inconsistencies in manipulated images.

To the best of our knowledge, we are the first to work directly with a spatio-frequency wavelet \textit{packet} representation in a multi-scale fashion. Our analysis goes beyond previous Fourier-based studies. Fourier loses all spatial information while wavelets preserve it. Compared to previous wavelet-based work, we also consider the packet approach and boundary wavelets. The packet approach allows us to decompose high-frequency bands in fine detail. Boundary wavelets enable us to work with higher-level wavelets, which the machine learning literature has largely ignored thus far.
As we demonstrate in the next section, wavelet packets allow visualization of frequency information while, at the same time, preserving local information.

Previously proposed deepfake detectors had to detect fully \citep{frank2020frequency} or partially~\citep{rossler2019faceforensics++} faked images. In this paper we do both. Sections~\ref{sec:ffhq_linear}, \ref{sec:train_celeba_lsun} and \ref{sec:big_ffhq} consider complete fakes and section~\ref{sec:ffpp} studies detection of partial neural fakes. Additionally we examine images generated by guided diffusion \citep{dhariwal2021diffusion} in section \ref{sec:guided_diffusion}. We demonstrate our ability to identify images from this source by training a forensic classifier. A recent survey on deepfake generation and detection can be found in~\cite{Zhang2022}.

\subsection{Wavelets}
Originally developed within applied mathematics \citep{mallat1989theory,daubechies1992tenwavelets},
wavelets are a long-established part of the image analysis and processing toolbox \citep{taubman2002jpeg2000,strang1996wavelets,jensen2001ripples}.
In the world of traditional image coding, wavelet-based codes have been developed to replace techniques based on the \ac{dct}~\citep{taubman2002jpeg2000}.
Early applications of wavelets in machine learning studied the integration of wavelets into neural networks for function learning \citep{zhang1995wavelet}.
Within this line of work, wavelets have previously served as input features \citep{chen2015automatic}, or as early layers of scatter nets \citep{mallat2012group,cotter2020uses}.
Deeper inside neural networks, wavelets have appeared within pooling layers using static Haar \citep{wang2020haar,williams2018wavelet} or adaptive wavelets \citep{wolter2021adaptive,ha2021adaptive}.
Within the subfield of generative networks, concurrent research \citep{gal2021swagan} explored the use of Haar-wavelets to improve \ac{gan} generated image content.
\cite{Bruna2013InvariantSC} and \cite{oyallon2015deep} found that fixed wavelet filters in early convolution layers can lead to performance similar to trained neural networks. The resulting architectures are known as scatter nets. This paper falls into a similar line of work by building neural network classifiers on top of wavelet packets. We find this approach is particularly efficient in limited training data scenarios. Section~\ref{sec:data_reduced} discusses this observation in detail. 

The machine learning literature often limits itself to the Haar wavelet \citep{wang2020haar,williams2018wavelet,gal2021swagan}. Perhaps because it is padding-free and does not require boundary value treatment. The following sections will explain the \ac{fwt} and how to work with longer wavelets effectively.

\section{Methods}
In this section, we briefly present the nuts and bolts of the fast wavelet transform and its packet variant.
For multi-channel color images, we transform each color separately. For simplicity, we will only consider single channels in the following exposition.

\subsection{\texorpdfstring{The \acf{fwt}}{The fast wavelet transform (FWT)}}\label{subsec:fwt}
\acp{fwt}~\citep{mallat1989theory,daubechies1992tenwavelets,strang1996wavelets,jensen2001ripples,mallat2009wavelet} utilize convolutions to decompose an input signal into its frequency components. Repeated applications of the wavelet transform result in a multi-scale analysis. Convolution is here a linear operation and linear operations are often written in matrix form. Consequently, we aim to find a matrix that allows the computation of the fast wavelet transform~\citep{strang1996wavelets}:
\begin{align}
\mathbf{b} = \mathbf{A}\mathbf{x}.
\end{align}
$\mathbf{A}$ is a product of multiple scale-matrices. The non-zero elements in $\mathbf{A}$ are populated with the coefficients from the selected filter pair.
Given the wavelet filter degree $d$, each filter has $N = 2d$ coefficients. Repeating diagonals compute convolution operations with the so-called analysis filter vector pair $\mathbf{f}_\mathcal{L}$ and $\mathbf{f}_\mathcal{H}$, where the filters are arranged as vectors in $\mathbb{R}^N$. The subscripts $\mathcal{L}$ denote the one-dimensional low-pass and $\mathcal{H}$ the high pass filter, respectively.
The filter pair appears within the diagonal patterns of the stride two convolution matrices $\mathbf{H}_\mathcal{L}$ and $\mathbf{H}_\mathcal{H}$. Overall one observes the pattern \citep{strang1996wavelets}
\begin{align}\label{eq:analysis}
\mathbf{A}=
\dots
\begin{pmatrix}
\begin{array}{c|c}
\mathbf{H}_\mathcal{L} &  \\
\mathbf{H}_\mathcal{H} &  \\ \hline
  & \mathbf{I} \\
\end{array}
\end{pmatrix}
\begin{pmatrix}
\mathbf{H}_\mathcal{L} \\ \mathbf{H}_\mathcal{H}
\end{pmatrix}.
\end{align}
The equation describes the first two FWT-matrices. Instead of the dots, we can imagine additional analysis matrices.
The analysis matrix $\mathbf{A}$ records all operations by matrix multiplication.
\begin{figure}
  \includegraphics[width=\textwidth]{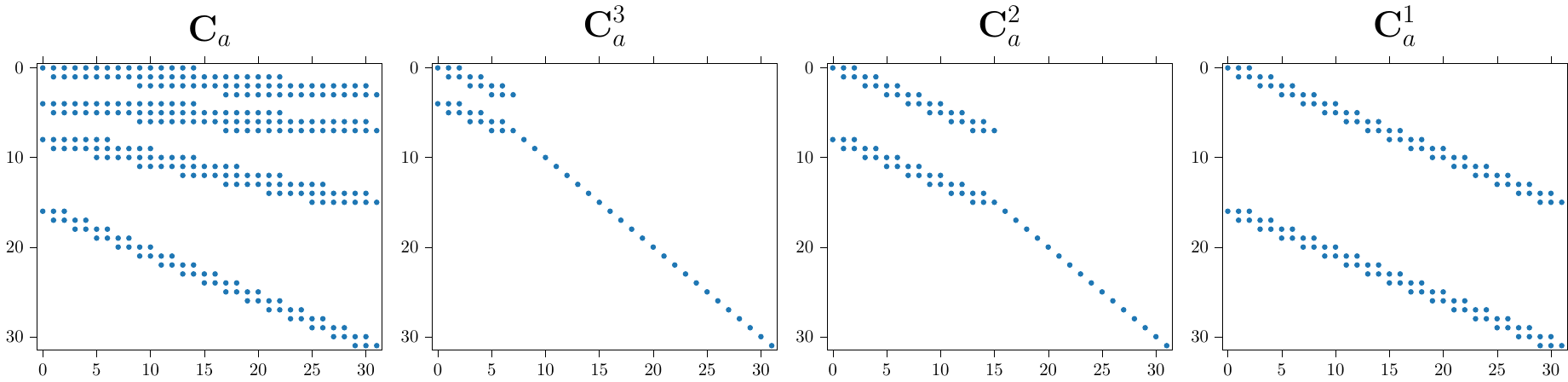}
  \caption{Truncated single-dimensional analysis convolution matrices for a signal of length 32 using for example a Daubechies-two wavelet. The decomposition level increases from right to left. On the leftmost side the product of the first three is shown. We are looking at truncated versions of the infinite matrices described by equation~(\ref{eq:analysis}). The title reads $\mathbf{C}_a$ because the analysis convolution matrix is different from the wavelet matrix, which we discuss further in section \ref{sec:boundary_wavelets} on boundary filters.}
  \label{fig:conv_analysis}
\end{figure}
In Figure~\ref{fig:conv_analysis} we illustrate a level three transform, where we see equation~(\ref{eq:analysis}) at work. Ideally, $\mathbf{A}$ is infinitely large and orthogonal. For finite signals, the ideal matrices have to be truncated. $\mathbf{C}$ denotes finite length untreated convolution matrices, subscript $a$ and $s$ mark analysis, and transposed synthesis convolutions. Second-degree wavelet coefficients from four filters populate the convolution matrices.
The identity tail of the individual matrices grows as scales complete. The final convolution matrix is shown on the left. Supplementary section~\ref{subsec:ifwt} presents the construction of the synthesis matrices, which undo the analysis steps.

Common choices for 1D wavelets are the \emph{Daubechies-wavelets} (\enquote{db}) and their less asymmetrical variant, the \emph{symlets} (\enquote{sym}).
We refer the reader to supplementary section~\ref{sec:daubsym} or \cite{mallat2009wavelet} for an in-depth discussion.
Note that the \ac{fwt} is invertible, to construct the synthesis matrix $\mathbf{S}$ for $\mathbf{S} \mathbf{A}~=~\mathbf{I}$, we require the synthesis filter pair $\mathbf{f}_\mathcal{L}, \mathbf{f}_\mathcal{H}$. The filter coefficients populate transposed convolution matrices. Details are discussed in supplementary section~\ref{subsec:ifwt}.
To ensure the transform is invertible and visualizations interpretable, not just any filter will do.
The perfect reconstruction and anti-aliasing conditions must hold. We briefly present these two in supplementary section~\ref{subsec:pc_ac} and refer to \cite{strang1996wavelets} and \cite{jensen2001ripples} for an excellent further discussion of these conditions.

\subsection{The two-dimensional wavelet transform}
To extend the single-dimensional case to two-dimensions, one dimensional wavelet pairs are often transformed. Outer products allow us to obtain two dimensional quadruples from single dimensional filter pairs~\citep{vyas2018multiscale}:
\begin{align}\label{eq:filter_conversion}
  \mathbf{f}_{a} &= \mathbf{f}_\mathcal{L}\mathbf{f}_\mathcal{L}^T, &
  \mathbf{f}_{h} &= \mathbf{f}_\mathcal{L}\mathbf{f}_\mathcal{H}^T, &
  \mathbf{f}_{v} &= \mathbf{f}_\mathcal{H}\mathbf{f}_\mathcal{L}^T, &
  \mathbf{f}_{d} &= \mathbf{f}_\mathcal{H}\mathbf{f}_\mathcal{H}^T.
\end{align}
In the equations above, $\mathbf{f}$ denotes a filter vector. In the two-dimensional case, \textit{a} denotes the approximation coefficients, \textit{h} denotes the horizontal coefficients, \textit{v} denotes vertical coefficients, and \textit{d} denotes the diagonal coefficients.
The 2D transformation at representation level $q+1$ requires the input $\mathbf{x}_q$ as well as a filter quadruple $\mathbf{f}_{k}$ for $k \in [a, h, v, d]$ and is computed by
\begin{align}
  \mathbf{x}_q*\mathbf{f}_k = \mathbf{k}_{q+1},
\end{align}
where $*$ denotes stride two convolution.
Note the input image is at level zero, i.e.\, $\mathbf{x}_0= \mathbf{I}$, while for stage $q$ the low pass result of $q-1$ is used as input.
\begin{figure}
  \includegraphics[width=\textwidth]{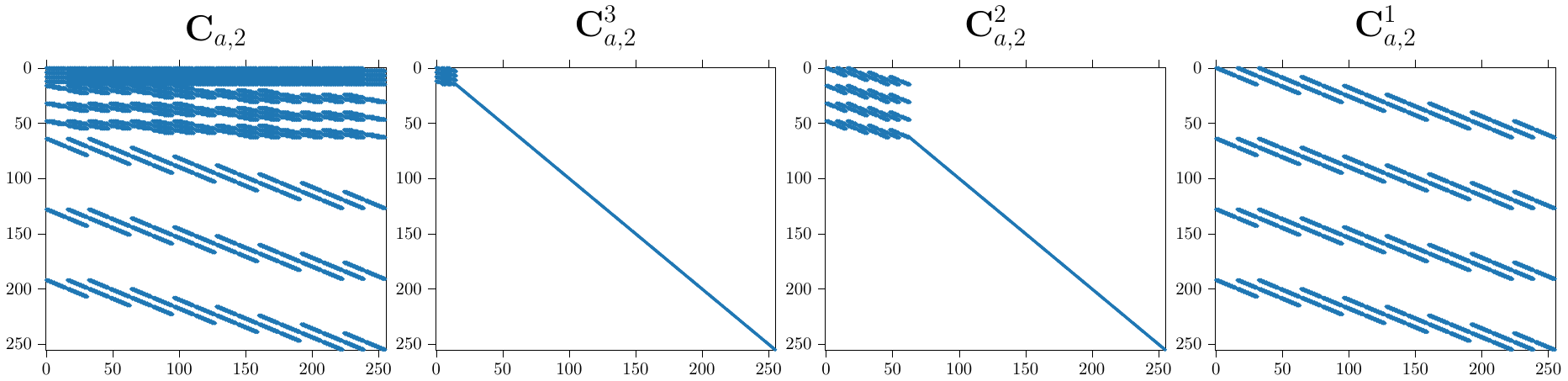}
  \caption{Two dimensional analysis convolution matrix sparsity patterns. The first three matrices from the right are individual convolution matrices. The matrix on the very left is the combined sparse convolution matrix.}
  \label{fig:raw_analysis2d}
\end{figure}
The two dimensional transform is also linear and can therefore be expressed in matrix form:
\begin{align}\label{eq:analysis2d}
  \mathbf{A}_{2d}=
  \dots
  \begin{pmatrix}
  \begin{array}{c|c}
  \mathbf{H}_a &  \\
  \mathbf{H}_h &  \\
  \mathbf{H}_v &  \\
  \mathbf{H}_d &  \\ \hline
    & \mathbf{I} \\
  \end{array}
  \end{pmatrix}
  \begin{pmatrix}
    \mathbf{H}_a &  \\
    \mathbf{H}_h &  \\
    \mathbf{H}_v &  \\
    \mathbf{H}_d &  \\
    \end{pmatrix}.
\end{align}
Similarly to the single dimensional case $\mathbf{H}_k$ denotes a stride two, two dimensional convolution matrix. We write the inverse or synthesis operation as $\mathbf{F}_k$. Figure~\ref{fig:raw_analysis2d} illustrates the sparsity patterns of the resulting two-dimensional convolution matrices.

\subsection{Boundary wavelets}\label{sec:boundary_wavelets}
\begin{figure}
  \centering
  \includegraphics[width=0.4\linewidth]{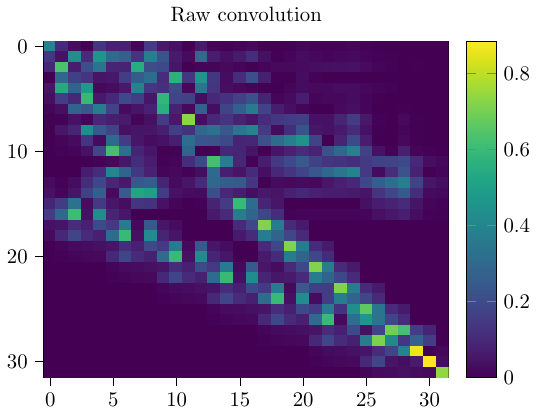}
  \includegraphics[width=0.4\linewidth]{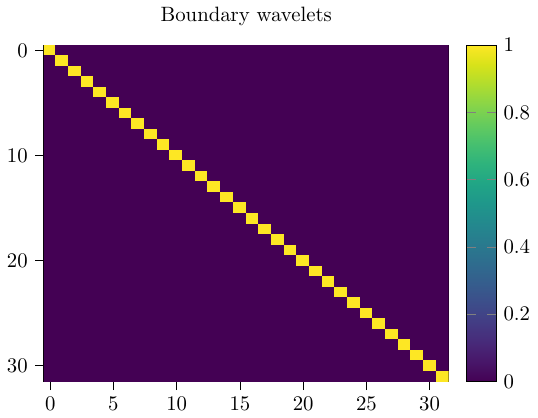}
  \caption{The effect of boundary wavelet treatment. Single-dimensional Transformation-Matrices of shape 32 by 32 are constructed. The plot on the left shows the element-wise absolute values of $\mathbf{C_s} \cdot \mathbf{C_a}$. The right plot shows the element-wise absolute values of $\mathbf{S} \cdot \mathbf{A}$ for orthogonalized analysis and synthesis matrices. The identity matrix indicates that our matrices have been correctly assembled.}
  \label{fig:boundary}
\end{figure}

So far, we have described the wavelet transform without considering the finite size of the images.
For example, the simple Haar wavelets can be used without modifications in such a case.
But, for the transform to preserve all information and be invertible, higher-order wavelets require modifications at the boundary~\citep{strang1996wavelets}.
There are different ways to handle the boundary, including zero-padding, symmetrization, periodic extension, and specific filters on the boundary. The disadvantage of zero-padding or periodic extensions is that discontinuities are artificially created at the border. With symmetrization, discontinuities of the first derivative arise at the border~\citep{jensen2001ripples}. For large images, the boundary effects might be negligible. However, for the employed multi-scale approach of wavelet-packets, as introduced in the next subsection, the artifacts become too severe. Furthermore, zero-padding increases the number of coefficients, which in our application would need different neural network architectures per wavelet.
Therefore we employ special boundary filters in the form of the so-called Gram-Schmidt boundary filters~\citep{jensen2001ripples}.

The idea is now to replace the filters at the boundary with specially constructed, shorter filters that preserve both the length and the perfect reconstruction property or other properties of the wavelet transform.
We illustrate the impact of the procedure in Figure~\ref{fig:boundary}. The product of the untreated convolution matrices appears on the left. The boundary wavelet matrices $\mathbf{S} \cdot \mathbf{A}$ on the right. Appendix Figures \ref{fig:boundary_analysis2d} and \ref{fig:boundary_synthesis2d} illustrate the sparsity patterns of the resulting sparse analysis and synthesis matrices for the Daubechies two case in 2D.

\subsection{Wavelet packets}\label{subsec:wavelet-packets}
Presuming to find the essential information in the lower frequencies, note that standard wavelet transformations decompose only the low-pass or $a$ coefficients further.
The $h$, $v$, and $d$ coefficients are left untouched.
While this is often a reasonable assumption~\citep{strang1996wavelets}, previous work~\citep{dzanic2019fourier,schwarz2021frequency} found higher frequencies equally relevant for deepfake detection.
For this analysis the wavelet tree will consequently be extended on both the low and high frequency sides.
This approach is known as wavelet packet analysis.
For a wavelet packet representation, one recursively continues to filter the low- and high-pass results.
Each recursion leads to a new level of filter coefficients,
starting with an input image $\mathbf{I} \in \mathbb{R}^{h,w}$, and using $\mathbf{n}_{0,0} = \mathbf{I}$.
A node $n_{q,j}$ at position $j$ of level $q$, is convolved with all filters $\mathbf{f}_k$, $k \in [a, h, v, d]$:
\begin{align} \label{eqn:wavpack2d}
  \mathbf{n}_{q,j}*\mathbf{f}_k = \mathbf{n}_{q+1,k}.
\end{align}
Once more, the star $*$ in the equation above denotes a stride two convolution.
Therefore every node at level $q$ will spawn four nodes at the next level $q+1$.
The result at the final level $Q$, assuming Haar or boundary wavelets without padding, will be a $4^Q \times \frac{h} {2^Q} \times \frac{w}{2^Q}$ tensor, i.e.\, the number of coefficients is the same as before and is denoted by $Q^\circ$.
Thereby wavelet packets provide filtering of the input into progressively finer equal-width blocks, with no redundancy.
For excellent presentations of the one-dimensional case we again refer to \cite{strang1996wavelets} and \cite{jensen2001ripples}.

\begin{figure}
  \centering
  \begin{tikzpicture}
  \node[draw, rectangle] at (0,0)   (I){$\mathbf{I}$};

  \node[draw, rectangle] at (-4.25, -1) (Ha){$\mathbf{H_a}\downarrow_2$};
  \node[draw, rectangle] at (-1, -1) (Hz){$\mathbf{H_h}\downarrow_2$};
  \node[draw, rectangle] at ( 1, -1) (Hv){$\mathbf{H_v}\downarrow_2$};
  \node[draw, rectangle] at ( 3, -1) (Hd){$\mathbf{H_d}\downarrow_2$};

  \draw[->] (I) -| (Ha);
  \draw[->] (I) -| (Hz);
  \draw[->] (I) -| (Hv);
  \draw[->] (I) -| (Hd);

  \node[draw, rectangle] at (-4.25, -2) (a){$\mathbf{a}$};
  \node[draw, rectangle] at (-1, -2) (z){$\mathbf{h}$};
  \node[draw, rectangle] at ( 1, -2) (v){$\mathbf{v}$};
  \node[draw, rectangle] at ( 3, -2) (d){$\mathbf{d}$};

  \draw[->] (Ha) -- (a);
  \draw[->] (Hz) -- (z);
  \draw[->] (Hv) -- (v);
  \draw[->] (Hd) -- (d);

  \node[draw, rectangle] at (-6.5, -3) (Haa){$\mathbf{H_a}$};
  \node[draw, rectangle] at (-5, -3) (Haz){$\mathbf{H_h}$};
  \node[draw, rectangle] at (-3.5, -3) (Hav){$\mathbf{H_v}$};
  \node[draw, rectangle] at (-2, -3) (Had){$\mathbf{H_d}$};

  \node[draw, rectangle] at (-6.5,  -3.75) (aa){$\mathbf{aa}$};
  \node[draw, rectangle] at (-5,  -3.75) (az){$\mathbf{ah}$};
  \node[draw, rectangle] at (-3.5,-3.75) (av){$\mathbf{av}$};
  \node[draw, rectangle] at (-2,  -3.75) (ad){$\mathbf{ad}$};

  \draw[->] (Haa) -- (aa);
  \draw[->] (Haz) -- (az);
  \draw[->] (Hav) -- (av);
  \draw[->] (Had) -- (ad);

  \draw[->] (a) -| (Haa);
  \draw[->] (a) -| (Haz);
  \draw[->] (a) -| (Hav);
  \draw[->] (a) -| (Had);

  \node[draw, rectangle] at (-6.5,  -6) (Faa){$\mathbf{F_a}$};
  \node[draw, rectangle] at (-5, -6) (Faz){$\mathbf{F_h}$};
  \node[draw, rectangle] at (-3.5, -6) (Fav){$\mathbf{F_v}$};
  \node[draw, rectangle] at (-2,   -6) (Fad){$\mathbf{F_d}$};
  
  \draw[->, dashed] (aa) -- (Faa);
  \draw[->, dashed] (az) -- (Faz);
  \draw[->, dashed] (av) -- (Fav);
  \draw[->, dashed] (ad) -- (Fad);

  \node[draw, rectangle] at (-4.25,-7) (Fa){$\mathbf{F_a}$};
  \node[draw, rectangle] at (-1,  -7) (Fz){$\mathbf{F_h}$};
  \node[draw, rectangle] at ( 1,  -7) (Fv){$\mathbf{F_v}$};
  \node[draw, rectangle] at ( 3,  -7) (Fd){$\mathbf{F_d}$};

  \draw[->] (Faa) |- (Fa);
  \draw[->] (Faz) |- (Fa);
  \draw[->] (Fav) |- (Fa);
  \draw[->] (Fad) |- (Fa);

  \draw[->, dashed] (z) -- (Fz);
  \draw[->, dashed] (v) -- (Fv);
  \draw[->, dashed] (d) -- (Fd);

  \node[draw, rectangle] at (0,-8)   (hI){$\hat{\mathbf{I}}$};

  \draw[->] (Fa) |- (hI);
  \draw[->] (Fz) |- (hI);
  \draw[->] (Fv) |- (hI);
  \draw[->] (Fd) |- (hI);

\end{tikzpicture}
  \caption{%
      Visualization of the 2D wavelet packet analysis and synthesis transform. The analysis filters are written as $\mathbf{H}_k$, synthesis filters as $\mathbf{F}_k$.
      We show all first level coefficients as well as some second level coefficients ${aa}, {ah}, {av}$, ${ad}$.
      The dotted lines indicate the omission of further possible analysis and synthesis steps.
      The transform is invertible in principle, $\hat{\mathbf{I}}$ denotes the reconstructed original input.}
      \label{fig:packet_schematic}
\end{figure}
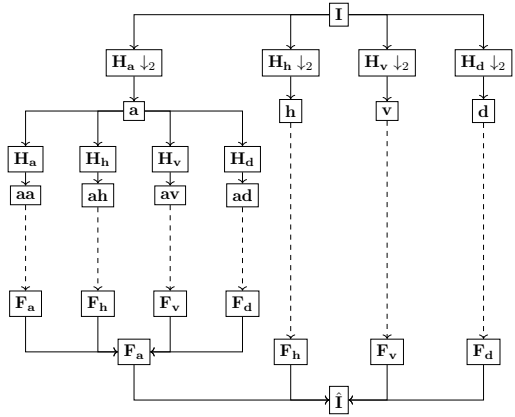
We show a schematic drawing of the process on the left of Figure~\ref{fig:packet_schematic}.
The upper half shows the analysis transform, which leads to the coefficients we are after. The lower half shows the synthesis transform, which allows inversion for completeness. Finally, for the correct interpretation of Figure~\ref{fig:proof_of_concept}, Supplementary-Figure~\ref{fig:packetlabels} lists the exact filter combinations for each packet.
The fusion networks in sections~\ref{sec:train_celeba_lsun} and onward use multiple complete levels at the same time.

\section{Classifier design and evaluation \label{sec:experiments}}
\begin{figure}
  \centering
  \includestandalone[width=.45\textwidth]{figures/concept/absolute_coeff_comparison}
\begin{tikzpicture}

\definecolor{color0}{rgb}{0.12156862745098,0.466666666666667,0.705882352941177}
\definecolor{color1}{rgb}{1,0.498039215686275,0.0549019607843137}
\definecolor{color2}{rgb}{0.172549019607843,0.627450980392157,0.172549019607843}
\definecolor{color3}{rgb}{0.83921568627451,0.152941176470588,0.156862745098039}

\begin{axis}[
legend cell align={left},
legend style={
  fill opacity=0.8,
  draw opacity=1,
  text opacity=1,
  at={(0.97,0.03)},
  anchor=south east,
  draw=white!80!black
},
tick align=outside,
tick pos=left,
title={FFHQ-StyleGAN source identification},
x grid style={white!69.0196078431373!black},
xlabel={training steps},
xmin=-67.5, xmax=2517.5,
xtick style={color=black},
y grid style={white!69.0196078431373!black},
ylabel={mean accuracy},
ymin=0.644521305914138, ymax=1.01544481639717,
ytick style={color=black}
]
\path [draw=color0, fill=color0, opacity=0.2]
(axis cs:50,0.697818534518451)
--(axis cs:50,0.661381465481549)
--(axis cs:100,0.748436122998221)
--(axis cs:150,0.728740182094711)
--(axis cs:200,0.758257301119027)
--(axis cs:250,0.759777148268905)
--(axis cs:300,0.790270957166156)
--(axis cs:350,0.757374414899444)
--(axis cs:400,0.797684135045505)
--(axis cs:450,0.799145065245338)
--(axis cs:500,0.779860317166525)
--(axis cs:550,0.789501801874927)
--(axis cs:600,0.823849458887134)
--(axis cs:650,0.747252510403055)
--(axis cs:700,0.782717292536025)
--(axis cs:750,0.795114756033832)
--(axis cs:800,0.838016713221965)
--(axis cs:850,0.76508506489111)
--(axis cs:900,0.80214276264013)
--(axis cs:950,0.792785656646343)
--(axis cs:1000,0.78167953967756)
--(axis cs:1050,0.834834945152372)
--(axis cs:1100,0.823872651426168)
--(axis cs:1150,0.77257812772408)
--(axis cs:1200,0.829863077484333)
--(axis cs:1250,0.820571618694718)
--(axis cs:1300,0.802597444589558)
--(axis cs:1350,0.801234851848282)
--(axis cs:1400,0.809491564092761)
--(axis cs:1450,0.844060285965415)
--(axis cs:1500,0.827947045007412)
--(axis cs:1550,0.832979064728366)
--(axis cs:1600,0.817747000359914)
--(axis cs:1650,0.806594454766719)
--(axis cs:1700,0.801384402653749)
--(axis cs:1750,0.856094144816182)
--(axis cs:1800,0.839848120433563)
--(axis cs:1850,0.804993637099755)
--(axis cs:1900,0.816571826573955)
--(axis cs:1950,0.808326584227748)
--(axis cs:2000,0.813548424233101)
--(axis cs:2050,0.816329593101099)
--(axis cs:2100,0.833233796978855)
--(axis cs:2150,0.838617764647235)
--(axis cs:2200,0.788960006404783)
--(axis cs:2250,0.836769178650687)
--(axis cs:2300,0.839361924855645)
--(axis cs:2350,0.829797673160243)
--(axis cs:2400,0.838332252240792)
--(axis cs:2400,0.867367747759208)
--(axis cs:2400,0.867367747759208)
--(axis cs:2350,0.871702326839757)
--(axis cs:2300,0.876038075144355)
--(axis cs:2250,0.881330821349313)
--(axis cs:2200,0.882639993595217)
--(axis cs:2150,0.873782235352764)
--(axis cs:2100,0.878766203021145)
--(axis cs:2050,0.863270406898901)
--(axis cs:2000,0.872251575766899)
--(axis cs:1950,0.862373415772252)
--(axis cs:1900,0.884128173426045)
--(axis cs:1850,0.860006362900246)
--(axis cs:1800,0.866451879566437)
--(axis cs:1750,0.873605855183818)
--(axis cs:1700,0.878015597346251)
--(axis cs:1650,0.863405545233281)
--(axis cs:1600,0.842752999640086)
--(axis cs:1550,0.861420935271634)
--(axis cs:1500,0.859252954992588)
--(axis cs:1450,0.867139714034585)
--(axis cs:1400,0.873808435907239)
--(axis cs:1350,0.846965148151718)
--(axis cs:1300,0.847602555410442)
--(axis cs:1250,0.845528381305281)
--(axis cs:1200,0.860436922515667)
--(axis cs:1150,0.83352187227592)
--(axis cs:1100,0.847027348573832)
--(axis cs:1050,0.848965054847629)
--(axis cs:1000,0.866620460322441)
--(axis cs:950,0.838814343353657)
--(axis cs:900,0.828057237359869)
--(axis cs:850,0.86101493510889)
--(axis cs:800,0.843983286778035)
--(axis cs:750,0.822885243966168)
--(axis cs:700,0.839082707463975)
--(axis cs:650,0.819947489596945)
--(axis cs:600,0.846950541112866)
--(axis cs:550,0.833898198125073)
--(axis cs:500,0.822739682833475)
--(axis cs:450,0.814854934754662)
--(axis cs:400,0.824015864954495)
--(axis cs:350,0.816125585100556)
--(axis cs:300,0.817329042833845)
--(axis cs:250,0.790322851731095)
--(axis cs:200,0.796942698880973)
--(axis cs:150,0.781459817905289)
--(axis cs:100,0.764663877001779)
--(axis cs:50,0.697818534518451)
--cycle;

\path [draw=color1, fill=color1, opacity=0.2]
(axis cs:50,0.993274807680927)
--(axis cs:50,0.990025192319073)
--(axis cs:100,0.989006262254949)
--(axis cs:150,0.983879021248497)
--(axis cs:200,0.988581114644945)
--(axis cs:250,0.984783651411076)
--(axis cs:300,0.992760928057033)
--(axis cs:350,0.995475595575915)
--(axis cs:400,0.994678416163742)
--(axis cs:450,0.993417112004706)
--(axis cs:500,0.992665343170243)
--(axis cs:550,0.995015343170243)
--(axis cs:600,0.992955689174251)
--(axis cs:650,0.993171328814745)
--(axis cs:700,0.995042893218813)
--(axis cs:750,0.993371364106259)
--(axis cs:800,0.994676731691925)
--(axis cs:850,0.994794577999171)
--(axis cs:900,0.994832292174797)
--(axis cs:950,0.995447226495737)
--(axis cs:1000,0.996387627564304)
--(axis cs:1050,0.995853580586141)
--(axis cs:1100,0.99264197578358)
--(axis cs:1150,0.994191032168799)
--(axis cs:1200,0.996019337613708)
--(axis cs:1250,0.995475595575915)
--(axis cs:1300,0.99501421833084)
--(axis cs:1350,0.996387596159536)
--(axis cs:1400,0.994981561285864)
--(axis cs:1450,0.994975204931802)
--(axis cs:1500,0.996063997742667)
--(axis cs:1550,0.994109651209943)
--(axis cs:1600,0.995868011627972)
--(axis cs:1650,0.995705795063664)
--(axis cs:1700,0.99468318452974)
--(axis cs:1750,0.995097416095367)
--(axis cs:1800,0.996206601886794)
--(axis cs:1850,0.993508846547471)
--(axis cs:1900,0.996383974596216)
--(axis cs:1950,0.993299038639272)
--(axis cs:2000,0.996640454702086)
--(axis cs:2050,0.995369659156917)
--(axis cs:2100,0.994918861169916)
--(axis cs:2150,0.995161590127327)
--(axis cs:2200,0.995359651209944)
--(axis cs:2250,0.995231394239305)
--(axis cs:2300,0.995915153077165)
--(axis cs:2350,0.996621767001687)
--(axis cs:2400,0.997016146087399)
--(axis cs:2400,0.998083853912602)
--(axis cs:2400,0.998083853912602)
--(axis cs:2350,0.997978232998312)
--(axis cs:2300,0.997384846922835)
--(axis cs:2250,0.997668605760695)
--(axis cs:2200,0.997940348790056)
--(axis cs:2150,0.997738409872672)
--(axis cs:2100,0.998081138830084)
--(axis cs:2050,0.997130340843083)
--(axis cs:2000,0.997959545297914)
--(axis cs:1950,0.998500961360728)
--(axis cs:1900,0.998116025403784)
--(axis cs:1850,0.996891153452529)
--(axis cs:1800,0.998093398113206)
--(axis cs:1750,0.998002583904633)
--(axis cs:1700,0.998416815470259)
--(axis cs:1650,0.996994204936336)
--(axis cs:1600,0.997931988372027)
--(axis cs:1550,0.996690348790056)
--(axis cs:1500,0.997836002257333)
--(axis cs:1450,0.998024795068198)
--(axis cs:1400,0.998118438714136)
--(axis cs:1350,0.998012403840464)
--(axis cs:1300,0.99728578166916)
--(axis cs:1250,0.996524404424085)
--(axis cs:1200,0.997680662386292)
--(axis cs:1150,0.996708967831201)
--(axis cs:1100,0.99825802421642)
--(axis cs:1050,0.997246419413859)
--(axis cs:1000,0.997612372435696)
--(axis cs:950,0.997252773504263)
--(axis cs:900,0.996967707825203)
--(axis cs:850,0.997105422000829)
--(axis cs:800,0.997923268308075)
--(axis cs:750,0.997328635893741)
--(axis cs:700,0.996457106781186)
--(axis cs:650,0.995728671185254)
--(axis cs:600,0.997344310825749)
--(axis cs:550,0.998584656829757)
--(axis cs:500,0.996234656829758)
--(axis cs:450,0.997782887995294)
--(axis cs:400,0.996321583836258)
--(axis cs:350,0.996524404424085)
--(axis cs:300,0.996639071942966)
--(axis cs:250,0.993516348588924)
--(axis cs:200,0.995218885355055)
--(axis cs:150,0.995220978751503)
--(axis cs:100,0.992193737745051)
--(axis cs:50,0.993274807680927)
--cycle;

\path [draw=color2, semithick]
(axis cs:2400,0.996783149944393)
--(axis cs:2400,0.998176850055607);

\path [draw=color3, semithick]
(axis cs:2400,0.805517508111212)
--(axis cs:2400,0.855602491888788);

\addplot [semithick, color0, mark=*, mark size=1.25, mark options={solid}]
table {%
50 0.6796
100 0.75655
150 0.7551
200 0.7776
250 0.77505
300 0.8038
350 0.78675
400 0.81085
450 0.807
500 0.8013
550 0.8117
600 0.8354
650 0.7836
700 0.8109
750 0.809
800 0.841
850 0.81305
900 0.8151
950 0.8158
1000 0.82415
1050 0.8419
1100 0.83545
1150 0.80305
1200 0.84515
1250 0.83305
1300 0.8251
1350 0.8241
1400 0.84165
1450 0.8556
1500 0.8436
1550 0.8472
1600 0.83025
1650 0.835
1700 0.8397
1750 0.86485
1800 0.85315
1850 0.8325
1900 0.85035
1950 0.83535
2000 0.8429
2050 0.8398
2100 0.856
2150 0.8562
2200 0.8358
2250 0.85905
2300 0.8577
2350 0.85075
2400 0.85285
};
\addlegendentry{pixel validation accuracy}
\addplot [semithick, color1, mark=*, mark size=1.25, mark options={solid}]
table {%
50 0.99165
100 0.9906
150 0.98955
200 0.9919
250 0.98915
300 0.9947
350 0.996
400 0.9955
450 0.9956
500 0.99445
550 0.9968
600 0.99515
650 0.99445
700 0.99575
750 0.99535
800 0.9963
850 0.99595
900 0.9959
950 0.99635
1000 0.997
1050 0.99655
1100 0.99545
1150 0.99545
1200 0.99685
1250 0.996
1300 0.99615
1350 0.9972
1400 0.99655
1450 0.9965
1500 0.99695
1550 0.9954
1600 0.9969
1650 0.99635
1700 0.99655
1750 0.99655
1800 0.99715
1850 0.9952
1900 0.99725
1950 0.9959
2000 0.9973
2050 0.99625
2100 0.9965
2150 0.99645
2200 0.99665
2250 0.99645
2300 0.99665
2350 0.9973
2400 0.99755
};
\addlegendentry{packet validation acc}
\addplot [semithick, color2, mark=x, mark size=1.25, mark options={solid}]
table {%
2400 0.99748
};
\addlegendentry{packet test accuracy}
\addplot [semithick, color3, mark=x, mark size=1.25, mark options={solid}]
table {%
2400 0.83056
};
\addlegendentry{pixel test accuracy}
\end{axis}

\end{tikzpicture}
  \caption{%
      The mean of level $3$ Haar wavelet packet coefficient values for each 63k $128\times128$ pixel \ac{ffhq} (blue) and StyleGAN (orange) images are shown on the left.
      The shaded area indicates a single standard deviation $\sigma$.
      We find higher mean coefficient values for the StyleGAN samples across the board. As the frequency increases from left to right, the differences become more pronounced.
      The plot on the right side shows validation and test accuracy for the identification experiment. Linear regression networks where used to identify \ac{ffhq} and StyleGAN. The blue line shows the pixel and the orange line the training accuracy using $\ln$-scaled absolute wavelet packet coefficients.
      Shaded areas indicate a single standard deviation for different initializations.
      We find that working with $\ln$-scaled absolute packets allowed linear separation of all three images sources. Furthermore, it significantly improves the convergence speed and final result.
      We found a mean test accuracy of $99.75 \pm 0.07 \% $ for the packet and for the pixel regression $83.06 \pm 2.5 \% $.}
  \label{fig:mean_packets_ffhq}
\end{figure}
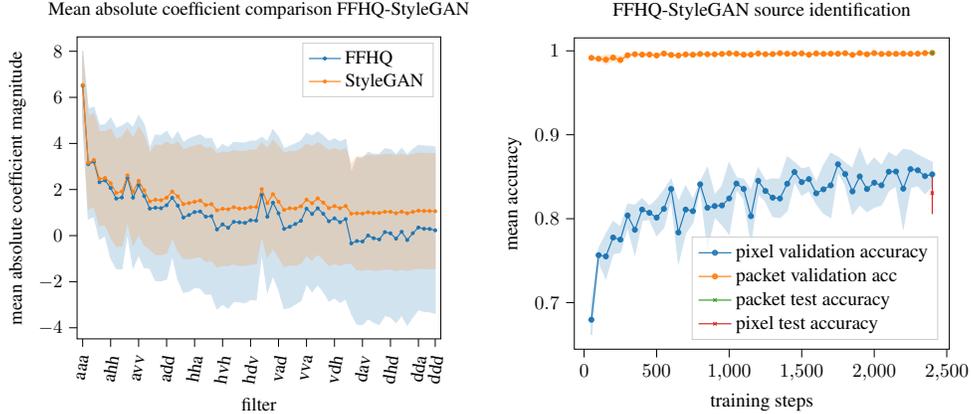
In Figure~\ref{fig:proof_of_concept} we saw significantly different mean wavelet-coefficients and standard deviations, shown in the rightmost column.
The disparity of the absolute mean packet difference widened as the frequency increased along the diagonal.
Additionally, background and edge coefficients appeared to diverge.
Exclusively for plotting purposes, we for now remove the spatial dimensions by averaging over these as well.
We observe in the left of Figure~\ref{fig:mean_packets_ffhq} increasing mean differences across the board. Differences are especially pronounced at the higher frequencies on the right. In comparison to the \ac{ffhq} standard deviation, the variance produced by the StyleGAN, shown in orange, is smaller for all coefficients.
In the following sections, we aim to exploit these differences for the identification of artificial images.
We will start with an interpretable linear network and will move on to highly performant nonlinear \ac{cnn}-architectures.

\subsection{Proof of concept: Linearly separating \ac{ffhq} and style-gan images}\label{sec:ffhq_linear}

Encouraged by the differences, we saw in Figures~\ref{fig:proof_of_concept} and \ref{fig:mean_packets_ffhq}, we attempt to linearly separate the $\ln$-scaled $3rd$ level Haar wavelet packets by training an interpretable linear regression model. We aim to obtain a classifier separating \ac{ffhq} wavelet packets from StyleGAN-packets. We work with 63k images per class for training, 2k for validation, and 5k for testing. All images have a $128\times128$ pixel resolution. The spatial dimensions are left intact.
Wavelet coefficients and raw pixels are normalized per color channel using mean $\mu$ and standard deviation $\sigma$. We subtract the training-set mean and divide using the standard deviation. On both normalized features, we train identical networks using the Adam optimizer~\citep{kingma2015adam} with a step size of 0.001 using PyTorch~\citep{paszke2017automatic}.

We plot the mean validation and test accuracy over 5 runs with identical seeds of $0,1,2,3,4$ in Figure~\ref{fig:mean_packets_ffhq} on the right. The shaded areas indicate a single $\sigma$-deviation. Networks initialized with the seeds $0,1,2,3,4$ have a mean accuracy of 99.75 $\pm$ 0.07 \%.
In the Haar packet coefficient space, we are able to separate the two sources linearly. Working in the Haar-wavelet space improved both the final result and convergence speed.

\begin{figure}
\centering
\begin{tikzpicture}

\begin{axis}[
colorbar,
colorbar style={ylabel={}},
colormap/viridis,
point meta max=0.2,
point meta min=-0.2,
tick align=outside,
tick pos=left,
title={StyleGAN},
x grid style={white!69.0196078431373!black},
xmin=-0.5, xmax=127.5,
xtick style={color=black},
y dir=reverse,
y grid style={white!69.0196078431373!black},
ymin=-0.5, ymax=127.5,
ytick style={color=black},
hide x axis,
hide y axis
]
\addplot graphics [includegraphics cmd=\pgfimage,xmin=-0.5, xmax=127.5, ymin=127.5, ymax=-0.5] {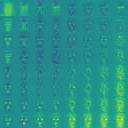};
\end{axis}

\end{tikzpicture}
\begin{tikzpicture}

\begin{axis}[
colorbar,
colorbar style={ylabel={}},
colormap/viridis,
point meta max=0.2,
point meta min=-0.2,
tick align=outside,
tick pos=left,
title={FFHQ},
x grid style={white!69.0196078431373!black},
xmin=-0.5, xmax=127.5,
xtick style={color=black},
y dir=reverse,
y grid style={white!69.0196078431373!black},
ymin=-0.5, ymax=127.5,
ytick style={color=black},
hide x axis,
hide y axis
]
\addplot graphics [includegraphics cmd=\pgfimage,xmin=-0.5, xmax=127.5, ymin=127.5, ymax=-0.5] {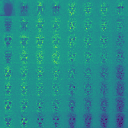};
\end{axis}

\end{tikzpicture}
\caption{Color-encoded matrix visualization of the learned classifier weights for the binary classification problem. We reshape the classifier-matrix into the $[2, 64, 16, 16, 3]$ packet form. Next we average over the three color channels and and arrange the $16$ by $16$ pixel packets in frequency order. Figure~\ref{fig:proof_of_concept} suggested high frequency packets would be crucial for the task of deepfake identification, this figure confirms our initial observation.
Input packet labels are available in Figure~\ref{fig:packetlabels}.}
\label{fig:classifier_plot}
\end{figure}
We visualize the weights for both classes for the classifier with seed $0$ in Figure~\ref{fig:classifier_plot}. The learned classifier confirms our observation from Figure~\ref{fig:proof_of_concept}. To identify deep-fakes the classifier does indeed rely heavily on the high-frequency packets. The two images are essentially inverses of each other. High frequency coefficients stimulate the StyleGAN neuron with a positive impact, while the impact is negative for the FFHQ neuron.

\subsection{Large scale detection of fully-fabricated-images}\label{sec:train_celeba_lsun}

In the previous experiment, two image sources, authentic \ac{ffhq} and StyleGAN images, had to be identified.
This section will study a standardized problem with more image sources.
To allow comparison with previous work, we exactly reproduce the experimental setup from \cite{frank2020frequency}, and \cite{yu2019attributing}. We choose this setup since we, in particular, want to compare to the spatial approach from \cite{yu2019attributing} and the frequency only DCT-representation presented in \cite{frank2020frequency}. We cite their results for comparison.
Additionally, we benchmark against the \ac{prnu} approach, as well as eigenfaces~\citep{sirovich1987low}.
The experiments in this section use four \acp{gan}: CramerGAN~\citep{Bellmare2017CramerGAN}, MMDGAN~\citep{Binkowski2018Demystifying}, ProGAN~\citep{karras2019pro}, and SN-DCGAN~\citep{Miyato2018Spectral}.

150k images were randomly selected from the \acf{celeba}~\citep{liu2015faceattributes} and \acf{lsun} bedroom~\citep{yu15lsun} data sets. Our pre-processing is identical for both.
The real images are cropped and resized to $128 \times 128$ pixels each.
With each of the four \acp{gan} an additional 150k images are generated at the same resolution. The 750k total images are split into 500k training, 100k validation, and 150k test images. To ensure stratified sampling, we draw the same amount of samples from each generator or the original data set.
As a result, the train, validation, and test sets contain equally many images from each source.

We compute wavelet packets with three levels for each image. We explore the use of Haar and Daubechies wavelets as well as symlets~\citep{daubechies1992tenwavelets,strang1996wavelets,jensen2001ripples,mallat2009wavelet}. The Haar wavelet is also the first Daubechies wavelet. Daubechies-wavelets and symlets are identical up to a degree of 3. Table~\ref{tab:source_seperation_results}, therefore, starts comparing both families above a degree of 4.

Both raw pixels and wavelet coefficients are normalized for a fair comparison. Normalization is always the last step before the models start their analysis.
We normalize by subtracting the training-set color-channel mean and dividing each color channel by its standard deviation. The \ac{ln}-scaled coefficients are normalized after the rescaling.
Given the original images as well as images generated by the four \acp{gan}, our classifiers must identify an image as either authentic or point out the generating \ac{gan} architecture. We train identical \ac{cnn}-classifiers on top of pixel and various wavelet representations. Additionally, we evaluate eigenface and PRNU baselines.
The wavelet packet transform, regression, and convolutional models are using PyTorch~\citep{Paszke2019Pytorch}.
Adam~\citep{kingma2015adam} optimizes our convolutional-models for 10 epochs.
For all experiments the batch size is set to 512 and the learning rate to 0.001.
\begin{table}[t]
  \centering
  \caption{The CNN architectures we used in our experiments. We show the convolution and pooling kernel, as well as matrix dimensions in brackets.
  For convolution layers we list the number of input and output filters as well as kernel height and width.
  As the packet representation generates additional channels we adapt the CNN architecture. We denote the number of classes as $c$, or equivalently
  the number of \acp{gan} in the problem plus one for the real data label. We set $d_i = 3$ when fusing pixel and packet representations. To additionally accommodate the Fourier representation we set $d_i = 6$. In the fusion case the input dimensions are the output channels of the previous layer plus the additional packet channels.
  } \label{tab:nets}
  \begin{tabular}{c c c c c c} \toprule
    \multicolumn{2}{c}{Wavelet-Packet} & \multicolumn{2}{c}{Fourier,Pixel} & \multicolumn{2}{c}{Fusion}     \\  \cmidrule(lr){1-2}\cmidrule(lr){3-4}\cmidrule(lr){5-6}
    Conv        & (192,24,3,3)         & Conv        & (3,8,3,3)           & Conv        & ($d_i$,8,3,3)    \\
    ReLU        &                      & ReLU        &                     & ReLU        &                  \\
    Conv        & (24,24,6,6)          & Conv        & (8,8,3)             & AvgPool     & (2,2)            \\
    ReLU        &                      & ReLU        &                     & Conv        & (20,16,3,3)      \\
    Conv        & (24,24,9,9)          & AvgPool     & (2,2)               & ReLU        &                  \\
    ReLU        &                      & Conv        & (8,16,3,3)          & AvgPool     & (2,2)            \\
                &                      & ReLU        &                     & Conv        & (64,32,3,3)      \\
                &                      & AvgPool     & (2,2)               & ReLU        &                  \\
                &                      & Conv        & (16,32,3,3)         & AvgPool     & (2,2)            \\
                &                      & ReLU        &                     & Conv        & (224,32,3,3)     \\
                &                      &             &                     & ReLU        &                  \\  \cmidrule(lr){1-2}\cmidrule(lr){3-4}\cmidrule(lr){5-6}
    Dense       & (24,  $c$)           & Dense       & ($32\cdot 28 \cdot 28$, $c$)     & Dense       & ($32\cdot 16\cdot 16$, $c$)  \\ \bottomrule
  \end{tabular}
\end{table}
Table~\ref{tab:nets} lists the exact network architectures trained. The Fourier and pixel architectures use the convolutional network architecture described in \cite{frank2020frequency}. Since the Fourier transform does not change the input dimensions, no architectural changes are required. Consequently, our Fourier and pixel input experiments share the same architecture.
The wavelet packet transform, as described in section~\ref{subsec:wavelet-packets}, employs sub-sampling operations and multiple filters. The filtered features stack up along the channel dimension. Hence the channel number increases with every level. At the same time, analyzing additional scales cuts the spatial dimensions in half every time. The network in the leftmost column of Table~\ref{tab:nets} addresses the changes in dimensionality. Its first layer has enough input filters to process all incoming packets. It does not use the fundamental similarities connecting wavelet packets and convolutional neural networks.
The fusion architecture on the very right of Table~\ref{tab:nets} does. Its design allows it to process multiple packet representations alongside its own internal \ac{cnn}-features. Using an average pooling operation per packet layer leads to \ac{cnn} features and wavelet packet coefficients with the same spatial dimensions. Both are concatenated along the channel dimension and processed further in the next layer. The concatenation of the wavelet packets requires additional input filters in each convolution layer. We use the number of output filters from the previous layer plus the packet channels.

\begin{table}[t]
	\centering
	\caption{%
	\ac{cnn} source identification results on the \ac{celeba} and \ac{lsun} bedroom data sets.
  We explore the use of boundary-wavelet packets up to a wavelet degree of 5.
  We report the test set accuracy. To add additional context, we report mean test set accuracy and standard deviation over 5 runs for all deep learning experiments. \ac{ln} denotes the natural logarithm. Logarithmically scaled \ac{fft2} coefficients as well as Daubechies (db) and symlets (sym) are compared.
	}\label{tab:source_seperation_results}
	\begin{tabular}{lrrrrr}\toprule
	                      &           & \multicolumn{4}{c}{Accuracy[\%]} \\
	                      &           & \multicolumn{2}{c}{\acs{celeba}}      & \multicolumn{2}{c}{\acs{lsun} bedroom}\\\cmidrule(lr){3-4}\cmidrule(lr){5-6}
	Method                &    params & $\max$         & $\mu\pm\sigma$       & $\max$     & $\mu\pm\sigma$    \\ \midrule
	Eigenfaces (ours)     &     0     & 68.56          &  -               & 62.24          &  -                \\
	Eigenfaces-DCT (\citeauthor{frank2020frequency})& 0 & 88.39  &  -               & \textit{94.31} &  -                \\
	Eigenfaces-$\ln$-Haar (ours) & 0 & \textit{92.67} &  -               & 87.91          &  -                \\ \midrule
	\acs{prnu} (ours)    &         0& 83.13          &  -               & 66.10          &  -                \\ \midrule
	CNN-Pixel (ours)     &         132k   & 98.87          & 98.74$\pm$0.14 & 97.06          & 95.02$\pm$1.14  \\
  CNN-$\ln$-fft2 (ours)&         132k   & 99.55          & 99.23$\pm$0.24 & 99.61          & 99.53$\pm$0.07  \\
	CNN-$\ln$-Haar (ours)&         109k   & 97.09          & 96.79$\pm$0.29 & 97.14          & 96.89$\pm$0.19  \\
  CNN-$\ln$-db2  (ours)&         109k   & 99.20          & 98.50$\pm$0.96 & 98.90          & 98.35$\pm$0.70  \\
  CNN-$\ln$-db3  (ours)&         109k   & 99.38          & 99.11$\pm$0.49 & 99.19          & 99.01$\pm$0.17  \\
  CNN-$\ln$-db4  (ours)&         109k   & 99.43          & 99.27$\pm$0.15 & 99.02          & 98.46$\pm$0.67  \\
  CNN-$\ln$-db5  (ours)&         109k   & 99.02          & 98.66$\pm$0.59 & 98.32          & 97.64$\pm$0.83  \\
  CNN-$\ln$-sym4 (ours)&         109k   & 99.43          & 98.98$\pm$0.36 & 99.09          & 98.57$\pm$0.72  \\
  CNN-$\ln$-sym5 (ours)&         109k   & 99.49          & 98.79$\pm$0.48 & 99.07          & 98.78$\pm$0.24  \\
  CNN-$\ln$-db4-fuse (ours)     & 127k  & 99.73          & 99.50$\pm$0.25 & 99.61          & 99.21$\pm$0.41  \\
  CNN-$\ln$-db4-fuse-fft2 (ours)& 127k  & \textbf{99.75} & 99.41$\pm$0.33 & \textbf{99.71} & 98.41$\pm$1.78  \\
  \midrule
	CNN-Pixel (\citeauthor{yu2019attributing})& - & 99.43          &      -          & 98.58          &  -                \\
	CNN-Pixel (\citeauthor{frank2020frequency})& 170k & 97.80          &      -          & 98.95          &  -                \\
	CNN-DCT (\citeauthor{frank2020frequency})& 170k & 99.07          &      -          & 99.64          &  -                \\\bottomrule
\end{tabular}
\end{table}

Table~\ref{tab:source_seperation_results} lists the classification accuracies of the previously described networks and benchmark-methods with various features for comparison.
We will first study the effect of third-level wavelet packets in comparison to pixel representation.
On CelebA the eigenface approach, in particular, was able to classify \textit{24.11\%} more images correctly when run on \ac{ln}-scaled Haar-Wavelet packages instead of pixels.
For the more complex \ac{cnn}, Haar wavelets are not enough. More complex wavelets, however, significantly improve the accuracy.
We find accuracy maxima superior to the DCT approach from \cite{frank2020frequency} for five wavelets, using a comparable \ac{cnn}. The db3 and db4 packets are better on average, demonstrating the robustness of the wavelet approach.
We choose the db4-wavelet representation for the feature fusion experiments.
Fusing wavelet packets into a pixel-CNN improves the result while reducing the total number of trainable parameters.
On both \ac{celeba} and \ac{lsun} the Fourier representation emerged as a strong contender. However, we argue that Fourier and wavelet features are compatible and can be used jointly. On both \ac{celeba} and \ac{lsun} the best performing networks fused Fourier-pixel as well as the first three wavelet packet scales.

\begin{table}[t]
	\centering
	\caption{%
	Confusion matrix for our CNN using db4-wavelet-packets on CelebA.
  Classification errors for the original data set as well as
  the CramerGAN, MMDGAN, ProGAN, and SN-DCGAN architectures are shown. The test set contains 30,000 entries per label. The detector classifies 99.43\% of all images correctly. }
	\label{tab:confusion_matrix}
	\begin{tabular}{lrrrrr}\toprule
	           & \multicolumn{5}{c}{Predicted label}\\\cmidrule(lr){2-6}
    True label & CelebA & CramerGAN & MMDGAN & ProGAN & SN-DCGAN \\ \midrule
    CelebA     & 29,759 &   7   & 11   &107   &116 \\
    CramerGAN  &  19    & 29,834&  144 &    3 &    0 \\
    MMDGAN     &  17    & 120   &29,862 &    1 &    0\\
    ProGAN     &  136   &  1    & 0    &29,755 &  108\\
    SN-DCGAN   &  55    &  0    & 0    &  5   & 29,940\\ \bottomrule
	\end{tabular}
\end{table}
We show the confusion matrix for our best performing network on CelebA in Table~\ref{tab:confusion_matrix}.
Among the \acp{gan} we considered the ProGAN-architecture and SN-DCGAN-architecture.
Images drawn from both architectures were almost exclusively misclassified as original data and rarely attributed to another GAN. The CramerGAN and MMDGAN generated images were often misattributed to each other, but rarely confused with the original data set. Of all misclassified real CelebA images most were confused with ProGAN images, making it the most advanced network of the four. Overall we find our, in comparison to a GAN, lightweight 109k classifier able to accurately spot 99.45\% of all fakes in this case.
Note that our convolutional networks have only 109k or 127k parameters, respectively, while \cite{yu2019attributing} used roughly 9 million parameters and \cite{frank2020frequency} utilized around 170,000 parameters.
We also observed that using our packet representation improved convergence speed.

\begin{figure}
  \centering
\begin{tikzpicture}

\definecolor{color0}{rgb}{0.12156862745098,0.466666666666667,0.705882352941177}
\definecolor{color1}{rgb}{1,0.498039215686275,0.0549019607843137}
\definecolor{color2}{rgb}{0.172549019607843,0.627450980392157,0.172549019607843}
\definecolor{color3}{rgb}{0.83921568627451,0.152941176470588,0.156862745098039}
\definecolor{color4}{rgb}{0.580392156862745,0.403921568627451,0.741176470588235}

\begin{axis}[
legend cell align={left},
legend style={fill opacity=0.8, draw opacity=1, text opacity=1, draw=white!80!black},
tick align=outside,
tick pos=left,
title={Mean absolute coefficient comparison real data-GAN},
x grid style={white!69.0196078431373!black},
xlabel={filter},
xmin=-1, xmax=64,
xtick style={color=black},
xticklabel style={rotate=80.0},
xtick=      {  0,  5, 10, 15, 20, 25, 30, 35, 40, 45, 50, 55, 60, 63},
xticklabels={aaa,ahh,avv,add,hha,hvh,hdv,vad,vva,vdh,dav,dhd,dda,ddd},
y grid style={white!69.0196078431373!black},
ylabel={mean absolute coefficient magnitude},
ymin=-2.19718509912491, ymax=8.30703383684158,
ytick style={color=black}
]

\addplot [semithick, color0, mark=*, mark size=0.75, mark options={solid}]
table {%
0 6.01465940475464
1 3.4729540348053
2 3.67855954170227
3 2.8280131816864
4 2.00567269325256
5 2.59860634803772
6 1.73844087123871
7 2.17423605918884
8 2.41226053237915
9 1.89528751373291
10 2.92899966239929
11 2.3455445766449
12 1.2608277797699
13 1.49220836162567
14 1.52865087985992
15 1.82956397533417
16 0.881014049053192
17 1.22504341602325
18 0.800334215164185
19 0.950005412101746
20 1.37277936935425
21 1.66108429431915
22 1.30123925209045
23 1.55118441581726
24 0.597272336483002
25 0.660310864448547
26 0.720736920833588
27 0.817619025707245
28 0.93329906463623
29 0.954048216342926
30 1.08230495452881
31 1.11716771125793
32 0.959399878978729
33 0.85349828004837
34 1.30958807468414
35 1.14957702159882
36 0.5917729139328
37 0.707824647426605
38 0.763651192188263
39 0.933478236198425
40 1.62828385829926
41 1.53047406673431
42 1.78383874893188
43 1.60926139354706
44 1.02934789657593
45 1.2626668214798
46 1.04161095619202
47 1.26895391941071
48 0.211168929934502
49 0.246902659535408
50 0.290212869644165
51 0.34762892127037
52 0.423273354768753
53 0.3807293176651
54 0.53387725353241
55 0.499570161104202
56 0.474697589874268
57 0.541726589202881
58 0.399879157543182
59 0.474222719669342
60 0.80670839548111
61 0.74063777923584
62 0.70450234413147
63 0.664617359638214
};
\addlegendentry{CelebA}
\addplot [semithick, color1, mark=*, mark size=0.75, mark options={solid}]
table {%
0 6.80913352966309
1 3.33471512794495
2 3.63676953315735
3 2.47717237472534
4 1.81762850284576
5 2.31651258468628
6 1.36532485485077
7 1.70523607730865
8 2.41854500770569
9 1.52738666534424
10 2.83055114746094
11 1.90421164035797
12 0.886915922164917
13 0.996964693069458
14 1.1144608259201
15 1.2973974943161
16 0.609946489334106
17 0.813819050788879
18 0.266595184803009
19 0.311121046543121
20 0.933640420436859
21 1.28839862346649
22 0.779020488262177
23 1.1109961271286
24 0.0407417081296444
25 0.0612712167203426
26 0.152032658457756
27 0.216510534286499
28 0.357982903718948
29 0.430365949869156
30 0.487430363893509
31 0.585397183895111
32 0.639637887477875
33 0.316718488931656
34 0.867562711238861
35 0.640069007873535
36 0.0468290708959103
37 0.148428902029991
38 0.269166171550751
39 0.403715878725052
40 1.16378486156464
41 1.01250839233398
42 1.40932071208954
43 1.19925630092621
44 0.501046121120453
45 0.72066992521286
46 0.601619839668274
47 0.801980137825012
48 -0.0616541989147663
49 -0.292156159877777
50 -0.229322731494904
51 -0.214497089385986
52 -0.198910161852837
53 -0.200290232896805
54 -0.0996512696146965
55 -0.0856761708855629
56 -0.134660810232162
57 -0.0778527706861496
58 -0.160370364785194
59 -0.0947199538350105
60 0.211288765072823
61 0.138470143079758
62 0.105216011404991
63 0.0893872231245041
};
\addlegendentry{CramerGan}
\addplot [semithick, color2, mark=o, mark size=0.75, mark options={solid}]
table {%
0 6.8083324432373
1 3.33357644081116
2 3.63026213645935
3 2.47006177902222
4 1.80782163143158
5 2.31904006004333
6 1.35608196258545
7 1.7050724029541
8 2.40143251419067
9 1.51048612594604
10 2.81892490386963
11 1.90023624897003
12 0.853743135929108
13 0.984867632389069
14 1.0998523235321
15 1.30350172519684
16 0.541523396968842
17 0.845356643199921
18 0.227124348282814
19 0.327255874872208
20 0.914188146591187
21 1.30737435817719
22 0.756661534309387
23 1.12339317798615
24 0.0614185109734535
25 0.0884735658764839
26 0.141028389334679
27 0.226770102977753
28 0.344785690307617
29 0.444705665111542
30 0.477160006761551
31 0.592826068401337
32 0.585794985294342
33 0.29420593380928
34 0.871981501579285
35 0.646479368209839
36 0.0212220344692469
37 0.138315111398697
38 0.241927936673164
39 0.39843362569809
40 1.14002859592438
41 0.99515300989151
42 1.40871429443359
43 1.20123946666718
44 0.476837396621704
45 0.703849136829376
46 0.583839654922485
47 0.796781539916992
48 -0.149808138608932
49 -0.335976898670197
50 -0.297594159841537
51 -0.273580104112625
52 -0.187531769275665
53 -0.200405910611153
54 -0.0922312960028648
55 -0.108378909528255
56 -0.182834789156914
57 -0.0930851399898529
58 -0.205528199672699
59 -0.131533578038216
60 0.187206506729126
61 0.143775254487991
62 0.0999632030725479
63 0.0819371342658997
};
\addlegendentry{MMDGAN}
\addplot [semithick, color3, mark=*, mark size=0.75, mark options={solid}]
table {%
0 6.02881288528442
1 3.4786205291748
2 3.67460489273071
3 2.8435173034668
4 2.01852631568909
5 2.6028163433075
6 1.75253772735596
7 2.18552374839783
8 2.41219592094421
9 1.92786026000977
10 2.92645406723022
11 2.36710357666016
12 1.3044718503952
13 1.52551567554474
14 1.55331218242645
15 1.84256994724274
16 0.933612823486328
17 1.18086278438568
18 0.798911511898041
19 0.96875935792923
20 1.36585772037506
21 1.63519251346588
22 1.31791806221008
23 1.58225786685944
24 0.550029397010803
25 0.641150116920471
26 0.65531063079834
27 0.799324095249176
28 0.984898209571838
29 0.985758125782013
30 1.11166191101074
31 1.14288604259491
32 1.00568294525146
33 0.791610717773438
34 1.31583380699158
35 1.15171790122986
36 0.707819402217865
37 0.726797580718994
38 0.826854646205902
39 0.972508668899536
40 1.63479268550873
41 1.57438588142395
42 1.78339874744415
43 1.63879323005676
44 1.10263669490814
45 1.31229305267334
46 1.09279179573059
47 1.3015798330307
48 0.144446328282356
49 0.147445827722549
50 0.283898264169693
51 0.375030785799026
52 0.283521115779877
53 0.263955891132355
54 0.471172243356705
55 0.457052648067474
56 0.454510480165482
57 0.555726826190948
58 0.416764974594116
59 0.510142743587494
60 0.890246450901031
61 0.786732912063599
62 0.750423014163971
63 0.699592590332031
};
\addlegendentry{ProGAN}
\addplot [semithick, color4, mark=o, mark size=0.75, mark options={solid}]
table {%
0 5.99483728408813
1 3.44801807403564
2 3.65120625495911
3 2.80616188049316
4 1.99145984649658
5 2.58644700050354
6 1.72471189498901
7 2.16618490219116
8 2.38950204849243
9 1.89813208580017
10 2.90309047698975
11 2.33473205566406
12 1.26406538486481
13 1.48536765575409
14 1.53083574771881
15 1.82029223442078
16 0.921741485595703
17 1.22612988948822
18 0.707363605499268
19 0.913736760616302
20 1.3533661365509
21 1.65644323825836
22 1.27408969402313
23 1.53958034515381
24 0.45324245095253
25 0.619458615779877
26 0.678491532802582
27 0.841003358364105
28 0.906218111515045
29 0.929774284362793
30 1.10684871673584
31 1.14728629589081
32 1.03967928886414
33 0.72719794511795
34 1.3507753610611
35 1.12657082080841
36 0.444608807563782
37 0.653249144554138
38 0.735603392124176
39 0.941015720367432
40 1.6077173948288
41 1.5220388174057
42 1.7725533246994
43 1.61597514152527
44 1.03808224201202
45 1.28270184993744
46 1.03987383842468
47 1.30080831050873
48 0.232410594820976
49 0.25051349401474
50 0.355630159378052
51 0.481660038232803
52 0.536981105804443
53 0.431418657302856
54 0.694437682628632
55 0.626148998737335
56 0.505324304103851
57 0.681760132312775
58 0.448993116617203
59 0.618861794471741
60 0.806342780590057
61 0.757990837097168
62 0.804378509521484
63 0.776825428009033
};
\addlegendentry{SNGAN}
\end{axis}
\end{tikzpicture}
\begin{tikzpicture}

\begin{axis}[
colorbar,
colorbar style={ylabel={}},
colormap/viridis,
point meta max=0.413612049558761,
point meta min=0.00129407558160742,
tick align=outside,
tick pos=left,
x grid style={white!69.0196078431373!black},
xmin=-0.5, xmax=127.5,
xtick style={color=black},
y dir=reverse,
y grid style={white!69.0196078431373!black},
ymin=-0.5, ymax=127.5,
ytick style={color=black},
title={Saliency Analysis CelebA},
xlabel={$\;$}
]
\addplot graphics [includegraphics cmd=\pgfimage,xmin=-0.5, xmax=127.5, ymin=127.5, ymax=-0.5] {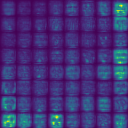};
\end{axis}

\end{tikzpicture}
\caption{The left plots depicts the mean \ac{ln}-db4-wavelet packet plots for CelebA, CramerGAN, MMDGAN, ProGAN and SN-DCGAN, where CelebA is mostly hidden behind the curves for ProGAN and SN-DCGAN. CramerGAN and MMDGAN show similar mean, which relates to their more often pairwise misattribution.
        The right presents the mean saliency map over all labels for the CNN-$\ln$-db4 trained with seed 0.
        }\label{fig:packets_sali_celebA}
\end{figure}
In Figure \ref{fig:packets_sali_celebA} (left) we show mean ln-db4-wavelet packet plots, where two \acp{gan} show similar behavior to CelebA, while two are different.
For interpretation, we use saliency maps~\citep{SimonyanVZ13} as an exemplary approach. We find that the classifier relies on the edges of the spectrum, Figure \ref{fig:packets_sali_celebA} (right).

\subsubsection{Training with a fifth of the original data}\label{sec:data_reduced}
We study the effect of a drastic reduction of the training data size, by reducing the training set size by 80\%.
\begin{table}[t]
\centering
\caption{%
	\ac{cnn} source identification results on the \ac{celeba} data set with only 20\,\% of the original data.
	We report the test set accuracy mean and standard deviation over 5 runs as well as the accuracy loss compared to the corresponding network trained on the full data set.}
\label{tab:reduced_data}
\begin{tabular}{lrrrr}\toprule
	& \multicolumn{4}{c}{Accuracy on \acs{celeba}[\%]}\\ \cmidrule(lr){2-5}
	Method & $\max$ & Loss & $\mu \pm \sigma$ & Loss\\ \midrule
	CNN-Pixel (ours)                     & 96.37 & -2.50 & 95.16 $\pm$ 0.86 & -3.58\\
	CNN-$\ln$-db4 (ours)              & \textbf{99.01} & -0.41 & 96.96 $\pm$ 3.47 & -2.58 \\ \midrule
	CNN-Pixel~\citep{frank2020frequency} & 96.33 & -1.47 &       -          & - \\
	CNN-DCT~\citep{frank2020frequency}   & 98.47 & -0.60 &       -          & - \\ \bottomrule
\end{tabular}
\end{table}
Table~\ref{tab:reduced_data} reports test-accuracies in this case. The classifiers are retrained as described in section~\ref{sec:train_celeba_lsun}. We find that our Daubechies 4 packet-CNN classifiers are robust to training data reductions, in particular in comparison to the pixel representations. We find that wavelet representations share this property with the Fourier-inputs proposed by \cite{frank2020frequency}.
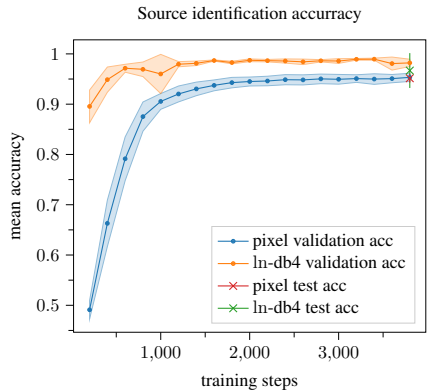
\begin{figure}
	\centering
\begin{tikzpicture}

\definecolor{color0}{rgb}{0.12156862745098,0.466666666666667,0.705882352941177}
\definecolor{color1}{rgb}{1,0.498039215686275,0.0549019607843137}
\definecolor{color2}{rgb}{0.172549019607843,0.627450980392157,0.172549019607843}
\definecolor{color3}{rgb}{0.83921568627451,0.152941176470588,0.156862745098039}

\begin{axis}[
legend cell align={left},
legend style={
  fill opacity=0.8,
  draw opacity=1,
  text opacity=1,
  at={(0.97,0.03)},
  anchor=south east,
  draw=white!80!black
},
tick align=outside,
tick pos=left,
title={Source identification accurracy},
x grid style={white!69.0196078431373!black},
xlabel={training steps},
xmin=20, xmax=3980,
xtick style={color=black},
minor tick num=1,
y grid style={white!69.0196078431373!black},
ylabel={mean accuracy},
ymin=0.446388465760411, ymax=1.02794984864199,
ytick style={color=black}
]
\path [draw=color0, fill=color0, fill opacity=0.2, draw opacity=0.5]
(axis cs:200,0.50934092592679)
--(axis cs:200,0.47282307407321)
--(axis cs:400,0.616843513074324)
--(axis cs:600,0.747890299706753)
--(axis cs:800,0.846173242354882)
--(axis cs:1000,0.889782318544423)
--(axis cs:1200,0.904492780439194)
--(axis cs:1400,0.916871993485344)
--(axis cs:1600,0.926604186969408)
--(axis cs:1800,0.932470091958852)
--(axis cs:2000,0.936207084001301)
--(axis cs:2200,0.936612184484189)
--(axis cs:2400,0.938625440882161)
--(axis cs:2600,0.938264030315163)
--(axis cs:2800,0.940775551416346)
--(axis cs:3000,0.939789866406042)
--(axis cs:3200,0.943335261089885)
--(axis cs:3400,0.939410277295402)
--(axis cs:3600,0.942581362378804)
--(axis cs:3800,0.945205657986045)
--(axis cs:3800,0.961474342013956)
--(axis cs:3800,0.961474342013956)
--(axis cs:3600,0.959030637621196)
--(axis cs:3400,0.960441722704598)
--(axis cs:3200,0.958444738910115)
--(axis cs:3000,0.959162133593958)
--(axis cs:2800,0.959840448583654)
--(axis cs:2600,0.958319969684837)
--(axis cs:2400,0.958750559117839)
--(axis cs:2200,0.955583815515811)
--(axis cs:2000,0.953932915998699)
--(axis cs:1800,0.953261908041148)
--(axis cs:1600,0.948211813030593)
--(axis cs:1400,0.943888006514657)
--(axis cs:1200,0.936115219560805)
--(axis cs:1000,0.921149681455577)
--(axis cs:800,0.904306757645118)
--(axis cs:600,0.834881700293247)
--(axis cs:400,0.709104486925676)
--(axis cs:200,0.50934092592679)
--cycle;

\path [draw=color1, fill=color1, fill opacity=0.2, draw opacity=0.5]
(axis cs:200,0.928038138917657)
--(axis cs:200,0.863049861082343)
--(axis cs:400,0.923703856778718)
--(axis cs:600,0.96361076279221)
--(axis cs:800,0.954901873407757)
--(axis cs:1000,0.921118918696663)
--(axis cs:1200,0.974338185044239)
--(axis cs:1400,0.97679476877981)
--(axis cs:1600,0.98567433221607)
--(axis cs:1800,0.980200714384236)
--(axis cs:2000,0.984545644190499)
--(axis cs:2200,0.984141881866463)
--(axis cs:2400,0.98172103402775)
--(axis cs:2600,0.978745163613819)
--(axis cs:2800,0.984049448648934)
--(axis cs:3000,0.981132937860143)
--(axis cs:3200,0.98762701518916)
--(axis cs:3400,0.987491397309641)
--(axis cs:3600,0.967752165794587)
--(axis cs:3800,0.974898286154402)
--(axis cs:3800,0.989365713845598)
--(axis cs:3800,0.989365713845598)
--(axis cs:3600,0.993711834205413)
--(axis cs:3400,0.991520602690358)
--(axis cs:3200,0.99085298481084)
--(axis cs:3000,0.990515062139857)
--(axis cs:2800,0.988562551351067)
--(axis cs:2600,0.989654836386181)
--(axis cs:2400,0.98997896597225)
--(axis cs:2200,0.988950118133537)
--(axis cs:2000,0.989998355809501)
--(axis cs:1800,0.985379285615763)
--(axis cs:1600,0.98796166778393)
--(axis cs:1400,0.98571723122019)
--(axis cs:1200,0.984605814955761)
--(axis cs:1000,0.999037081303337)
--(axis cs:800,0.983782126592243)
--(axis cs:600,0.97928923720779)
--(axis cs:400,0.973864143221282)
--(axis cs:200,0.928038138917657)
--cycle;

\addplot [semithick, color0, mark=*, mark size=1, mark options={solid}]
table {%
200 0.491082
400 0.662974
600 0.791386
800 0.87524
1000 0.905466
1200 0.920304
1400 0.93038
1600 0.937408
1800 0.942866
2000 0.94507
2200 0.946098
2400 0.948688
2600 0.948292
2800 0.950308
3000 0.949476
3200 0.95089
3400 0.949926
3600 0.950806
3800 0.95334
};
\addlegendentry{pixel validation acc}
\addplot [semithick, color1, mark=*, mark size=1, mark options={solid}]
table {%
200 0.895544
400 0.948784
600 0.97145
800 0.969342
1000 0.960078
1200 0.979472
1400 0.981256
1600 0.986818
1800 0.98279
2000 0.987272
2200 0.986546
2400 0.98585
2600 0.9842
2800 0.986306
3000 0.985824
3200 0.98924
3400 0.989506
3600 0.980732
3800 0.982132
};
\addlegendentry{$\ln$-db4 validation acc}

\addplot [semithick, color3, mark=x, mark size=3, mark options={solid}]
table {%
3800 0.951557333333333
};

\addlegendentry{pixel test acc}

\path [draw=color2, semithick]
(axis cs:3800,0.932199426337478)
--(axis cs:3800,1.00151524032919);

\path [draw=color3, semithick]
(axis cs:3800,0.942914112123108)
--(axis cs:3800,0.960200554543559);

\addplot [semithick, color2, mark=x, mark size=3, mark options={solid}]
table {%
3800 0.966857333333333
};

\addlegendentry{$\ln$-db4 test acc}
\end{axis}

\end{tikzpicture}
	\caption{Mean validation and test set accuracy of 5 runs of source identification on \acs{celeba} for a CNN trained on the raw images and $\ln$-db4 wavelet packets, each using only 20\,\% of the original training data.
	The shaded areas indicate a single standard deviation $\sigma$.
	}
	\label{fig:celeba_accuracy_reduced}
\end{figure}
Figure~\ref{fig:celeba_accuracy_reduced} plots validation accuracy during training as well as the test accuracy at the end. The wavelet packet representation produces a significant initial boost, which persists throughout the training process. This observation is in-line with the training behaviour of the linear regressor as shown in figure~\ref{fig:mean_packets_ffhq} on the right.
In addition to reducing the training set sizes we discuss removing a \ac{gan} entirely in the supplementary section~\ref{sec:unkown_gan}.

\subsection{Detecting additional \acp{gan} on \ac{ffhq}}\label{sec:big_ffhq}
In this section, we consider a more complex setup on \ac{ffhq} taking into account what we learned in the previous section. We extend the setup we studied in section~\ref{sec:ffhq_linear} by adding StyleGAN2 \citep{karras2020analyzing} and StyleGAN3~\citep{Karras2021Alias} generated images. Unlike section~\ref{sec:train_celeba_lsun}, this setup has not been explored in previous work. We consider it here to work with some of the most recent architectures. \ac{ffhq}-pre-trained networks are available for all three \acp{gan}. Here, we re-evaluate the most important models from section~\ref{sec:train_celeba_lsun} in the \ac{ffhq} setting.
\begin{table}[tbp]
\centering
\caption{Classification accuracy for the \ac{ffhq} , StyleGAN~\citep{karras2019pro}, StyleGAN2~\citep{karras2020analyzing} and StyleGAN3~\citep{Karras2021Alias} separation problem. This problem has four classes, which changes the shape of the final classifier. The different classifiers cause variations in the total number of weights in comparison to Table~\ref{tab:source_seperation_results}.}
\begin{tabular}{l r r r} \toprule
                & &\multicolumn{2}{c}{Accuracy on \ac{ffhq} [\%]} \\ \cmidrule(lr){3-4}
   Method                    & parameters  &  max  & $\mu \pm \sigma$ \\ \midrule
   Regression-pixel          & 197k       &72.42& $70.55\pm 1.19$  \\
   Regression-\ac{ln}-db4    & 197k       &95.60& $95.00\pm 0.52$  \\
   CNN-pixel                 & 107k       &93.71& $90.90\pm 2.19$  \\
   CNN-\ac{ln}-fft2          & 107k       &86.03& $85.52\pm 0.50$  \\
   CNN-\ac{ln}-db4           & 109k       &96.28& $95.85\pm 0.59$  \\
   CNN-\ac{ln}-db4-fuse      & 119k       &\textbf{97.52}& $96.91\pm 0.45$  \\
   CNN-\ac{ln}-db4-fft2-fuse & 119k       &97.49& $96.67\pm 0.68$  \\ \bottomrule
   \label{tab:ffhq_hard}
\end{tabular}
\end{table}
We re-use the training hyperparameters as described in section~\ref{sec:train_celeba_lsun}. Results appear in Table~\ref{tab:ffhq_hard}, in comparison to Table~\ref{tab:source_seperation_results} we observe a surprising performance drop for the Fourier features. For the Daubechies-four wavelet-packets we observe consistent performance gains, both in the regression and convolutional setting. The fusing pixels and three wavelet packet levels performed best. Fusing Fourier, as well as wavelet packet features, does not help in this case.

\begin{figure}
  \centering
\begin{tikzpicture}

\definecolor{color0}{rgb}{0.12156862745098,0.466666666666667,0.705882352941177}
\definecolor{color1}{rgb}{1,0.498039215686275,0.0549019607843137}
\definecolor{color2}{rgb}{0.172549019607843,0.627450980392157,0.172549019607843}
\definecolor{color3}{rgb}{0.83921568627451,0.152941176470588,0.156862745098039}

\begin{axis}[
legend cell align={left},
legend style={fill opacity=0.8, draw opacity=1, text opacity=1, draw=white!80!black},
tick align=outside,
tick pos=left,
title={Mean absolute coefficient comparison real data-StyleGANs},
x grid style={white!69.0196078431373!black},
xlabel={filter},
xmin=-1, xmax=64,
xtick style={color=black},
xticklabel style={rotate=80.0},
xtick=      {  0,  5, 10, 15, 20, 25, 30, 35, 40, 45, 50, 55, 60, 63},
xticklabels={aaa,ahh,avv,add,hha,hvh,hdv,vad,vva,vdh,dav,dhd,dda,ddd},
y grid style={white!69.0196078431373!black},
ylabel={mean absolute coefficient magnitude},
ymin=-2.16403103470802, ymax=7.93413675427437,
ytick style={color=black},
ytick={-4,-2,0,2,4,6,8},
yticklabels={\ensuremath{-}4,\ensuremath{-}2,0,2,4,6,8},
xlabel={$\;$}
]

\addplot [semithick, color0, mark=*, mark size=1.25, mark options={solid}]
table {%
0 6.07450008392334
1 3.60966300964355
2 3.83470296859741
3 2.9805862903595
4 2.13896322250366
5 2.72692203521729
6 1.881502866745
7 2.30792117118835
8 2.55471277236938
9 1.98593747615814
10 3.07677984237671
11 2.454021692276
12 1.37373113632202
13 1.6012464761734
14 1.65278029441833
15 1.95859229564667
16 0.592816472053528
17 1.11352455615997
18 0.456450521945953
19 0.812050223350525
20 1.41313624382019
21 1.66937983036041
22 1.36090612411499
23 1.57141268253326
24 0.301038026809692
25 0.542889893054962
26 0.41833907365799
27 0.694874286651611
28 1.01079416275024
29 0.988956153392792
30 1.14404106140137
31 1.13349056243896
32 0.660061955451965
33 0.59458464384079
34 1.16996312141418
35 1.08082711696625
36 0.39011886715889
37 0.509587109088898
38 0.704161047935486
39 0.89174622297287
40 1.59191060066223
41 1.55709731578827
42 1.75720798969269
43 1.65064740180969
44 1.08896231651306
45 1.33885431289673
46 1.10620093345642
47 1.34743225574493
48 -0.303613543510437
49 -0.109532833099365
50 -0.144318372011185
51 0.10481122136116
52 0.23806220293045
53 0.168030440807343
54 0.464929729700089
55 0.408054947853088
56 0.148717612028122
57 0.421602576971054
58 0.0711986795067787
59 0.341338068246841
60 0.874322056770325
61 0.771118521690369
62 0.762959480285645
63 0.684643447399139
};
\addlegendentry{ffhq}
\addplot [semithick, color1, mark=*, mark size=1.25, mark options={solid}]
table {%
0 6.09244346618652
1 3.62697410583496
2 3.87641000747681
3 3.0107147693634
4 2.26289033889771
5 2.79053664207458
6 2.023108959198
7 2.37482571601868
8 2.65114450454712
9 2.10129189491272
10 3.14889907836914
11 2.52957487106323
12 1.62862300872803
13 1.77581095695496
14 1.86044251918793
15 2.07552003860474
16 1.34974122047424
17 1.62999963760376
18 1.30339896678925
19 1.40461981296539
20 1.70544707775116
21 2.00339341163635
22 1.63663911819458
23 1.89324820041656
24 1.21282553672791
25 1.24001789093018
26 1.26427114009857
27 1.32124412059784
28 1.38686370849609
29 1.42303347587585
30 1.48407375812531
31 1.52595973014832
32 1.41463875770569
33 1.37698554992676
34 1.70935547351837
35 1.60174143314362
36 1.23448526859283
37 1.3105480670929
38 1.36142218112946
39 1.46500492095947
40 1.84349310398102
41 1.78528261184692
42 2.09022092819214
43 1.95127475261688
44 1.4470226764679
45 1.61222898960114
46 1.53228664398193
47 1.69086790084839
48 1.07631313800812
49 1.11110711097717
50 1.11585092544556
51 1.15991163253784
52 1.17702031135559
53 1.16306447982788
54 1.23091578483582
55 1.22612965106964
56 1.16127192974091
57 1.20257127285004
58 1.14161789417267
59 1.19440579414368
60 1.32979035377502
61 1.29413437843323
62 1.29543244838715
63 1.29062104225159
};
\addlegendentry{Stylegan}
\addplot [semithick, color2, mark=*, mark size=1.25, mark options={solid}]
table {%
0 6.08049726486206
1 3.57643413543701
2 3.82460999488831
3 2.94471836090088
4 2.07009649276733
5 2.67697024345398
6 1.79753088951111
7 2.24992632865906
8 2.5096275806427
9 1.91828894615173
10 3.04633474349976
11 2.39808940887451
12 1.25041627883911
13 1.49777889251709
14 1.55823302268982
15 1.8799489736557
16 0.502096712589264
17 1.00441479682922
18 0.349537938833237
19 0.679293870925903
20 1.30528163909912
21 1.56310665607452
22 1.23667168617249
23 1.45450282096863
24 0.162485957145691
25 0.37017098069191
26 0.296047151088715
27 0.544218599796295
28 0.839464128017426
29 0.816530764102936
30 0.996208488941193
31 0.993605077266693
32 0.580958545207977
33 0.506429374217987
34 1.06396281719208
35 0.971673309803009
36 0.246746361255646
37 0.402561604976654
38 0.537291646003723
39 0.74968683719635
40 1.47918140888214
41 1.44576394557953
42 1.65327787399292
43 1.53611958026886
44 0.917668879032135
45 1.20164752006531
46 0.933083772659302
47 1.2078891992569
48 -0.429628014564514
49 -0.286194711923599
50 -0.310498803853989
51 -0.111236430704594
52 0.0698592886328697
53 -0.00726996175944805
54 0.26937660574913
55 0.190933629870415
56 -0.0124184647575021
57 0.218539714813232
58 -0.0986084118485451
59 0.123818330466747
60 0.666546285152435
61 0.563744902610779
62 0.553709387779236
63 0.475711196660995
};
\addlegendentry{Stylegan2}
\addplot [semithick, color3, mark=o, mark size=1.25, mark options={solid}]
table {%
0 6.07744550704956
1 3.59728837013245
2 3.83062434196472
3 2.96199703216553
4 2.08548331260681
5 2.69255471229553
6 1.81259369850159
7 2.26596784591675
8 2.51700043678284
9 1.9248012304306
10 3.05909872055054
11 2.4172351360321
12 1.25206768512726
13 1.51067090034485
14 1.56126356124878
15 1.89714217185974
16 0.501448810100555
17 1.00735437870026
18 0.359202235937119
19 0.691763281822205
20 1.30585527420044
21 1.56921970844269
22 1.24547779560089
23 1.46312236785889
24 0.169863224029541
25 0.370447337627411
26 0.306937992572784
27 0.553068459033966
28 0.840910613536835
29 0.825126171112061
30 1.00408291816711
31 1.00166189670563
32 0.578603088855743
33 0.509569108486176
34 1.0696816444397
35 0.970267474651337
36 0.257836103439331
37 0.40316054224968
38 0.539294540882111
39 0.757333993911743
40 1.49824178218842
41 1.45751392841339
42 1.66507089138031
43 1.55191767215729
44 0.928058445453644
45 1.21466720104218
46 0.944511950016022
47 1.222416639328
48 -0.435364425182343
49 -0.278990030288696
50 -0.314322590827942
51 -0.107393637299538
52 0.0691655427217484
53 -0.0041497815400362
54 0.259980887174606
55 0.196431308984756
56 -0.01834930293262
57 0.216042190790176
58 -0.104993134737015
59 0.126857489347458
60 0.672769069671631
61 0.56843101978302
62 0.55586963891983
63 0.472281783819199
};
\addlegendentry{Stylegan3}
\end{axis}
\end{tikzpicture}
\begin{tikzpicture}

\begin{axis}[
colorbar,
colorbar style={ylabel={}},
colormap/viridis,
point meta max=0.579453621658024,
point meta min=0.00226394714518854,
tick align=outside,
tick pos=left,
title={FFHQ-Stylegan-Stylegan2-Stylegan3 db4 CNN saliency },
x grid style={white!69.0196078431373!black},
xmin=-0.5, xmax=127.5,
xtick style={color=black},
y dir=reverse,
y grid style={white!69.0196078431373!black},
ymin=-0.5, ymax=127.5,
ytick style={color=black}
]
\addplot graphics [includegraphics cmd=\pgfimage,xmin=-0.5, xmax=127.5, ymin=127.5, ymax=-0.5] {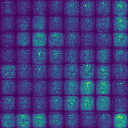};
\end{axis}

\end{tikzpicture}
  \caption{The left figure shows db4 mean wavelets for \ac{ffhq}, StyleGAN~\citep{karras2019pro}, StyleGAN2~\citep{karras2020analyzing} and StyleGAN3~\citep{Karras2021Alias} images.
  Standard deviations are included in Supplementary plot~\ref{fig:ffhq_hard_mean_std}. The right side shows a saliency map \citep{SimonyanVZ13} for the \ac{ln}-db4-\ac{cnn}-classifier from Table~\ref{tab:ffhq_hard}. All wavelet filter combinations are labeled in supplementary Figure~\ref{fig:packetlabels}}\label{fig:wp_sali_ffhq_hard}
\end{figure}
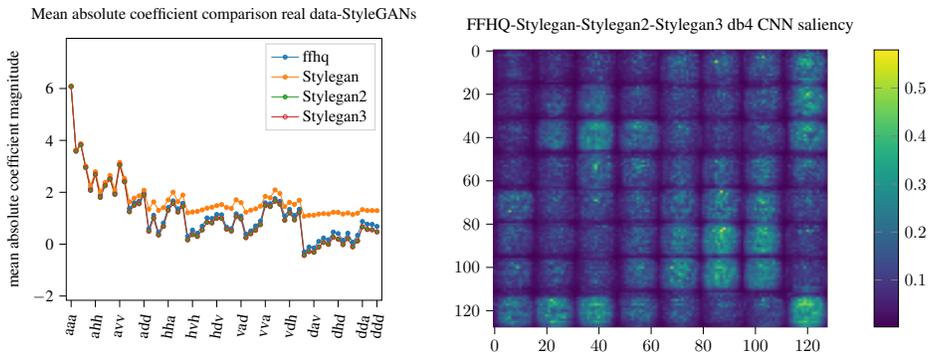
As before, we investigate mean ln-db4-wavelet packet plots and saliency maps.
The mean wavelet coefficients on the left of Figure~\ref{fig:wp_sali_ffhq_hard} reveal an improved ability to accurately model the spectrum for the StyleGAN2  and StyleGAN2 architectures, yet differences to FFHQ remain. Supplementary Table~\ref{tab:conf_ffhq_hard} confirms this observation, the two newer GANs are misclassified more often.
The per packet mean and standard deviation for StyleGAN2 and StyleGAN3 is almost identical.
In difference to Figure~\ref{fig:packets_sali_celebA}, the saliency score indicates a more wide-spread importance of packets.
We see a peak at the very high frequency packets and generally a higher saliency at higher frequencies.

\subsection{Detecting partial manipulations in Face-Forensics++}\label{sec:ffpp}
In all previously studied images, all pixels were either original or fake. This section studies the identification of partial manipulations.
To this end, we consider a subset of the \ac{ffpp} video-deepfake detection benchmark \citep{rossler2019faceforensics++}.
The full data set contains original videos as well as manipulated videos. Altered clips are deepfaked \citep{deepfake2022github}, include neural textures \citep{thies2019deferred}, or have been modified using the more traditional face-swapping \citep{Kowalski2022face} or face-to-face \citep{thies2016face2face} translation methods. The first two methods are deep learning-based.

Data preparation and frame extraction follows the standardized setup from \cite{rossler2019faceforensics++}. Seven hundred twenty videos are used for training and 140 each for validation and testing. We sample 260\footnote{\cite{rossler2019faceforensics++} tells us to use 270, but the training set contains a video, where the face extraction pipeline could only find 262 frames with a face. We, therefore, reduced the total number accordingly across the board.} images from each training video and 100 from each test and validation clip. A fake or real binary label is assigned to each frame. After re-scaling each frame to $[0, 1]$, data preprocessing and network training follows the description from section~\ref{sec:train_celeba_lsun}. We trained all networks for 20 epochs.

\begin{table}[tbp]
  \centering
  \caption{Detection results on the neural-subset of the face-forensics++ \citep{rossler2019faceforensics++}. The networks are tasked to identify original videos, deepfakes \citep{deepfake2022github} and neural texture modifications \citep{thies2019deferred}. The Pixel-CNN architecture has a large parameter cluster in the final classifier. Since \ac{ffpp} is a real-fake binary problem, we see a parameter reduction for the Pixel-CNN in comparison to previous experiments.} \label{tab:ff++}
  \begin{tabular}{l r r r} \toprule
    & &\multicolumn{2}{c}{Accuracy on neural \ac{ffpp} [\%]} \\ \cmidrule(lr){3-4}
    Method                    & parameters  &  max  & $\mu \pm \sigma$ \\ \midrule
    CNN-pixel                 & 57k        &  96.02  & $94.43\pm 1.81 $   \\
    CNN-db4                   & 109k       &  97.66  & $97.50\pm 0.25 $   \\
    CNN-sym4                  & 109k       &  \textbf{98.02}  & $97.71\pm 0.25 $   \\ \bottomrule
  \end{tabular}
\end{table}
We first consider the neural subset of pristine images, deepfakes and images containing neural textures. Results are tabulated in Table~\ref{tab:ff++}. We find that our classifiers outperform the pixel-\ac{cnn} baseline for both db4 and sym4 wavelets. Note, the \ac{ffpp}-data set only modifies the facial region of each image. A possible explanation could be that the first and second scales produce coarser frequency resolutions that do not help in this specific case.

According to supplementary Figures~\ref{fig:ffpp_wp_single} and \ref{fig:ffpp_wp_grouped} the deep learning based methods create frequency domain artifacts similar to those we saw in Figures~\ref{fig:mean_packets_ffhq} and \ref{fig:classifier_plot}. Therefore, we again see an improved performance for the packet-based classifiers in Table~\ref{tab:ff++}.

Supplementary Table~\ref{tab:fullffpp} lists results on the complete data set for three compression-levels. The wavelet-based networks are competitive if compared to approaches without Image-Net pretraining. We highlight the high-quality setting C23 in particular. However, the full data set includes classic computer-vision methods that do not rely on deep learning. These methods produce fewer divergences in the higher wavelet packet coefficients, making these harder to detect. See supplementary section~\ref{sec:fullffpp} for an extended discussion of supplementary Table~\ref{tab:fullffpp}.

\subsection{Guided diffusion on LSUN}\label{sec:guided_diffusion}
\begin{figure}
\includestandalone[width=0.4\linewidth]{./figures/guided_diffusion/absolute_coeff_comparison_lsun_guide_diff}
\begin{tikzpicture}

\definecolor{darkgray176}{RGB}{176,176,176}

\begin{axis}[
colorbar,
colorbar style={ylabel={}},
colormap/viridis,
point meta max=0.056955710146775,
point meta min=1.59559614792158e-05,
tick align=outside,
tick pos=left,
x grid style={darkgray176},
xmin=-0.5, xmax=255.5,
xtick style={color=black},
y dir=reverse,
y grid style={darkgray176},
ymin=-0.5, ymax=255.5,
ytick style={color=black},
title={Guided-Diffusion-LSUN CNN saliency },
]
\addplot graphics [includegraphics cmd=\pgfimage,xmin=-0.5, xmax=255.5, ymin=255.5, ymax=-0.5] {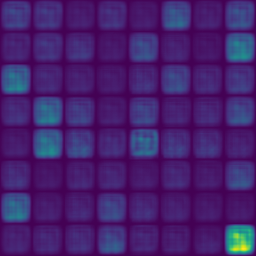};
\end{axis}

\end{tikzpicture}
\caption{A logarithmically scaled Haar-wavelet packet representation is shown on the left. The mean packets of 40k LSUN-bedroom images are shown in blue. The orange plot shows the mean packets of 40k images generated by guided diffusion. In both cases the image resolution was 256 by 256 pixels. We see that guided diffusion approximates the frequency-spectrum well. However, a small systematic increase is visible for higher frequencies. On the right we show the CNN-saliency for the CNN classifier trained on top of the high resolution packet representation. Similarly to the GAN classifier it appears to focus on the highest frequency packet.}
\label{fig:diffusion_analysis}
\end{figure}
To evaluate our detection method on non-\ac{gan}-based deep learning generators, we consider the recent guided diffusion approach.
We study images generated using the supplementary code written by \cite{dhariwal2021diffusion}\footnote{available online at: \url{https://github.com/openai/guided-diffusion}} on the \ac{lsun}-bedrooms data set.
\begin{table}
  \centering
  \caption{Classification accuracy using various feature representations on the LSUN and Guided-Diffusion seperation problem.}\label{tab:diff_lsun}
  \begin{tabular}{l r r r} \toprule
    & &\multicolumn{2}{c}{Accuracy[\%]} \\ \cmidrule(lr){3-4}
    Method                    & parameters  &  max  & $\mu \pm \sigma$           \\ \midrule
    CNN-pixel                 & 57k        &  80.03  & $71.46\pm 10.99 $         \\
    CNN-ln-Fourier            & 57k        &  73.55  & $69.59\pm 3.94 $          \\
    CNN-ln-Haar               & 109k       &  \textbf{96.75}  & $96.50\pm 0.23 $ \\ \bottomrule
  \end{tabular}
\end{table}
We start by considering 40k images per class at a 128x128 pixel resolution. We use 38k images for training and set 2k aside. We split equally and work with 1k images for validation and 1k for testing. For our analysis, we work with the wavelet packet representation as described in section \ref{sec:ffhq_linear}. Additionally, we test a pixel and Fourier representation. Table~\ref{tab:diff_lsun} shows that the wavelet packet approach works comparatively well in this setting.

We investigate further and study an additional 40k images per class at a higher 256x256-Pixel resolution. We visualize the third-level Haar-wavelet packets in Figure~\ref{fig:diffusion_analysis} on the left.
The coefficients for the images generated by guided diffusion exhibit a slightly elevated mean and a reduced standard deviation. Using stacked wavelet packets with the spatial dimension intact, we train a Haar-CNN as described in section \ref{sec:train_celeba_lsun}. To work at 256 by 256 pixels, we set the size of the final fully connected layer for the Wavelet-Packet architecture from Table~\ref{tab:nets} to $24 \cdot 17 \cdot 17, 2$.
For five experiments, we observe a test accuracy of $99.04 \pm 0.6$. We conclude that the wavelet-packet approach reliably identifies images generated by guided diffusion. We did not find higher-order wavelets beneficial for the guided-diffusion data and leave the investigation of possible causes for future work.

\section{Conclusion and outlook}\label{sec:conclusion}
We introduced a wavelet packet-based approach for deepfake analysis and detection, which is based on a multi-scale image representation in space and frequency.
We saw that wavelet-packets allow the visualization of frequency domain differences while preserving some spatial information.
In particular, we found diverging mean values for packets at high frequencies.
At the same time, the bulk of the standard deviation differences were at the edges and within the background portions of the images.
This observation suggests contemporary \ac{gan} architectures still fail to capture the backgrounds and high-frequency information in full detail.
We found that coupling higher-order wavelets and \ac{cnn} led to an improved or competitive performance in comparison to a \ac{dct} approach or working directly on the raw images, where fused architectures shows the best performance.
Furthermore, the employed lean neural network architecture allows efficient training and also can achieve similar accuracy with only 20\% of the training data. Overall our classifiers deliver state-of-the-art performance, require few learnable parameters, and converge quickly.

Even though releasing our detection code will allow potential bad actors to test against it, we hope our approach will complement the publicly available deepfake identification toolkits.
We propose to further investigate the resilience of multi-classifier ensembles in future work
and envision a framework of multiple forensic classifiers, which together give strong and robust results for artificial image identification.
Future work could also explore the use of curvelets and shearlets for deepfake detection.
Integrating image data from diffusion models and GANs into a single standardized data set will also be an important task in future work.

\section{Declarations}
\textbf{Funding} This work was supported by the Fraunhofer Cluster of Excellence Cognitive Internet Technologies (CCIT). \\
\textbf{Conflicts of interest/Competing interests} Not applicable  \\
\textbf{Ethics approval} Not Applicable               \\
\textbf{Consent to participate} Not Applicable             \\
\textbf{Consent for publication} The human image in Figure~\ref{fig:proof_of_concept} is part of the Flickr Faces high qualiy data set. \cite{karras2019stylebased} created the data set and clarified the license: ``The individual images were published in Flickr by their respective authors under either Creative Commons BY 2.0, Creative Commons BY-NC 2.0, Public Domain Mark 1.0, Public Domain CC0 1.0, or U.S. Government Works license''.\\
\textbf{Availability of data and material} The \ac{ffhq} data set is available at \url{https://github.com/NVlabs/ffhq-dataset}, \ac{lsun}-data is available via download scripts from \url{https://github.com/fyu/lsun}. Download links for \ac{celeba} are online at \url{https://mmlab.ie.cuhk.edu.hk/projects/CelebA.html}. Finally, the \ac{ffpp}-videos we worked with are available via code from \url{https://github.com/ondyari/FaceForensics}. \\
\textbf{Code availability}  Tested and documented code for the boundary wavelet computations is available at \url{https://github.com/v0lta/PyTorch-Wavelet-Toolbox}. The source for the Deepfake detection experiments is available at \url{https://github.com/gan-police/frequency-forensics} . \\
\textbf{Authors' contributions}:
\textit{Conceptualization}: MW, JG;
\textit{Data curation}: FB;
\textit{Formal analysis}: MW, FB;
\textit{Funding acquisition}: JG;
\textit{Investigation}: FB, MW;
\textit{Methodology}: MW;
\textit{Project Administration}: JG, MW;
\textit{Resources}: JG;
\textit{Software}: MW, FB, RH;
\textit{Supervision}: MW, JG;
\textit{Validation}: MW, FB, JG;
\textit{Visualization}: MW, FB;
\textit{Writing - original draft}: MW;
\textit{Writing - review and editing}: JG, RH, MW, FB.
\\

\section{Acknowledgements}
We are greatly indebted to Charly Hoyt for his invaluable help with the setup of our unit-test and pip packaging framework, as well as his feedback regarding earlier versions of this paper.
We thank Helmut Harbrecht and Daniel Domingo-Fern\`andez for insightful discussions. Finally, we thank Andr\'e Gemünd, and Olga Rodikow for administrating the compute clusters at SCAI. Without their help, this project would not exist.

\bibliographystyle{apalike}
\bibliography{library}

\printacronyms

\section{Supplementary}
In this supplementary, we share additional sparsity  patterns of our fast wavelet transformation matrices, as well as additional experimental details.

\subsection{The perfect reconstruction and alias cancellation conditions}\label{subsec:pc_ac}
When using wavelet filters, we expect a lossless representation free of aliasing. Consequently, we want to restrict ourselves to filters, which guarantee both. Classic literature like \cite{strang1996wavelets} presents two equations to enforce this, the perfect reconstruction and alias cancellation conditions.
Starting from the analysis filter coefficients $\mathbf{h}$ and the synthesis filter coefficients $\mathbf{f}$.
For a complex number $z \in \mathbb{C}$ their $z$-transformed counterparts are $H(z) = \sum_n \mathbf{h}(n) z^{-n}$ and $F(z)$ respectively. The reconstruction condition can now be expressed as
\begin{equation}
H_\mathcal{L}(z) F_\mathcal{L}(z) + H_\mathcal{H}(z) F_\mathcal{H}(z) = 2z^c,
\label{eq:pr_condition}
\end{equation}
and the anti-aliasing condition as
\begin{equation} \label{eq:ac}
H_\mathcal{L}(-z) F_\mathcal{L}(z) + H_\mathcal{H}(-z) F_\mathcal{H}(z) = 0.
\end{equation}
For the perfect reconstruction condition in Eq.~\ref{eq:pr_condition}, the center term $z^l$ of the resulting $z$-transformed expression must be a two; all other coefficients should be zero. $c$ denotes the power of the center.
The effect of the alias cancellation condition can be studied by looking at Daubechies wavelets and symlets, which we will do in the next section. The perfect reconstruction condition is visible too, but harder to see. For an in-depth discussion of both conditions, we refer the interested reader to the excellent textbooks by \cite{strang1996wavelets} and \cite{jensen2001ripples}.

\subsection{Daubechies wavelets and symlets}\label{sec:daubsym}
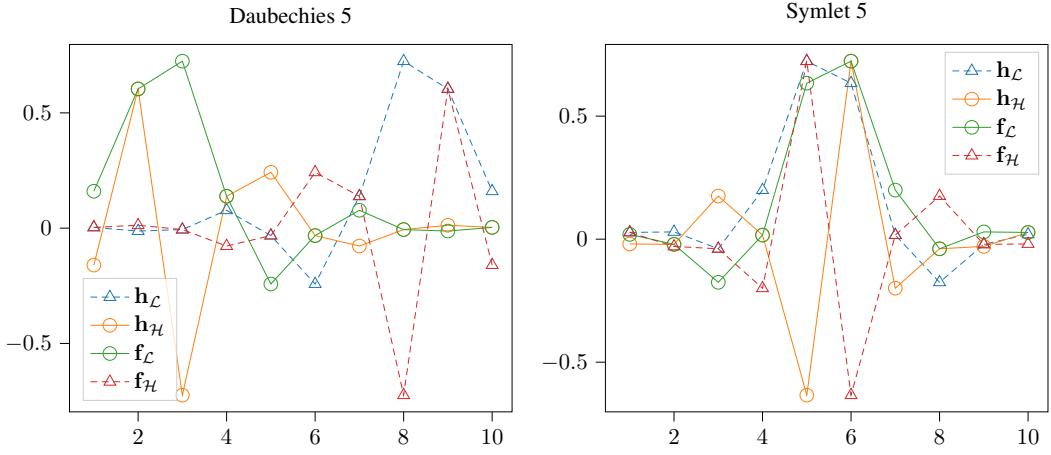
\begin{figure}[tbp]
\begin{tikzpicture}

\definecolor{color0}{rgb}{0.12156862745098,0.466666666666667,0.705882352941177}
\definecolor{color1}{rgb}{1,0.498039215686275,0.0549019607843137}
\definecolor{color2}{rgb}{0.172549019607843,0.627450980392157,0.172549019607843}
\definecolor{color3}{rgb}{0.83921568627451,0.152941176470588,0.156862745098039}

\begin{axis}[
legend cell align={left},
legend style={
  fill opacity=0.8,
  draw opacity=1,
  text opacity=1,
  at={(0.03,0.03)},
  anchor=south west,
  draw=white!80!black
},
tick align=outside,
tick pos=left,
title={Daubechies 5},
x grid style={white!69.0196078431373!black},
xmin=0.45, xmax=10.45,
xtick style={color=black},
y grid style={white!69.0196078431373!black},
ymin=-0.79673938128155, ymax=0.79673938128155,
ytick style={color=black}
]
\addplot [densely dashed, mark=triangle, mark size=3, mark options={solid}, color0]
table {%
1 0.00333572528547377
2 -0.012580751999082
3 -0.00624149021279827
4 0.0775714938400457
5 -0.0322448695846384
6 -0.242294887066382
7 0.138428145901321
8 0.724308528437773
9 0.60382926979719
10 0.160102397974193
};
\addlegendentry{$\mathbf{h}_\mathcal{L}$}
\addplot [solid, mark=o, mark size=3, mark options={solid}, color1]
table {%
1 -0.160102397974193
2 0.60382926979719
3 -0.724308528437773
4 0.138428145901321
5 0.242294887066382
6 -0.0322448695846384
7 -0.0775714938400457
8 -0.00624149021279827
9 0.012580751999082
10 0.00333572528547377
};
\addlegendentry{$\mathbf{h}_\mathcal{H}$}
\addplot [solid, mark=o, mark size=3, mark options={solid}, color2]
table {%
1 0.160102397974193
2 0.60382926979719
3 0.724308528437773
4 0.138428145901321
5 -0.242294887066382
6 -0.0322448695846384
7 0.0775714938400457
8 -0.00624149021279827
9 -0.012580751999082
10 0.00333572528547377
};
\addlegendentry{$\mathbf{f}_\mathcal{L}$}
\addplot [densely dashed, mark=triangle, mark size=3, mark options={solid}, color3]
table {%
1 0.00333572528547377
2 0.012580751999082
3 -0.00624149021279827
4 -0.0775714938400457
5 -0.0322448695846384
6 0.242294887066382
7 0.138428145901321
8 -0.724308528437773
9 0.60382926979719
10 -0.160102397974193
};
\addlegendentry{$\mathbf{f}_\mathcal{H}$}
\end{axis}

\end{tikzpicture}\hfill%
\begin{tikzpicture}

\definecolor{color0}{rgb}{0.12156862745098,0.466666666666667,0.705882352941177}
\definecolor{color1}{rgb}{1,0.498039215686275,0.0549019607843137}
\definecolor{color2}{rgb}{0.172549019607843,0.627450980392157,0.172549019607843}
\definecolor{color3}{rgb}{0.83921568627451,0.152941176470588,0.156862745098039}

\begin{axis}[
legend cell align={left},
legend style={fill opacity=0.8, draw opacity=1, text opacity=1, draw=white!80!black},
tick align=outside,
tick pos=left,
title={Symlet 5},
x grid style={white!69.0196078431373!black},
xmin=0.45, xmax=10.45,
xtick style={color=black},
y grid style={white!69.0196078431373!black},
ymin=-0.701848296151244, ymax=0.791277023095452,
ytick style={color=black}
]
\addplot [densely dashed, mark=triangle, mark size=3, mark options={solid}, color0]
table {%
1 0.027333068345078
2 0.0295194909257746
3 -0.0391342493023831
4 0.199397533977394
5 0.723407690402421
6 0.633978963458212
7 0.0166021057645223
8 -0.17532808990845
9 -0.0211018340247589
10 0.0195388827352867
};
\addlegendentry{$\mathbf{h}_\mathcal{L}$}
\addplot [solid, mark=o, mark size=3, mark options={solid}, color1]
table {%
1 -0.0195388827352867
2 -0.0211018340247589
3 0.17532808990845
4 0.0166021057645223
5 -0.633978963458212
6 0.723407690402421
7 -0.199397533977394
8 -0.0391342493023831
9 -0.0295194909257746
10 0.027333068345078
};
\addlegendentry{$\mathbf{h}_\mathcal{H}$}
\addplot [solid, mark=o, mark size=3, mark options={solid}, color2]
table {%
1 0.0195388827352867
2 -0.0211018340247589
3 -0.17532808990845
4 0.0166021057645223
5 0.633978963458212
6 0.723407690402421
7 0.199397533977394
8 -0.0391342493023831
9 0.0295194909257746
10 0.027333068345078
};
\addlegendentry{$\mathbf{f}_\mathcal{L}$}
\addplot [densely dashed, mark=triangle, mark size=3, mark options={solid}, color3]
table {%
1 0.027333068345078
2 -0.0295194909257746
3 -0.0391342493023831
4 -0.199397533977394
5 0.723407690402421
6 -0.633978963458212
7 0.0166021057645223
8 0.17532808990845
9 -0.0211018340247589
10 -0.0195388827352867
};
\addlegendentry{$\mathbf{f}_\mathcal{H}$}
\end{axis}

\end{tikzpicture}%
\caption{Fifth degree Daubechies and Symlet filters side by side.}
\label{fig:db5sym5}
\end{figure}
Figure~\ref{fig:db5sym5} plots both Daubechies filter pairs on the left.
A possible way to solve the anti-aliasing condition formulated in equation~\ref{eq:ac} is to require, $F_\mathcal{L}(z) = H_\mathcal{H}(-z)$ and $F_\mathcal{H}(z) = - H_\mathcal{L}(-z)$ \citep{strang1996wavelets}.
The patterns this solution produces are visible for both the Daubechies filters as well as their symmetric counterparts shown above. To see why substitute $(-z)$. It will produce a minus sign at odd powers in the coefficient polynomial. Multiplication with $(-1)$ shifts the pattern to even powers. Whenever $F_\mathcal{L}$ and $H_\mathcal{H}$ share the same sign $F_\mathcal{H}$ and $H_\mathcal{L}$ do not and the other way around. The same pattern is visible for the Symlet on the right of figure~\ref{fig:db5sym5}.

The Daubechies wavelets are very anti-symmetric \citep{mallat2009wavelet}. Symlets have been designed as an alternative with similar properties. But, as shown in \ref{fig:db5sym5}, Symlets are symmetric and centered.

\subsection{Constructing the inverse fwt matrix}\label{subsec:ifwt}
Section~\ref{subsec:fwt} presented the structure of the Analysis matrix $\mathbf{A}$. Its inverse is again a linear operation. We can write:
\begin{align}
\mathbf{S}\mathbf{b} = \mathbf{x}.
\end{align}
The synthesis matrix $\mathbf{S}$ reverts the operations in $\mathbf{A}$. To construct it one requires the synthesis filter pair $\mathbf{f}_\mathcal{L}, \mathbf{f}_\mathcal{H}$~\cite{strang1996wavelets}.  Structurally the synthesis matrices are transposed in comparison to their analysis counterparts.
Using the transposed convolution matrices $\mathbf{F}_\mathcal{L}$ and $\mathbf{F}_\mathcal{H}$, $\mathbf{S}$, one builds:
\begin{align}
\mathbf{S}=
\begin{pmatrix}
\mathbf{F}_\mathcal{L} & \mathbf{F}_\mathcal{H}
\end{pmatrix}
\begin{pmatrix}
\begin{array}{c c| c}
\mathbf{F}_\mathcal{L} & \mathbf{F}_\mathcal{H} &  \\ \hline
  & & \mathbf{I} \\
\end{array}
\end{pmatrix}
\dots .
\end{align}
The pattern here is analog to the one we saw in section~\ref{subsec:fwt}.
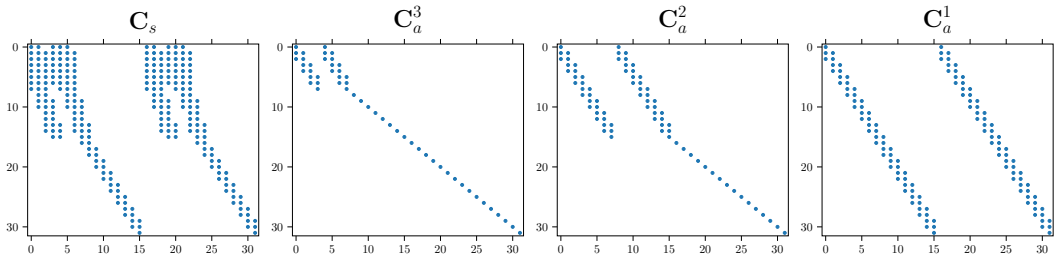
\begin{figure}[p]
\begin{tikzpicture}

\definecolor{color0}{rgb}{0.12156862745098,0.466666666666667,0.705882352941177}

\begin{groupplot}[group style={group size=4 by 1}]
\nextgroupplot[
tick align=outside,
title={\huge \(\displaystyle \mathbf{C}_s\)},
x grid style={white!69.0196078431373!black},
xmin=-0.5, xmax=31.5,
xtick pos=both,
xtick style={color=black},
y dir=reverse,
y grid style={white!69.0196078431373!black},
ymin=-0.5, ymax=31.5,
ytick pos=left,
ytick style={color=black}
]
\addplot [semithick, color0, mark=*, mark size=1.25, mark options={solid}, only marks]
table {%
0 0
1 0
3 0
4 0
5 0
16 0
17 0
19 0
20 0
21 0
0 1
1 1
2 1
3 1
4 1
5 1
6 1
16 1
17 1
18 1
19 1
20 1
21 1
22 1
0 2
1 2
2 2
3 2
4 2
5 2
6 2
16 2
17 2
18 2
19 2
20 2
21 2
22 2
0 3
1 3
2 3
3 3
4 3
5 3
6 3
16 3
17 3
18 3
19 3
20 3
21 3
22 3
0 4
1 4
2 4
3 4
4 4
5 4
6 4
16 4
17 4
18 4
19 4
20 4
21 4
22 4
0 5
1 5
2 5
3 5
4 5
5 5
6 5
16 5
17 5
18 5
19 5
20 5
21 5
22 5
0 6
1 6
2 6
3 6
4 6
5 6
6 6
16 6
17 6
18 6
19 6
20 6
21 6
22 6
0 7
1 7
2 7
4 7
5 7
6 7
16 7
17 7
18 7
20 7
21 7
22 7
1 8
2 8
5 8
6 8
17 8
18 8
21 8
22 8
1 9
2 9
3 9
5 9
6 9
7 9
17 9
18 9
19 9
21 9
22 9
23 9
1 10
2 10
3 10
5 10
6 10
7 10
17 10
18 10
19 10
21 10
22 10
23 10
2 11
3 11
6 11
7 11
18 11
19 11
22 11
23 11
2 12
3 12
6 12
7 12
18 12
19 12
22 12
23 12
2 13
3 13
4 13
6 13
7 13
8 13
18 13
19 13
20 13
22 13
23 13
24 13
2 14
3 14
4 14
6 14
7 14
8 14
18 14
19 14
20 14
22 14
23 14
24 14
3 15
4 15
7 15
8 15
19 15
20 15
23 15
24 15
7 16
8 16
23 16
24 16
8 17
9 17
24 17
25 17
8 18
9 18
24 18
25 18
9 19
10 19
25 19
26 19
9 20
10 20
25 20
26 20
10 21
11 21
26 21
27 21
10 22
11 22
26 22
27 22
11 23
12 23
27 23
28 23
11 24
12 24
27 24
28 24
12 25
13 25
28 25
29 25
12 26
13 26
28 26
29 26
13 27
14 27
29 27
30 27
13 28
14 28
29 28
30 28
14 29
15 29
30 29
31 29
14 30
15 30
30 30
31 30
15 31
31 31
};

\nextgroupplot[
tick align=outside,
title={\huge \(\displaystyle \mathbf{C}_a^3\)},
x grid style={white!69.0196078431373!black},
xmin=-0.5, xmax=31.5,
xtick pos=both,
xtick style={color=black},
y dir=reverse,
y grid style={white!69.0196078431373!black},
ymin=-0.5, ymax=31.5,
ytick pos=left,
ytick style={color=black}
]
\addplot [semithick, color0, mark=*, mark size=1.25, mark options={solid}, only marks]
table {%
0 0
4 0
0 1
1 1
4 1
5 1
0 2
1 2
4 2
5 2
1 3
2 3
5 3
6 3
1 4
2 4
5 4
6 4
2 5
3 5
6 5
7 5
2 6
3 6
6 6
7 6
3 7
7 7
8 8
9 9
10 10
11 11
12 12
13 13
14 14
15 15
16 16
17 17
18 18
19 19
20 20
21 21
22 22
23 23
24 24
25 25
26 26
27 27
28 28
29 29
30 30
31 31
};

\nextgroupplot[
tick align=outside,
title={\huge \(\displaystyle \mathbf{C}_a^2\)},
x grid style={white!69.0196078431373!black},
xmin=-0.5, xmax=31.5,
xtick pos=both,
xtick style={color=black},
y dir=reverse,
y grid style={white!69.0196078431373!black},
ymin=-0.5, ymax=31.5,
ytick pos=left,
ytick style={color=black}
]
\addplot [semithick, color0, mark=*, mark size=1.25, mark options={solid}, only marks]
table {%
0 0
8 0
0 1
1 1
8 1
9 1
0 2
1 2
8 2
9 2
1 3
2 3
9 3
10 3
1 4
2 4
9 4
10 4
2 5
3 5
10 5
11 5
2 6
3 6
10 6
11 6
3 7
4 7
11 7
12 7
3 8
4 8
11 8
12 8
4 9
5 9
12 9
13 9
4 10
5 10
12 10
13 10
5 11
6 11
13 11
14 11
5 12
6 12
13 12
14 12
6 13
7 13
14 13
15 13
6 14
7 14
14 14
15 14
7 15
15 15
16 16
17 17
18 18
19 19
20 20
21 21
22 22
23 23
24 24
25 25
26 26
27 27
28 28
29 29
30 30
31 31
};

\nextgroupplot[
tick align=outside,
title={\huge \(\displaystyle \mathbf{C}_a^1\)},
x grid style={white!69.0196078431373!black},
xmin=-0.5, xmax=31.5,
xtick pos=both,
xtick style={color=black},
y dir=reverse,
y grid style={white!69.0196078431373!black},
ymin=-0.5, ymax=31.5,
ytick pos=left,
ytick style={color=black}
]
\addplot [semithick, color0, mark=*, mark size=1.25, mark options={solid}, only marks]
table {%
0 0
16 0
0 1
1 1
16 1
17 1
0 2
1 2
16 2
17 2
1 3
2 3
17 3
18 3
1 4
2 4
17 4
18 4
2 5
3 5
18 5
19 5
2 6
3 6
18 6
19 6
3 7
4 7
19 7
20 7
3 8
4 8
19 8
20 8
4 9
5 9
20 9
21 9
4 10
5 10
20 10
21 10
5 11
6 11
21 11
22 11
5 12
6 12
21 12
22 12
6 13
7 13
22 13
23 13
6 14
7 14
22 14
23 14
7 15
8 15
23 15
24 15
7 16
8 16
23 16
24 16
8 17
9 17
24 17
25 17
8 18
9 18
24 18
25 18
9 19
10 19
25 19
26 19
9 20
10 20
25 20
26 20
10 21
11 21
26 21
27 21
10 22
11 22
26 22
27 22
11 23
12 23
27 23
28 23
11 24
12 24
27 24
28 24
12 25
13 25
28 25
29 25
12 26
13 26
28 26
29 26
13 27
14 27
29 27
30 27
13 28
14 28
29 28
30 28
14 29
15 29
30 29
31 29
14 30
15 30
30 30
31 30
15 31
31 31
};
\end{groupplot}

\end{tikzpicture}
  \caption{Sparsity pattern of the truncated 32 by 32 level 3 synthesis convolution matrix, and its scale components.
           The three individual decomposition matrices are shown in increasing order from the right to the left.
           On the very left the product of all three is shown.
           The pattern occurs for second degree wavelets.}
  \label{fig:conv_synthesis}
\end{figure}
In Figure~\ref{fig:conv_synthesis} we show a truncated example. In comparison to Figure~\ref{fig:conv_analysis} the structure is transposed.
Note, in order to guarantee invertibility one must have $\mathbf{S} \mathbf{A}~=~\mathbf{I}$. Which is the case for infinitely large matrices.
When working with real truncated matrices, one requires boundary wavelet treatment, see section~\ref{sec:boundary_wavelets}.

\subsection{Sparsity patterns of boundary wavelet matrices}
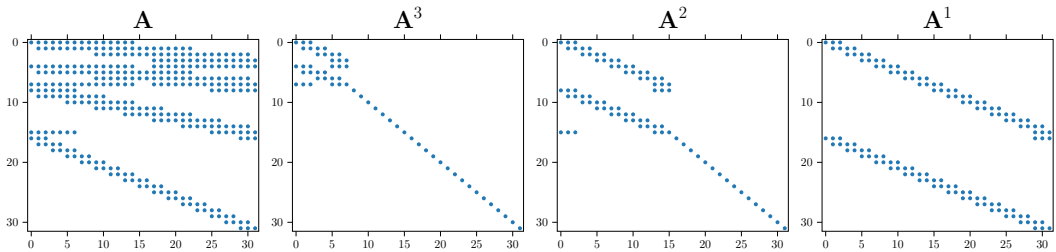
\begin{figure}[p]
\begin{tikzpicture}

\definecolor{color0}{rgb}{0.12156862745098,0.466666666666667,0.705882352941177}

\begin{groupplot}[group style={group size=4 by 1}]
\nextgroupplot[
tick align=outside,
title={\huge \(\displaystyle \mathbf{A}\)},
x grid style={white!69.0196078431373!black},
xmin=-0.5, xmax=31.5,
xtick pos=both,
xtick style={color=black},
y dir=reverse,
y grid style={white!69.0196078431373!black},
ymin=-0.5, ymax=31.5,
ytick pos=left,
ytick style={color=black}
]
\addplot [semithick, color0, mark=*, mark size=1.25, mark options={solid}, only marks]
table {%
0 0
1 0
2 0
3 0
4 0
5 0
6 0
7 0
8 0
9 0
10 0
11 0
12 0
13 0
14 0
1 1
2 1
3 1
4 1
5 1
6 1
7 1
8 1
9 1
10 1
11 1
12 1
13 1
14 1
15 1
16 1
17 1
18 1
19 1
20 1
21 1
22 1
9 2
10 2
11 2
12 2
13 2
14 2
15 2
16 2
17 2
18 2
19 2
20 2
21 2
22 2
23 2
24 2
25 2
26 2
27 2
28 2
29 2
30 2
17 3
18 3
19 3
20 3
21 3
22 3
23 3
24 3
25 3
26 3
27 3
28 3
29 3
30 3
31 3
0 4
1 4
2 4
3 4
4 4
5 4
6 4
7 4
8 4
9 4
10 4
11 4
12 4
13 4
14 4
17 4
18 4
19 4
20 4
21 4
22 4
23 4
24 4
25 4
26 4
27 4
28 4
29 4
30 4
31 4
1 5
2 5
3 5
4 5
5 5
6 5
7 5
8 5
9 5
10 5
11 5
12 5
13 5
14 5
15 5
16 5
17 5
18 5
19 5
20 5
21 5
22 5
9 6
10 6
11 6
12 6
13 6
14 6
15 6
16 6
17 6
18 6
19 6
20 6
21 6
22 6
23 6
24 6
25 6
26 6
27 6
28 6
29 6
30 6
0 7
1 7
2 7
3 7
4 7
5 7
6 7
7 7
8 7
9 7
10 7
11 7
12 7
13 7
14 7
17 7
18 7
19 7
20 7
21 7
22 7
23 7
24 7
25 7
26 7
27 7
28 7
29 7
30 7
31 7
0 8
1 8
2 8
3 8
4 8
5 8
6 8
25 8
26 8
27 8
28 8
29 8
30 8
31 8
1 9
2 9
3 9
4 9
5 9
6 9
7 9
8 9
9 9
10 9
5 10
6 10
7 10
8 10
9 10
10 10
11 10
12 10
13 10
14 10
9 11
10 11
11 11
12 11
13 11
14 11
15 11
16 11
17 11
18 11
13 12
14 12
15 12
16 12
17 12
18 12
19 12
20 12
21 12
22 12
17 13
18 13
19 13
20 13
21 13
22 13
23 13
24 13
25 13
26 13
21 14
22 14
23 14
24 14
25 14
26 14
27 14
28 14
29 14
30 14
0 15
1 15
2 15
3 15
4 15
5 15
6 15
25 15
26 15
27 15
28 15
29 15
30 15
31 15
0 16
1 16
2 16
29 16
30 16
31 16
1 17
2 17
3 17
4 17
3 18
4 18
5 18
6 18
5 19
6 19
7 19
8 19
7 20
8 20
9 20
10 20
9 21
10 21
11 21
12 21
11 22
12 22
13 22
14 22
13 23
14 23
15 23
16 23
15 24
16 24
17 24
18 24
17 25
18 25
19 25
20 25
19 26
20 26
21 26
22 26
21 27
22 27
23 27
24 27
23 28
24 28
25 28
26 28
25 29
26 29
27 29
28 29
27 30
28 30
29 30
30 30
29 31
30 31
31 31
};

\nextgroupplot[
tick align=outside,
title={\huge \(\displaystyle \mathbf{A}^3\)},
x grid style={white!69.0196078431373!black},
xmin=-0.5, xmax=31.5,
xtick pos=both,
xtick style={color=black},
y dir=reverse,
y grid style={white!69.0196078431373!black},
ymin=-0.5, ymax=31.5,
ytick pos=left,
ytick style={color=black}
]
\addplot [semithick, color0, mark=*, mark size=1.25, mark options={solid}, only marks]
table {%
0 0
1 0
2 0
1 1
2 1
3 1
4 1
3 2
4 2
5 2
6 2
5 3
6 3
7 3
0 4
1 4
2 4
5 4
6 4
7 4
1 5
2 5
3 5
4 5
3 6
4 6
5 6
6 6
0 7
1 7
2 7
5 7
6 7
7 7
8 8
9 9
10 10
11 11
12 12
13 13
14 14
15 15
16 16
17 17
18 18
19 19
20 20
21 21
22 22
23 23
24 24
25 25
26 26
27 27
28 28
29 29
30 30
31 31
};

\nextgroupplot[
tick align=outside,
title={\huge \(\displaystyle \mathbf{A}^2\)},
x grid style={white!69.0196078431373!black},
xmin=-0.5, xmax=31.5,
xtick pos=both,
xtick style={color=black},
y dir=reverse,
y grid style={white!69.0196078431373!black},
ymin=-0.5, ymax=31.5,
ytick pos=left,
ytick style={color=black}
]
\addplot [semithick, color0, mark=*, mark size=1.25, mark options={solid}, only marks]
table {%
0 0
1 0
2 0
1 1
2 1
3 1
4 1
3 2
4 2
5 2
6 2
5 3
6 3
7 3
8 3
7 4
8 4
9 4
10 4
9 5
10 5
11 5
12 5
11 6
12 6
13 6
14 6
13 7
14 7
15 7
0 8
1 8
2 8
13 8
14 8
15 8
1 9
2 9
3 9
4 9
3 10
4 10
5 10
6 10
5 11
6 11
7 11
8 11
7 12
8 12
9 12
10 12
9 13
10 13
11 13
12 13
11 14
12 14
13 14
14 14
0 15
1 15
2 15
13 15
14 15
15 15
16 16
17 17
18 18
19 19
20 20
21 21
22 22
23 23
24 24
25 25
26 26
27 27
28 28
29 29
30 30
31 31
};

\nextgroupplot[
tick align=outside,
title={\huge \(\displaystyle \mathbf{A}^1\)},
x grid style={white!69.0196078431373!black},
xmin=-0.5, xmax=31.5,
xtick pos=both,
xtick style={color=black},
y dir=reverse,
y grid style={white!69.0196078431373!black},
ymin=-0.5, ymax=31.5,
ytick pos=left,
ytick style={color=black}
]
\addplot [semithick, color0, mark=*, mark size=1.25, mark options={solid}, only marks]
table {%
0 0
1 0
2 0
1 1
2 1
3 1
4 1
3 2
4 2
5 2
6 2
5 3
6 3
7 3
8 3
7 4
8 4
9 4
10 4
9 5
10 5
11 5
12 5
11 6
12 6
13 6
14 6
13 7
14 7
15 7
16 7
15 8
16 8
17 8
18 8
17 9
18 9
19 9
20 9
19 10
20 10
21 10
22 10
21 11
22 11
23 11
24 11
23 12
24 12
25 12
26 12
25 13
26 13
27 13
28 13
27 14
28 14
29 14
30 14
29 15
30 15
31 15
0 16
1 16
2 16
29 16
30 16
31 16
1 17
2 17
3 17
4 17
3 18
4 18
5 18
6 18
5 19
6 19
7 19
8 19
7 20
8 20
9 20
10 20
9 21
10 21
11 21
12 21
11 22
12 22
13 22
14 22
13 23
14 23
15 23
16 23
15 24
16 24
17 24
18 24
17 25
18 25
19 25
20 25
19 26
20 26
21 26
22 26
21 27
22 27
23 27
24 27
23 28
24 28
25 28
26 28
25 29
26 29
27 29
28 29
27 30
28 30
29 30
30 30
29 31
30 31
31 31
};
\end{groupplot}

\end{tikzpicture}
  \caption{Sparsity pattern of a 32 by 32 boundary wavelet analysis matrix, and its scale components.
           This pattern occurs for second degree wavelets. All non-zero entries are shown. Additional entries appear in comparison to the raw-convolution matrix (Figure~\ref{fig:conv_analysis}).}
  \label{fig:boundary_analysis}
\end{figure}
\begin{figure}[p]
\begin{tikzpicture}

\definecolor{color0}{rgb}{0.12156862745098,0.466666666666667,0.705882352941177}

\begin{groupplot}[group style={group size=4 by 1}]
\nextgroupplot[
tick align=outside,
title={\huge \(\displaystyle \mathbf{S}\)},
x grid style={white!69.0196078431373!black},
xmin=-0.5, xmax=31.5,
xtick pos=both,
xtick style={color=black},
y dir=reverse,
y grid style={white!69.0196078431373!black},
ymin=-0.5, ymax=31.5,
ytick pos=left,
ytick style={color=black}
]
\addplot [semithick, color0, mark=*, mark size=1.25, mark options={solid}, only marks]
table {%
0 0
4 0
7 0
8 0
15 0
16 0
0 1
1 1
4 1
5 1
7 1
8 1
9 1
15 1
16 1
17 1
0 2
1 2
4 2
5 2
7 2
8 2
9 2
15 2
16 2
17 2
0 3
1 3
4 3
5 3
7 3
8 3
9 3
15 3
17 3
18 3
0 4
1 4
4 4
5 4
7 4
8 4
9 4
15 4
17 4
18 4
0 5
1 5
4 5
5 5
7 5
8 5
9 5
10 5
15 5
18 5
19 5
0 6
1 6
4 6
5 6
7 6
8 6
9 6
10 6
15 6
18 6
19 6
0 7
1 7
4 7
5 7
7 7
9 7
10 7
19 7
20 7
0 8
1 8
4 8
5 8
7 8
9 8
10 8
19 8
20 8
0 9
1 9
2 9
4 9
5 9
6 9
7 9
9 9
10 9
11 9
20 9
21 9
0 10
1 10
2 10
4 10
5 10
6 10
7 10
9 10
10 10
11 10
20 10
21 10
0 11
1 11
2 11
4 11
5 11
6 11
7 11
10 11
11 11
21 11
22 11
0 12
1 12
2 12
4 12
5 12
6 12
7 12
10 12
11 12
21 12
22 12
0 13
1 13
2 13
4 13
5 13
6 13
7 13
10 13
11 13
12 13
22 13
23 13
0 14
1 14
2 14
4 14
5 14
6 14
7 14
10 14
11 14
12 14
22 14
23 14
1 15
2 15
5 15
6 15
11 15
12 15
23 15
24 15
1 16
2 16
5 16
6 16
11 16
12 16
23 16
24 16
1 17
2 17
3 17
4 17
5 17
6 17
7 17
11 17
12 17
13 17
24 17
25 17
1 18
2 18
3 18
4 18
5 18
6 18
7 18
11 18
12 18
13 18
24 18
25 18
1 19
2 19
3 19
4 19
5 19
6 19
7 19
12 19
13 19
25 19
26 19
1 20
2 20
3 20
4 20
5 20
6 20
7 20
12 20
13 20
25 20
26 20
1 21
2 21
3 21
4 21
5 21
6 21
7 21
12 21
13 21
14 21
26 21
27 21
1 22
2 22
3 22
4 22
5 22
6 22
7 22
12 22
13 22
14 22
26 22
27 22
2 23
3 23
4 23
6 23
7 23
13 23
14 23
27 23
28 23
2 24
3 24
4 24
6 24
7 24
13 24
14 24
27 24
28 24
2 25
3 25
4 25
6 25
7 25
8 25
13 25
14 25
15 25
28 25
29 25
2 26
3 26
4 26
6 26
7 26
8 26
13 26
14 26
15 26
28 26
29 26
2 27
3 27
4 27
6 27
7 27
8 27
14 27
15 27
29 27
30 27
2 28
3 28
4 28
6 28
7 28
8 28
14 28
15 28
29 28
30 28
2 29
3 29
4 29
6 29
7 29
8 29
14 29
15 29
16 29
30 29
31 29
2 30
3 30
4 30
6 30
7 30
8 30
14 30
15 30
16 30
30 30
31 30
3 31
4 31
7 31
8 31
15 31
16 31
31 31
};

\nextgroupplot[
tick align=outside,
title={\huge \(\displaystyle \mathbf{S}^3\)},
x grid style={white!69.0196078431373!black},
xmin=-0.5, xmax=31.5,
xtick pos=both,
xtick style={color=black},
y dir=reverse,
y grid style={white!69.0196078431373!black},
ymin=-0.5, ymax=31.5,
ytick pos=left,
ytick style={color=black}
]
\addplot [semithick, color0, mark=*, mark size=1.25, mark options={solid}, only marks]
table {%
0 0
4 0
7 0
0 1
1 1
4 1
5 1
7 1
0 2
1 2
4 2
5 2
7 2
1 3
2 3
5 3
6 3
1 4
2 4
5 4
6 4
2 5
3 5
4 5
6 5
7 5
2 6
3 6
4 6
6 6
7 6
3 7
4 7
7 7
8 8
9 9
10 10
11 11
12 12
13 13
14 14
15 15
16 16
17 17
18 18
19 19
20 20
21 21
22 22
23 23
24 24
25 25
26 26
27 27
28 28
29 29
30 30
31 31
};

\nextgroupplot[
tick align=outside,
title={\huge \(\displaystyle \mathbf{S}^2\)},
x grid style={white!69.0196078431373!black},
xmin=-0.5, xmax=31.5,
xtick pos=both,
xtick style={color=black},
y dir=reverse,
y grid style={white!69.0196078431373!black},
ymin=-0.5, ymax=31.5,
ytick pos=left,
ytick style={color=black}
]
\addplot [semithick, color0, mark=*, mark size=1.25, mark options={solid}, only marks]
table {%
0 0
8 0
15 0
0 1
1 1
8 1
9 1
15 1
0 2
1 2
8 2
9 2
15 2
1 3
2 3
9 3
10 3
1 4
2 4
9 4
10 4
2 5
3 5
10 5
11 5
2 6
3 6
10 6
11 6
3 7
4 7
11 7
12 7
3 8
4 8
11 8
12 8
4 9
5 9
12 9
13 9
4 10
5 10
12 10
13 10
5 11
6 11
13 11
14 11
5 12
6 12
13 12
14 12
6 13
7 13
8 13
14 13
15 13
6 14
7 14
8 14
14 14
15 14
7 15
8 15
15 15
16 16
17 17
18 18
19 19
20 20
21 21
22 22
23 23
24 24
25 25
26 26
27 27
28 28
29 29
30 30
31 31
};

\nextgroupplot[
tick align=outside,
title={\huge \(\displaystyle \mathbf{S}^1\)},
x grid style={white!69.0196078431373!black},
xmin=-0.5, xmax=31.5,
xtick pos=both,
xtick style={color=black},
y dir=reverse,
y grid style={white!69.0196078431373!black},
ymin=-0.5, ymax=31.5,
ytick pos=left,
ytick style={color=black}
]
\addplot [semithick, color0, mark=*, mark size=1.25, mark options={solid}, only marks]
table {%
0 0
16 0
0 1
1 1
16 1
17 1
0 2
1 2
16 2
17 2
1 3
2 3
17 3
18 3
1 4
2 4
17 4
18 4
2 5
3 5
18 5
19 5
2 6
3 6
18 6
19 6
3 7
4 7
19 7
20 7
3 8
4 8
19 8
20 8
4 9
5 9
20 9
21 9
4 10
5 10
20 10
21 10
5 11
6 11
21 11
22 11
5 12
6 12
21 12
22 12
6 13
7 13
22 13
23 13
6 14
7 14
22 14
23 14
7 15
8 15
23 15
24 15
7 16
8 16
23 16
24 16
8 17
9 17
24 17
25 17
8 18
9 18
24 18
25 18
9 19
10 19
25 19
26 19
9 20
10 20
25 20
26 20
10 21
11 21
26 21
27 21
10 22
11 22
26 22
27 22
11 23
12 23
27 23
28 23
11 24
12 24
27 24
28 24
12 25
13 25
28 25
29 25
12 26
13 26
28 26
29 26
13 27
14 27
29 27
30 27
13 28
14 28
29 28
30 28
14 29
15 29
16 29
30 29
31 29
14 30
15 30
16 30
30 30
31 30
15 31
16 31
31 31
};
\end{groupplot}

\end{tikzpicture}
  \caption{Sparsity pattern of a 32 by 32 boundary wavelet synthesis matrix, and its scale components.}
  \label{fig:boundary_synthesis}
\end{figure}
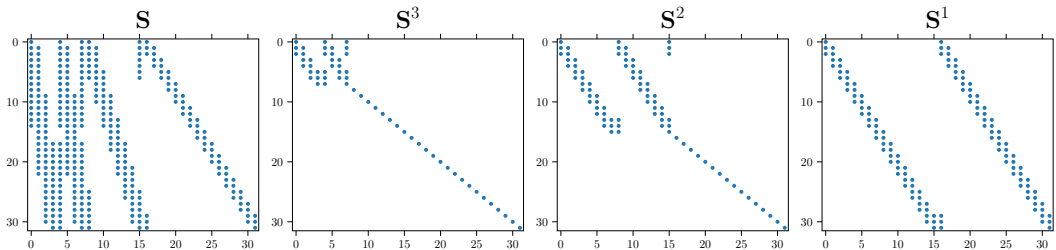

Figure~\ref{fig:conv_analysis} presented the single dimensional truncated analysis convolution matrices. Its synthesis counterpart with a transposed diagonal pattern is visible in Figure~\ref{fig:conv_synthesis}.  The effect of the boundary wavelet treatment discussed in section ~\ref{sec:boundary_wavelets} is illustrated in Figures \ref{fig:boundary_analysis} and \ref{fig:boundary_synthesis}. Additional entries created by the orthogonalization procedure are clearly visible.

\subsection{Two dimensional sparse-transformation matrix plots}
\begin{figure}[p]
  \includegraphics[width=\textwidth]{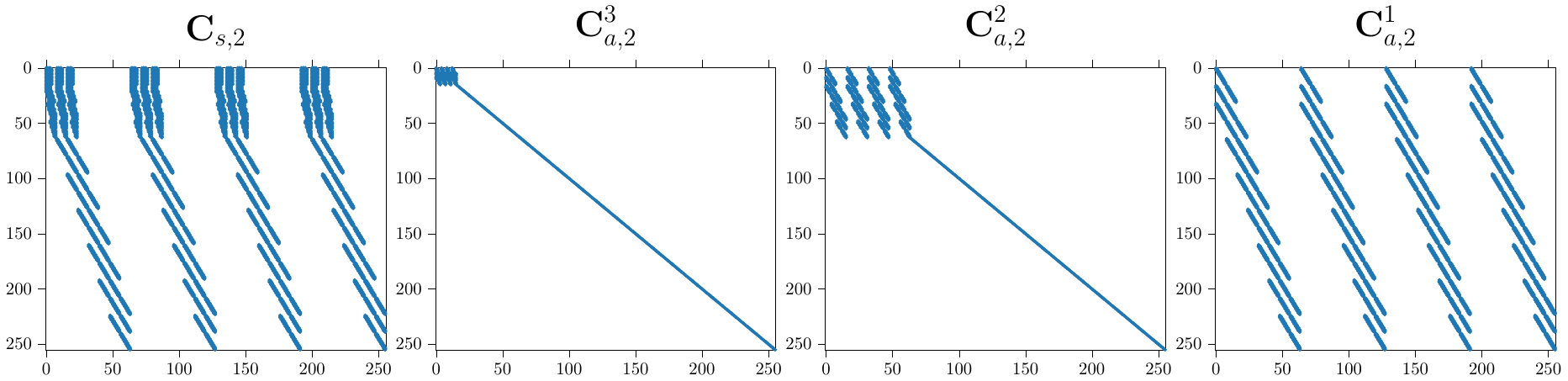}
  \caption{Sparsity patterns of two-dimensional synthesis convolution matrices. Upper indices indicate individual scale matrices.
           The transformation matrix on the left is the matrix-product of all three scale-matrices.}
  \label{fig:raw_synthesis2d}
\end{figure}
\begin{figure}[p]
  \includegraphics[width=\textwidth]{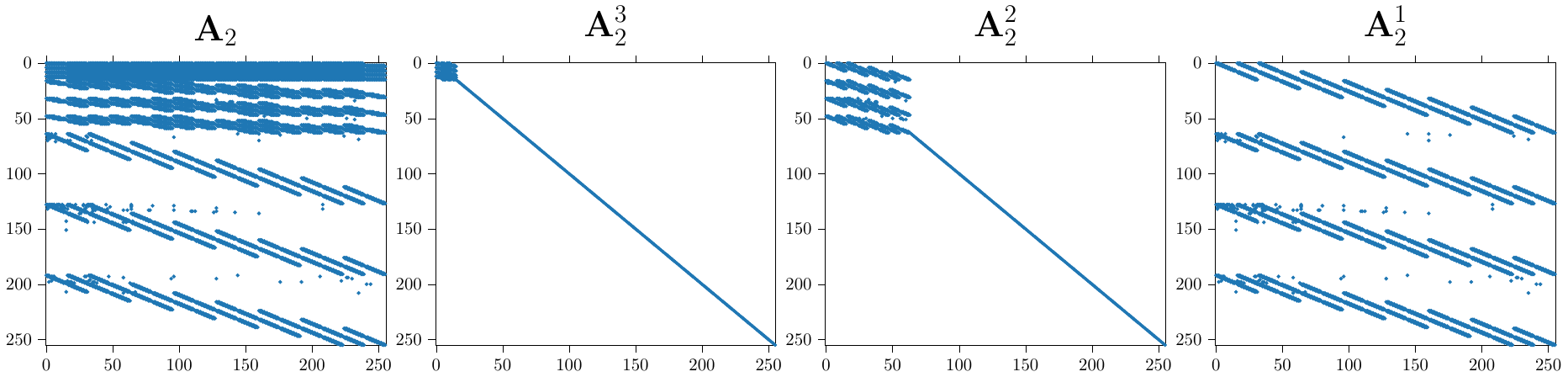}
  \caption{Sparsity patterns of two-dimensional analysis \ac{fwt}-matrices. Upper indices indicate individual scale matrices.
  The transformation matrix on the left is the matrix-product of all three scale-matrices.}
  \label{fig:boundary_analysis2d}
\end{figure}
\begin{figure}[p]
  \includegraphics[width=\textwidth]{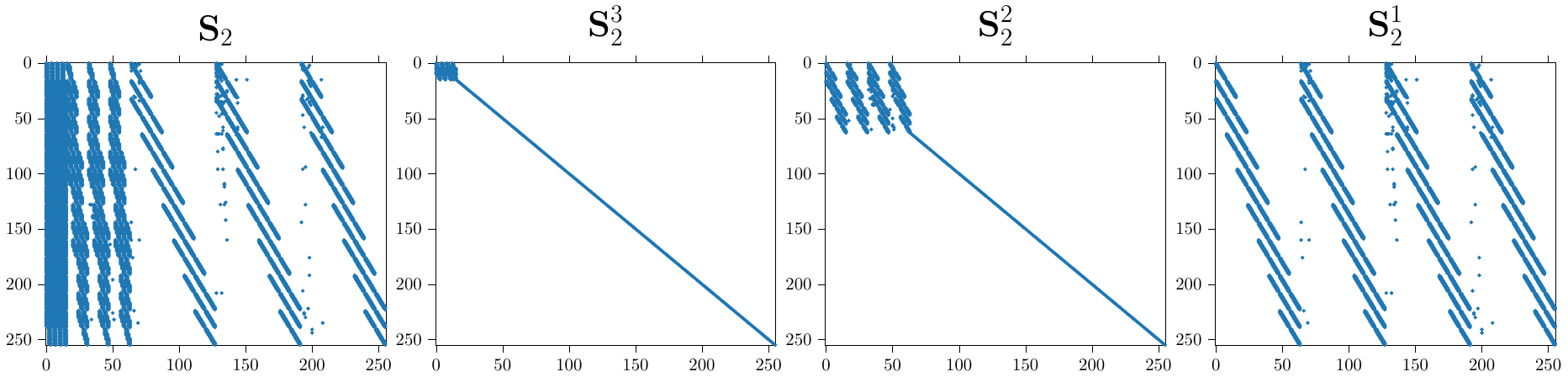}
  \caption{Sparsity patterns of two-dimensional synthesis or \ac{ifwt} matrices. Upper indices indicate individual scale matrices.
  The transformation matrix on the left is the matrix-product of all three scale-matrices.}
  \label{fig:boundary_synthesis2d}
\end{figure}
Figures \ref{fig:raw_analysis2d}, \ref{fig:raw_synthesis2d}, \ref{fig:boundary_analysis2d} and \ref{fig:boundary_synthesis2d}. Illustrate the differences between two dimensional convolution and orthogonal fast wavelet transformation matrices. In comparison to the convolution matrices in Figures \ref{fig:raw_analysis2d} and \ref{fig:raw_synthesis2d}, the orthogonalization procedure has created additional entries in the two dimensional analysis and synthesis matrices on display in Figures \ref{fig:boundary_analysis2d} and \ref{fig:boundary_synthesis2d}.

\subsection{Hyperparameters}\label{sec:hyperparameters}
Unless explicitly stated otherwise, we train with a batch size of 512 and a learning rate of 0.001 using Adam \citep{kingma2015adam} for ten epochs. For the 5 repetitions we report, the seed values are always 0,1,2,3,4.

\subsection{Detection of images from an unknown generator}\label{sec:unkown_gan}
To exemplarily study the detection of images from an unknown \ac{gan},
we remove the SN-DCGAN generated images from our \ac{lsun} training and validation sets.
10,000 SN-DCGAN images now appear exclusively in the test set, where they never influence the learning process.
The training hyperparameters are identical to those discussed in section~\ref{sec:train_celeba_lsun}.
We rebalance the data to contain equally many real and GAN-generated images.
The task now is to produce a real or fake label in the presence of fake images which were not present during the optimization process.
For this initial proof-of-concept investigation, we apply only the simple Haar wavelet.
We find that our approach does allow detection of the SN-DCGAN samples on \ac{lsun}-bedrooms, without their presence in the training set.
Using the $\ln$-scaled Haar-Wavelet packages, we achieve for the unknown \ac{gan} a result of $78.8 \pm 1.8\%$,
with a max of $81.7\%$. For the real and artificial generators present in the training set, we achieve $98.6 \pm 0.1\%$, with a max of 98.6\%. The accuracy improvement in comparison to Table~\ref{tab:source_seperation_results} likely is due to the binary classification problem here instead of the multi-classification one before.
This result indicates that the wavelet representation alone can allow a transfer to unseen generators, or other manipulations, to some extent.

\subsection{\ac{lsun} perturbation analysis}
\begin{table}[tbp]
  \centering
  \caption{Perturbation analysis of the db3 wavelet-packet- and pixel- \ac{cnn} classifiers on \ac{lsun}-bedrooms.}
  \label{tab:suppl_perturbation}
  \begin{tabular}{lrrr}\toprule
                            &                & \multicolumn{2}{c}{Accuracy on \acs{lsun}[\%]}\\ \cmidrule(lr){3-4}
    Method                  & Perturbation   & $\max$ & $\mu \pm \sigma$         \\ \midrule
    CNN-$\ln$-db3 (ours) & center-crop    & \textbf{95.68}  & 95.49$\pm$ 0.26          \\
    CNN-Pixel (ours)        & center-crop    & 92.03  & 90.04$\pm$ 1.18          \\ \midrule
    CNN-$\ln$-db3 (ours) & rotation       & 91.74  & 90.84$\pm$ 0.90          \\
    CNN-Pixel (ours)        & rotation       & \textbf{92.63}  & 91.99$\pm$ 0.97          \\ \midrule
    CNN-$\ln$-db3 (ours) & jpeg           & 84.73          & 84.25  $\pm$  00.33  \\
    CNN-Pixel (ours)        & jpeg           & \textbf{89.95} & 89.25  $\pm$  00.48  \\ \bottomrule
  \end{tabular}
\end{table}
We study the effect of image cropping, rotation, and jpeg compression on our classifiers in table~\ref{tab:suppl_perturbation}.
We re-trained both classifiers on the perturbed data as described in section~\ref{sec:train_celeba_lsun}.
The center crop perturbation randomly extracts images centers while preserving at least 80\% of the original width or height.
Both the pixel and packet approaches are affected by center cropping yet can classify more than 90\% of all images correctly.
In this case, the log-scaled db3-wavelet packets can outperform the pixel-space representation on average.
Rotation and jpeg compression change the picture.
The rotation perturbation randomly rotates all images by up to $15^\circ$ degrees.
The jpeg perturbation randomly compresses images with a compression factor drawn from $U(70, 90)$, without subsampling.
The wavelet-packet representation is less robust to rotation than its pixel counterpart.
For jpeg-compression, the effect is more pronounced. Looking at Figure~\ref{fig:mean_packets_ffhq} suggests that our classifiers rely on high-frequency information. Since jpeg-compression removes this part of the spectrum, perhaps explaining the larger drop in mean accuracy.

\begin{figure}
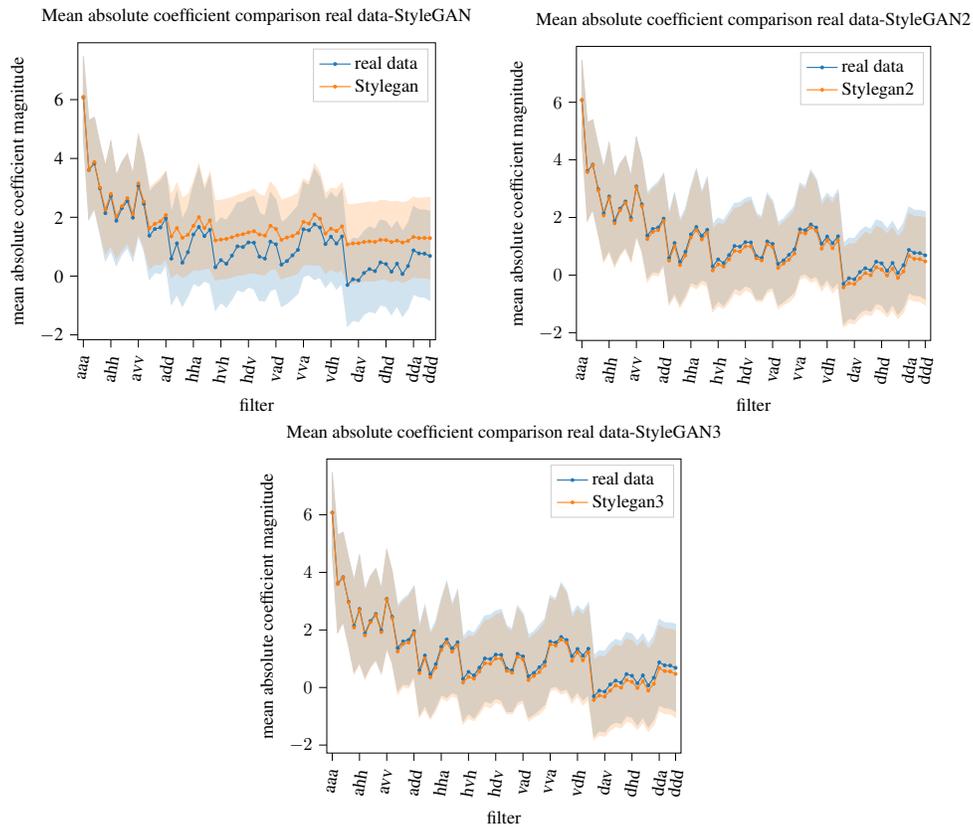

\centering
\includestandalone[width=.45\linewidth]{./figures/ffhq_hard/absolute_coeff_comparison_stylegan}
\includestandalone[width=.45\linewidth]{./figures/ffhq_hard/absolute_coeff_comparison_stylegan2}
\includestandalone[width=.45\linewidth]{./figures/ffhq_hard/absolute_coeff_comparison_stylegan3}
\caption{Mean and standard deviation plots for the three styleGAN variants on \ac{ffhq}.}\label{fig:ffhq_hard_mean_std}
\end{figure}

\subsection{Results on the full \ac{ffpp} with and without image corruption}\label{sec:fullffpp}
\begin{table}[p]
  \centering
  \caption{Classification accuracies for the complete \ac{ffpp}-data set. Results for no compression (C0), high quality compression (C23) as well as aggressive compression (C40) are tabulated. For comparison, we additionally cite the results presented in \cite{rossler2019faceforensics++}. We were able to find the parameter count for XceptionNet in \cite{Chollet2017XceptionDL} and added it to the table. XceptionNet was pretrained in Image-Net and is therefore not directly comparable to the rest of the table.}\label{tab:fullffpp}
  \begin{tabular}{l r r r} \toprule
    & &\multicolumn{2}{c}{Accuracy on \ac{ffpp} C0 [\%]} \\ \cmidrule(lr){3-4}
    Method                    & parameters &  max      & $\mu \pm \sigma$   \\ \midrule
    CNN-pixel (ours)          & 57k        &  97.24    &  $94.87\pm 1.47$   \\
    CNN-db4   (ours)          & 109k       &  96.07    &  $95.75\pm 0.23$   \\
    CNN-sym4  (ours)          & 109k       &  97.05    &  $96.45\pm 0.35$   \\  \addlinespace

    Steg. Feat. + SVM (\citeauthor{fridrich2012rich})     & -          &  \textit{97.63}    & -                  \\
    \citeauthor{cozzolino2017recasting}         & -          &  98.57    & -                  \\
    \citeauthor{bayar2016deep}          & -          &  98.74    & -                  \\
    \citeauthor{rahmouni2017distinguishing}          & -          &  97.03    & -                  \\
    MesoNet     (\citeauthor{afchar2018mesonet})              & -          &  95.23    & -                  \\ \addlinespace
    XceptionNet (\citeauthor{Chollet2017XceptionDL})   & \textbf{22,856k}     &  \textbf{99.26}   & -                  \\ \midrule
    & &\multicolumn{2}{c}{Accuracy on \ac{ffpp} C23 [\%]} \\ \cmidrule(lr){3-4}
    CNN-pixel (ours)          &  57k       &  80.41    & $77.97\pm 1.74$    \\
    CNN-db4   (ours)          &  109k      &  \textit{85.41} & $84.49\pm 0.83$ \\
    CNN-sym4  (ours)          &  109k      &  84.48    & $83.03\pm 1.16$    \\  \addlinespace

    Steg. Feat. + SVM (\citeauthor{fridrich2012rich})     & -          &  70.97    & -                  \\
    \citeauthor{cozzolino2017recasting}          & -          &  78.45    & -                  \\
    \citeauthor{bayar2016deep}          & -          &  82.87    & -                  \\
    \citeauthor{rahmouni2017distinguishing}          & -          &  79.08    & -                  \\
    MesoNet      (\citeauthor{afchar2018mesonet})             & -          &  83.10    & -                  \\ \addlinespace
    XceptionNet  (\citeauthor{Chollet2017XceptionDL})             &\textbf{22,856k} &  \textbf{95.73} & -        \\ \midrule

    & &\multicolumn{2}{c}{Accuracy on \ac{ffpp} C40 [\%]} \\ \cmidrule(lr){3-4}
    CNN-pixel (ours)          &  57k       &  \textit{75.54}    & $ 74.11 \pm 1.87$  \\
    CNN-db4   (ours)          &  109k      &  71.91    & $ 71.91 \pm 1.03$  \\
    CNN-sym4  (ours)          &  109k      &  72.92    & $ 69.91 \pm 2.24$  \\ \addlinespace

    Steg. Feat. + SVM (\citeauthor{fridrich2012rich})  & -          &  55.98    & -                  \\
    \citeauthor{cozzolino2017recasting}  & -          &  58.69    & -                  \\
    \citeauthor{bayar2016deep}  & -          &  66.84    & -                  \\
    \citeauthor{rahmouni2017distinguishing}  & -          &  61.18    & -                  \\
    MesoNet      (\citeauthor{afchar2018mesonet})    & -          &  70.47    & -                  \\ \addlinespace
    XceptionNet  (\citeauthor{Chollet2017XceptionDL})  &\textbf{22,856k} &  \textbf{81.00} & -                 \\ \bottomrule
  \end{tabular}
\end{table}
We train our networks as previously discussed in sections~\ref{sec:train_celeba_lsun} and \ref{sec:ffpp}. Table~\ref{tab:fullffpp} lists results for a pixel-cnn as well as wavelet-packet-architectures using db4 and sym4-wavelets. In comparison to various benchmark methods our networks perform competitively. Figures~\ref{fig:ffpp_wp_grouped} and ~\ref{fig:ffpp_wp_single} tell us that classical face modification methods produce fewer artifacts in the wavelet-domain. This observation probably explains the drop in performance for the network processing the uncompressed features. We note competitive performance in the high-quality (C23) case.

\begin{figure}
  \centering
\begin{tikzpicture}

\definecolor{crimson2143940}{RGB}{214,39,40}
\definecolor{darkgray176}{RGB}{176,176,176}
\definecolor{darkorange25512714}{RGB}{255,127,14}
\definecolor{forestgreen4416044}{RGB}{44,160,44}
\definecolor{lightgray204}{RGB}{204,204,204}
\definecolor{mediumpurple148103189}{RGB}{148,103,189}
\definecolor{steelblue31119180}{RGB}{31,119,180}

\begin{axis}[
legend cell align={left},
legend style={fill opacity=0.8, draw opacity=1, text opacity=1, draw=lightgray204},
tick align=outside,
tick pos=left,
title={Mean absolute db4 coefficient comparison},
x grid style={darkgray176},
xlabel={filter},
xmin=-1, xmax=64,
xtick style={color=black},
xtick=      {  0,  5, 10, 15, 20, 25, 30, 35, 40, 45, 50, 55, 60, 63},
xticklabels={aaa,ahh,avv,add,hha,hvh,hdv,vad,vva,vdh,dav,dhd,dda,ddd},
xticklabel style={rotate=80.0},
y grid style={darkgray176},
ylabel={mean absolute coefficient magnitude},
ymin=-7.56557063778814, ymax=1.01443073583653,
ytick style={color=black}
]
\addplot [semithick, steelblue31119180, mark=*, mark size=0.75, mark options={solid}]
table {%
0 0.624289514045366
1 -2.1880521053294
2 -1.90620615665719
3 -2.88477015632967
4 -3.83450996087794
5 -3.14861836061595
6 -4.21226367581952
7 -3.72978963096189
8 -3.37573822392761
9 -4.14027706034248
10 -2.69349527496073
11 -3.58477285121361
12 -5.06063915994781
13 -4.71348616471888
14 -4.6395725595397
15 -4.21076060462242
16 -5.63112255515618
17 -5.04443087463038
18 -6.02326571628744
19 -5.58246958747385
20 -4.76873127139689
21 -4.46052133279354
22 -4.96257441345254
23 -4.63835713748547
24 -6.3735517812305
25 -6.16573936058506
26 -6.12815785785681
27 -5.84854284467237
28 -5.6565089162372
29 -5.52416656718274
30 -5.38177922374517
31 -5.20005182782992
32 -5.47369753196975
33 -5.79966202982718
34 -4.96440370496146
35 -5.26747260897007
36 -6.34668996257898
37 -6.09472763722632
38 -6.04978786081862
39 -5.69553233328256
40 -4.51593936327997
41 -4.71661121338902
42 -4.33888396117618
43 -4.62236727755177
44 -5.56779397813098
45 -5.15010904018405
46 -5.56939842010876
47 -5.12074443570938
48 -7.0382171700611
49 -6.93497254933264
50 -6.9535280462629
51 -6.77465150745576
52 -6.63711947835086
53 -6.64662706785769
54 -6.42264130313993
55 -6.42909290983917
56 -6.6760606101539
57 -6.42074888173596
58 -6.73500152766579
59 -6.49545149328643
60 -5.94384278805723
61 -5.98691979056812
62 -6.06303262714443
63 -6.05890673207696
};
\addlegendentry{Real data}
\addplot [semithick, darkorange25512714, mark=*, mark size=0.75, mark options={solid}]
table {%
0 0.610176088791356
1 -2.21364183282095
2 -1.93985064139228
3 -2.9417802569404
4 -3.91228376995091
5 -3.19701395326585
6 -4.32403187041813
7 -3.80851575842372
8 -3.46500697886772
9 -4.26004345052475
10 -2.74802100195444
11 -3.66268470256422
12 -5.23318029243548
13 -4.85891809585521
14 -4.77686605602593
15 -4.31661076296981
16 -5.81199076263321
17 -5.15331674465119
18 -6.20277650937306
19 -5.73270583975895
20 -4.89968078892739
21 -4.56294077635375
22 -5.11394615871253
23 -4.77258393561804
24 -6.54045190996314
25 -6.35138841659504
26 -6.30696811481457
27 -6.02191523494759
28 -5.85186643678841
29 -5.71155175243115
30 -5.55599776240022
31 -5.36076497135863
32 -5.65176869642343
33 -5.96274228230315
34 -5.08236990710015
35 -5.4002401827174
36 -6.49300301139228
37 -6.25126546785063
38 -6.20863772546984
39 -5.84285350050313
40 -4.67759348518218
41 -4.88836930240872
42 -4.46233973152927
43 -4.76066675316899
44 -5.76753607950959
45 -5.33610056279327
46 -5.74577387516943
47 -5.27698715818777
48 -7.11148910969171
49 -7.02865199533845
50 -7.05426191176545
51 -6.90043081448801
52 -6.77471867458419
53 -6.76511342526938
54 -6.58540828071298
55 -6.57771678892856
56 -6.83236107971169
57 -6.60309613971484
58 -6.8683501465814
59 -6.65387755682209
60 -6.15373073805238
61 -6.18595336403753
62 -6.25070642472024
63 -6.23469037377289
};
\addlegendentry{Deepfakes}
\addplot [semithick, forestgreen4416044, mark=*, mark size=0.75, mark options={solid}]
table {%
0 0.612251760115121
1 -2.21722241814155
2 -1.93581160994766
3 -2.93741726932632
4 -3.91733536275199
5 -3.22593725029667
6 -4.33251067557092
7 -3.82485891834661
8 -3.46061693771441
9 -4.27197390710563
10 -2.75931778921785
11 -3.68184286005909
12 -5.25514126681204
13 -4.89262093720119
14 -4.80177417678396
15 -4.3559261153397
16 -5.84840695536956
17 -5.18114055317687
18 -6.24390242557452
19 -5.76034185656514
20 -4.92197788219694
21 -4.59155431772568
22 -5.13843420647082
23 -4.79247198362387
24 -6.57780336425188
25 -6.38390774248944
26 -6.33695662312798
27 -6.0496801903058
28 -5.88977725179301
29 -5.73400457023305
30 -5.59696279972653
31 -5.38696750162134
32 -5.72211178124331
33 -6.0368583888129
34 -5.11864111448723
35 -5.45177525331645
36 -6.56365177738444
37 -6.33300789380677
38 -6.27266634245617
39 -5.91292626486109
40 -4.6907741193239
41 -4.92134490881543
42 -4.47699316847962
43 -4.79428996737015
44 -5.8086154084142
45 -5.38879704729891
46 -5.78619353214415
47 -5.32409407904881
48 -7.17557057535066
49 -7.09677716894729
50 -7.11807961799646
51 -6.95891646822797
52 -6.85799920952116
53 -6.8482896238859
54 -6.66313103561121
55 -6.64578712545859
56 -6.89052606395254
57 -6.65573870463898
58 -6.92605138367427
59 -6.70219145789287
60 -6.21478380208767
61 -6.24209556380579
62 -6.30896077061286
63 -6.27849433415863
};
\addlegendentry{NeuralTextures}
\addplot [semithick, crimson2143940, mark=o, mark size=0.75, mark options={solid}]
table {%
0 0.624357766540916
1 -2.22189107061621
2 -1.9458091179519
3 -2.92432754799031
4 -3.89552264152948
5 -3.20652649077845
6 -4.26841734234496
7 -3.77978192058324
8 -3.43586566541549
9 -4.19150877000626
10 -2.75356722576463
11 -3.63607414251223
12 -5.10472857911727
13 -4.76357985376346
14 -4.69079524903581
15 -4.26363748777223
16 -5.74547214350918
17 -5.12727204220761
18 -6.07930068249798
19 -5.64178810609484
20 -4.87323727851882
21 -4.54601604810664
22 -5.04509398269334
23 -4.70282750389935
24 -6.38778598186138
25 -6.19210519729528
26 -6.15997228723063
27 -5.89137059292334
28 -5.71319119276482
29 -5.56472331736522
30 -5.44941734121763
31 -5.25493398905155
32 -5.56466489614727
33 -5.81340255351911
34 -5.03441235107039
35 -5.30125077741117
36 -6.33698477416138
37 -6.0894968397201
38 -6.05796274275137
39 -5.71339863146753
40 -4.60244741230576
41 -4.78977111356261
42 -4.41674374605954
43 -4.68135983311453
44 -5.61956794486895
45 -5.21145684173962
46 -5.60388926253403
47 -5.16592421978912
48 -7.00395015157979
49 -6.89658013200551
50 -6.92556491636228
51 -6.74917826948495
52 -6.62293205360783
53 -6.62002253497479
54 -6.42749166990614
55 -6.42061296428873
56 -6.68112128533084
57 -6.43480742907051
58 -6.72417222376076
59 -6.49178289579223
60 -5.99297939749397
61 -6.01828160547495
62 -6.09400660424145
63 -6.074563043088
};
\addlegendentry{Face2Face}
\addplot [semithick, mediumpurple148103189, mark=o, mark size=0.75, mark options={solid}]
table {%
0 0.624430673399043
1 -2.21086167726067
2 -1.9209748570114
3 -2.90701896392188
4 -3.86707366016429
5 -3.18511462485219
6 -4.23217598586482
7 -3.75463135305994
8 -3.3860492841647
9 -4.14423134173069
10 -2.71288553662646
11 -3.60159993972923
12 -5.03598915653235
13 -4.70482799454639
14 -4.63775798073053
15 -4.21927979561142
16 -5.62859829324036
17 -5.06209872020984
18 -5.9481845897735
19 -5.55291342997844
20 -4.802531234244
21 -4.49021187566755
22 -4.97340210353298
23 -4.63948688387583
24 -6.25491938139356
25 -6.07582702610225
26 -6.0282606505875
27 -5.78723398614262
28 -5.61817491923863
29 -5.46952368167234
30 -5.36683350444688
31 -5.17367696102197
32 -5.46045104874592
33 -5.72692693776489
34 -4.96085549162213
35 -5.23642174999136
36 -6.24444470585453
37 -6.00565214954168
38 -5.97856856949853
39 -5.64474063746459
40 -4.52582564737294
41 -4.70757927654106
42 -4.3439357973694
43 -4.60942314031085
44 -5.528897015752
45 -5.12603076243581
46 -5.52231849645845
47 -5.09090647436058
48 -6.90845773372783
49 -6.80889851278981
50 -6.82140372365394
51 -6.65673918485083
52 -6.52863772872126
53 -6.53264882364022
54 -6.33940852441769
55 -6.33451156242663
56 -6.55077679229049
57 -6.32136099439368
58 -6.60594061740957
59 -6.38877891024227
60 -5.89293662584307
61 -5.9205455750992
62 -6.00291049910372
63 -5.98363533581689
};
\addlegendentry{FaceSwap}
\end{axis}

\end{tikzpicture}\hfill%
\begin{tikzpicture}

\definecolor{crimson2143940}{RGB}{214,39,40}
\definecolor{darkgray176}{RGB}{176,176,176}
\definecolor{darkorange25512714}{RGB}{255,127,14}
\definecolor{forestgreen4416044}{RGB}{44,160,44}
\definecolor{lightgray204}{RGB}{204,204,204}
\definecolor{mediumpurple148103189}{RGB}{148,103,189}
\definecolor{steelblue31119180}{RGB}{31,119,180}

\begin{axis}[
legend cell align={left},
legend style={fill opacity=0.8, draw opacity=1, text opacity=1, draw=lightgray204},
tick align=outside,
tick pos=left,
title={Mean absolute sym4 coefficient comparison},
x grid style={darkgray176},
xlabel={filter},
xmin=-1, xmax=64,
xtick style={color=black},
xtick=      {  0,  5, 10, 15, 20, 25, 30, 35, 40, 45, 50, 55, 60, 63},
xticklabels={aaa,ahh,avv,add,hha,hvh,hdv,vad,vva,vdh,dav,dhd,dda,ddd},
xticklabel style={rotate=80.0},
y grid style={darkgray176},
ylabel={mean absolute coefficient magnitude},
ymin=-7.60053738822448, ymax=1.09677585502982,
ytick style={color=black}
]
\addplot [semithick, steelblue31119180, mark=*, mark size=0.75, mark options={solid}]
table {%
0 0.701316185928317
1 -2.09056438660115
2 -1.81924690512146
3 -2.72505862007176
4 -3.66434061846009
5 -3.1907121462449
6 -3.89545984068411
7 -3.45983144592549
8 -3.41503732251807
9 -3.85286066585246
10 -2.94983127687219
11 -3.44045652504098
12 -4.77420728932969
13 -4.38100804568837
14 -4.42946706000727
15 -4.00515667194601
16 -5.71677979307738
17 -5.34000278382349
18 -5.97765231799628
19 -5.554711974653
20 -4.89015682936216
21 -4.93683473638931
22 -5.0375345621023
23 -5.04735055933729
24 -6.33496929785009
25 -5.98949900985538
26 -6.16202285031367
27 -5.77435083903182
28 -5.51280410405645
29 -5.51312197380886
30 -5.27179656350656
31 -5.29078418838113
32 -5.51816367921018
33 -5.89753069200465
34 -5.12320037113422
35 -5.49717967697413
36 -6.34213749156852
37 -6.16263559922011
38 -6.07595824088016
39 -5.82736908801419
40 -4.53862267262979
41 -4.81484073231651
42 -4.57788872396787
43 -4.88338624718995
44 -5.56247659100499
45 -5.27150651102289
46 -5.65335677702439
47 -5.37172652198361
48 -7.07170476612677
49 -6.96430650239641
50 -6.98014352492313
51 -6.83862900316063
52 -6.68176093930104
53 -6.75306333077855
54 -6.48547898024811
55 -6.58318695161713
56 -6.72466308550256
57 -6.49942304638513
58 -6.80059060678781
59 -6.59981303758445
60 -6.0289615329856
61 -6.14556244117669
62 -6.15521268377608
63 -6.26534903035461
};
\addlegendentry{Real data}
\addplot [semithick, darkorange25512714, mark=*, mark size=0.75, mark options={solid}]
table {%
0 0.689275334611724
1 -2.11281636306134
2 -1.85687375001383
3 -2.77829842454043
4 -3.74264635454077
5 -3.24014853142358
6 -4.00563349104405
7 -3.53570462619693
8 -3.5054133061187
9 -3.97071177107506
10 -3.01157524983961
11 -3.51908674196068
12 -4.94735829496347
13 -4.52447552161736
14 -4.56490801827878
15 -4.10790974296437
16 -5.89197351443415
17 -5.44421420315108
18 -6.14648726929554
19 -5.69298520057898
20 -5.01962466486476
21 -5.04068922797851
22 -5.18404291716427
23 -5.17817697616584
24 -6.49697174863933
25 -6.16667984665913
26 -6.33199346748147
27 -5.93705537054715
28 -5.7045040302604
29 -5.69805653291284
30 -5.43926072991143
31 -5.44843592534561
32 -5.7016134990694
33 -6.06054864296305
34 -5.24722644894444
35 -5.63015372564481
36 -6.49241243281911
37 -6.32367696598506
38 -6.23603675590437
39 -5.97739749196948
40 -4.70518951901919
41 -4.99108623285558
42 -4.7034023104962
43 -5.02255807185354
44 -5.76456341982687
45 -5.46090362535814
46 -5.82978552823467
47 -5.52951493770187
48 -7.14494386170416
49 -7.05714722848041
50 -7.07471398251158
51 -6.95682688697176
52 -6.81794056701717
53 -6.87163853955448
54 -6.64116383979879
55 -6.72618475252561
56 -6.87475935203691
57 -6.67484381734396
58 -6.92567445854098
59 -6.74972058257838
60 -6.23455948333671
61 -6.34260288066899
62 -6.33487710136311
63 -6.43640352993721
};
\addlegendentry{Deepfakes}
\addplot [semithick, forestgreen4416044, mark=*, mark size=0.75, mark options={solid}]
table {%
0 0.685453047618953
1 -2.11968180766894
2 -1.85510277244944
3 -2.77777348199352
4 -3.75787932150546
5 -3.26553504925011
6 -4.00611871875298
7 -3.54831379855476
8 -3.52174766308998
9 -3.98358354866909
10 -3.03703610013666
11 -3.54099551641419
12 -4.96941179458784
13 -4.54605626426248
14 -4.5946485010247
15 -4.14043315914319
16 -5.96098632692557
17 -5.49119980403835
18 -6.20011349209426
19 -5.72929545991993
20 -5.06722313965153
21 -5.07777837325915
22 -5.23331480955888
23 -5.20908167713453
24 -6.5437727322001
25 -6.19704600786047
26 -6.37366371123617
27 -5.96628250787467
28 -5.74408565415242
29 -5.72064890237936
30 -5.48717239544475
31 -5.47911559134297
32 -5.79329201168229
33 -6.1361951067747
34 -5.29221407801672
35 -5.68905479603994
36 -6.56866287351244
37 -6.3930396042256
38 -6.30577997990628
39 -6.04110915749506
40 -4.73631518824764
41 -5.0229301414606
42 -4.73374763064888
43 -5.05877071905386
44 -5.81091380327342
45 -5.50142976270606
46 -5.88022265956007
47 -5.57545592565501
48 -7.20520496807656
49 -7.12348390835774
50 -7.13729480831679
51 -7.01639246422054
52 -6.89364122012225
53 -6.9442174213207
54 -6.7128956830731
55 -6.78741270612443
56 -6.93547095779304
57 -6.72468722885477
58 -6.99107959882599
59 -6.80299250086801
60 -6.28966608880519
61 -6.38443263947149
62 -6.39557565992038
63 -6.48025897405837
};
\addlegendentry{NeuralTextures}
\addplot [semithick, crimson2143940, mark=o, mark size=0.75, mark options={solid}]
table {%
0 0.701443434881901
1 -2.12277707409319
2 -1.85625853033714
3 -2.76137357569203
4 -3.72861221151054
5 -3.2475760808252
6 -3.95279960437196
7 -3.50872486929122
8 -3.47788125188039
9 -3.90629762628355
10 -3.01374044344251
11 -3.4946740675161
12 -4.82025692764775
13 -4.42870764545696
14 -4.48004880496969
15 -4.05276376383242
16 -5.84067412122543
17 -5.42502981065747
18 -6.02806739768419
19 -5.61209419334961
20 -4.99355732658866
21 -5.02262945092922
22 -5.11144587337005
23 -5.10929284111463
24 -6.34705413914809
25 -6.02071126843826
26 -6.18829135614856
27 -5.81968032522996
28 -5.55908140651497
29 -5.55026149288798
30 -5.32970184017427
31 -5.34004044451342
32 -5.61860035501375
33 -5.91179636474517
34 -5.20185501753367
35 -5.53430964187219
36 -6.33904691639925
37 -6.16132431174288
38 -6.09408303994107
39 -5.85151251167557
40 -4.63158716726656
41 -4.88905248177474
42 -4.65860699276076
43 -4.94051069694395
44 -5.61997206501117
45 -5.33383334765075
46 -5.69293756393858
47 -5.41681415127807
48 -7.04127474962508
49 -6.9357730926569
50 -6.95534702372633
51 -6.82362997561338
52 -6.66811108051052
53 -6.72914093797017
54 -6.49202805437958
55 -6.57902368116225
56 -6.72798360938452
57 -6.52175480066671
58 -6.78865312570601
59 -6.60418375974746
60 -6.07314379788207
61 -6.18097321861306
62 -6.181084549375
63 -6.28390295635651
};
\addlegendentry{Face2Face}
\addplot [semithick, mediumpurple148103189, mark=o, mark size=0.75, mark options={solid}]
table {%
0 0.70043785141495
1 -2.11350317457153
2 -1.83259478639831
3 -2.74455861362814
4 -3.69852816579775
5 -3.22652826033469
6 -3.91509236902179
7 -3.48285067940498
8 -3.42502347449389
9 -3.85459753218083
10 -2.9693111012721
11 -3.45623946742098
12 -4.75223584414838
13 -4.37150987831275
14 -4.42925908912649
15 -4.01280397220939
16 -5.72031031381358
17 -5.36120009237195
18 -5.91071017248083
19 -5.53132235430827
20 -4.92529857597373
21 -4.96879543201015
22 -5.04716713932775
23 -5.04939063620075
24 -6.22207754160598
25 -5.90628924313851
26 -6.06655967258723
27 -5.71813865031097
28 -5.47519919606574
29 -5.46048831149235
30 -5.25669635561986
31 -5.26502867051139
32 -5.5075697109494
33 -5.83064701983512
34 -5.11968790121645
35 -5.46899945274541
36 -6.24460149239387
37 -6.07682081365643
38 -6.01074081097012
39 -5.78001705639434
40 -4.54685042904217
41 -4.80791166665193
42 -4.58159975657334
43 -4.87148800472357
44 -5.52673838327321
45 -5.24969918510235
46 -5.60940725277289
47 -5.34385943681397
48 -6.94947168971298
49 -6.84355987761388
50 -6.8569258385037
51 -6.72652317334275
52 -6.57553621934407
53 -6.63834055632605
54 -6.40521080882819
55 -6.48892859143073
56 -6.60525154127071
57 -6.40397960382894
58 -6.67732176443611
59 -6.49694546016757
60 -5.97779049707728
61 -6.07839353391077
62 -6.09465736864294
63 -6.18934856623927
};
\addlegendentry{FaceSwap}
\end{axis}

\end{tikzpicture}
  \caption{
    Mean $\ln$-wavelet packet plots for all elements of the FaceForensics++ training set for the db4-wavelet (left) and the sym4-wavelet(right).
    A full list of the order of packets can be found in Figure~\ref{fig:packetlabels_natural}.
    Generally the frequencies increase from left to right.
    We observe a difference between the means of machine learning based fakes (Deepfakes~\citep{deepfake2022github} and NeuralTextures~\citep{thies2019deferred}) and the other methods,
    especially in the higher frequencies i.e. on the right hand side (best observed when zoomed into the plot).
  }
  \label{fig:ffpp_wp_single}
  \includestandalone[width=0.48\linewidth]{./figures/ffpp/absolute_coeff_comparison_db4_grouped}\hfill%
  \includestandalone[width=0.48\linewidth]{./figures/ffpp/absolute_coeff_comparison_sym4_grouped}
  \caption{%
    Wavelet packet mean and standard deviation using db4-wavelets (left) and sym4-wavelets (right) for the original data, machine learning-based methods, and classic computer-vision methods.
    Again, we see a difference in the mean of machine learning-based models and the other sources as well as the original data.
  }
  \label{fig:ffpp_wp_grouped}
\end{figure}

\begin{table}[tbp]
  \centering
  \caption{Confusion matrix for the $\ln$-db4-\ac{cnn} on the extended \ac{ffhq} problem.}\label{tab:conf_ffhq_hard}
  \begin{tabular}{l r r r r}\toprule
             & \multicolumn{4}{c}{Predicted label}          \\\cmidrule(lr){2-5}
  True label & \ac{ffhq} & StyleGAN & StyleGAN2 & StyleGAN3 \\ \midrule
  \ac{ffhq}  & 4626      &  0       &   14      &   360     \\
  StyleGAN   &   4       &  4991    &   0       &   5       \\
  StyleGAN2  &  13       &  0       &  4913     &   74      \\
  StyleGAN3  & 237       &  0       &  80       &   4683    \\ \bottomrule
  \end{tabular}
\end{table}

\subsection{Packet labels}
\begin{figure}
  \centering
\begin{tikzpicture}
\node at (0,0)    {aaa}; \node at (1,0)    {aav}; \node at (2,0)    {avv}; \node at (3,0)    {ava}; \node at (4,0)    {vva}; \node at (5,0)    {vvv}; \node at (6,0)    {vav}; \node at (7,0)    {vaa};
\node at (0,-.5)  {aah}; \node at (1,-.5)  {aad}; \node at (2,-.5)  {avd}; \node at (3,-.5)  {avh}; \node at (4,-.5)  {vvh}; \node at (5,-.5)  {vvd}; \node at (6,-.5)  {vad}; \node at (7,-.5)  {vah};
\node at (0,-1)   {ahh}; \node at (1,-1)   {ahd}; \node at (2,-1)   {add}; \node at (3,-1)   {adh}; \node at (4,-1)   {vdh}; \node at (5,-1)   {vdd}; \node at (6,-1)   {vhd}; \node at (7,-1)   {vhh};
\node at (0,-1.5) {aha}; \node at (1,-1.5) {ahv}; \node at (2,-1.5) {adv}; \node at (3,-1.5) {ada}; \node at (4,-1.5) {vda}; \node at (5,-1.5) {vdv}; \node at (6,-1.5) {vhv}; \node at (7,-1.5) {vha};
\node at (0,-2)   {hha}; \node at (1,-2)   {hhv}; \node at (2,-2)   {hdv}; \node at (3,-2)   {hda}; \node at (4,-2)   {dda}; \node at (5,-2)   {ddv}; \node at (6,-2)   {dhv}; \node at (7,-2)   {dha};
\node at (0,-2.5) {hhh}; \node at (1,-2.5) {hhd}; \node at (2,-2.5) {hdd}; \node at (3,-2.5) {hdh}; \node at (4,-2.5) {ddh}; \node at (5,-2.5) {ddd}; \node at (6,-2.5) {dhd}; \node at (7,-2.5) {dhh};
\node at (0,-3)   {hah}; \node at (1,-3)   {had}; \node at (2,-3)   {hvd}; \node at (3,-3)   {hvh}; \node at (4,-3)   {dvh}; \node at (5,-3)   {dvd}; \node at (6,-3)   {dad}; \node at (7,-3)   {dah};
\node at (0,-3.5) {haa}; \node at (1,-3.5) {hav}; \node at (2,-3.5) {hvv}; \node at (3,-3.5) {hva}; \node at (4,-3.5) {dva}; \node at (5,-3.5) {dvv}; \node at (6,-3.5) {dav}; \node at (7,-3.5) {daa};
\end{tikzpicture}
  \caption{Labels for the transforms of level 3 we showed previously in the frequency order. For an excellent introduction to the theory behind the frequency order see~\cite{jensen2001ripples}.}
  \label{fig:packetlabels}
\end{figure}

\begin{figure}
  \centering
  \small
  \begin{multicols}{8}
    \begin{enumerate}
      \item aaa
      \item aah
      \item aav
      \item aad
      \item aha
      \item ahh
      \item ahv
      \item ahd
      \item ava
      \item avh
      \item avv
      \item avd
      \item ada
      \item adh
      \item adv
      \item add
      \item haa
      \item hah
      \item hav
      \item had
      \item hha
      \item hhh
      \item hhv
      \item hhd
      \item hva
      \item hvh
      \item hvv
      \item hvd
      \item hda
      \item hdh
      \item hdv
      \item hdd
      \item vaa
      \item vah
      \item vav
      \item vad
      \item vha
      \item vhh
      \item vhv
      \item vhd
      \item vva
      \item vvh
      \item vvv
      \item vvd
      \item vda
      \item vdh
      \item vdv
      \item vdd
      \item daa
      \item dah
      \item dav
      \item dad
      \item dha
      \item dhh
      \item dhv
      \item dhd
      \item dva
      \item dvh
      \item dvv
      \item dvd
      \item dda
      \item ddh
      \item ddv
      \item ddd
    \end{enumerate}
  \end{multicols}
  \caption{Order of wavelet packet coefficients as used in the mean absolute coefficient plots.}
  \label{fig:packetlabels_natural}
\end{figure}

Figure~\ref{fig:packetlabels_natural} displays the 1d ordering and Figure~\ref{fig:packetlabels} the 2d ordering of the 64 image-patches of the level 3 wavelet packet coefficients.
Each reappear many times in the paper.

\subsection{Reproducibility Statement}
To ensure the reproducibility of this research project, we release our wavelet toolbox and the source code for our experiments in the supplementary material.
The wavelet toolbox ships multiple unit tests to ensure correctness.
Our code is carefully documented to make it as accessible as possible.
We have consistently seeded the Mersenne twister used by PyTorch to initialize our neural networks with 0, 1, 2, 3, 4. Since the seed is always set, rerunning our code always produces the same results.
Outliers can make it hard to reproduce results if the seed is unknown. To ensure we do not share outliers without context, we report mean and standard deviations of five runs for all neural network experiments.

\end{document}